\documentclass{article}
\usepackage{arxiv_mda,times}   
\usepackage{ifthen}
\newcommand{\revealIdentity}{true}
\newcommand{\redacted}[1]{\ifthenelse{\equal{\revealIdentity}{true}}{#1}{\texttt{REDACTED}}}
\AtBeginDocument{%
  \ifthenelse{\equal{\revealIdentity}{true}}{%
    \ifcsname iclrfinalcopy\endcsname\iclrfinalcopy\fi
  }{\author{Anonymous authors}}%
}
\usepackage{amsmath,amssymb,amsthm}
\usepackage{hyperref}
\usepackage{xurl} 
\usepackage{enumitem}
\usepackage{booktabs}
\usepackage{graphicx}
\usepackage{subcaption}
\usepackage{algorithm}
\usepackage{algpseudocode}
\makeatletter
\providecommand{\theHALG@line}{}
\renewcommand{\theHALG@line}{\thealgorithm.\arabic{ALG@line}}
\makeatother
\usepackage{tikz}
\usetikzlibrary{arrows.meta, positioning, fit, backgrounds, decorations.pathmorphing, calc, shapes.geometric}
\hypersetup{colorlinks=true,linkcolor=blue,citecolor=blue,urlcolor=blue}
\usepackage{bbm}
\usepackage{verbatim}
\usepackage{fvextra}   
\usepackage{listings} 
\lstdefinestyle{pytiny}{%
  language=Python, basicstyle=\ttfamily\tiny, breaklines=true, breakatwhitespace=false,
  columns=fullflexible, keepspaces=true, showstringspaces=false, commentstyle=\color{gray},
  aboveskip=2pt, belowskip=0pt, xleftmargin=3pt, frame=none,
  literate={"}{{\char34}}1{'}{{\char39}}1}   
\usepackage[most]{tcolorbox}   
\definecolor{promptblue}{HTML}{2B4A9B}
\newtcolorbox{promptboxenv}[1]{%
  enhanced, breakable, colback=promptblue!5!white, colframe=promptblue,
  boxrule=0.5pt, arc=2pt, left=5pt, right=5pt, top=2pt, bottom=2.5pt,
  fonttitle=\bfseries\small, coltitle=white,
  attach boxed title to top left={xshift=8pt, yshift=-2.4mm},
  boxed title style={colback=promptblue, arc=2pt, boxrule=0pt},
  title={#1}}
\newcommand{\promptfile}[2]{%
  \begin{promptboxenv}{#1}%
  \VerbatimInput[breaklines,breakanywhere,breaksymbolleft={},breaksymbolright={},fontsize=\scriptsize]{#2}%
  \end{promptboxenv}}
\usepackage{xspace}
\usepackage{circuitikz}
\usepackage[capitalize]{cleveref}

\newcommand{\dopo}{\mathrm{do}}
\newcommand{\Ind}{\mathbbm{1}}
\newcommand{\E}{\mathbb{E}}

\newcommand{\D}{\mathcal{D}}
\newcommand{\data}{\mathcal{D}}

\newcommand{\design}{\chi}
\newcommand{\designSpace}{\mathcal{X}}  
\newcommand{\perturb}{\delta}
\newcommand{\Nexpt}{K}
\newcommand{\expt}{k}
\newcommand{\real}{\mathbb{R}}
\newcommand{\Dtrain}{\D_\text{tr}}
\newcommand{\Dtest}{\D_\text{te}}

\newcommand{\test}{\text{test}}
\newcommand{\loss}{\mathcal{L}}
\newcommand{\summary}{s}
\newcommand{\summaryLearned}{s_{\phi}}
\newcommand{\target}{F}
\newcommand{\predtarget}{\hat{F}} 
\newcommand{\queryDist}{\mathcal{Q}}
\newcommand{\terminal}{d}
\newcommand{\terminalSpace}{\Delta}

\newcommand{\VOI}{\text{VoI}}
\newcommand{\EIG}{\text{EIG}}

\newcommand{\inputs}{a}
\newcommand{\inputSpace}{\mathcal{A}}
\newcommand{\initial}{z_0}
\newcommand{\gauss}{\mathcal{N}}
\newcommand{\new}{\mathrm{new}}

\newcommand{\Nnoise}{N_{\text{noise}}}

\newcommand{\zrebound}{\textsc{z-rebound}\xspace}
\newcommand{\hsag}{\textsc{h-sag}\xspace}
\newcommand{\nafatigue}{\textsc{na-fatigue}\xspace}
\newcommand{\carebound}{\textsc{ca-rebound}\xspace}
\newcommand{\dtype}{\textsc{d-type}\xspace}
\newcommand{\textbookM}{\textsc{textbook-M}\xspace}

\newcommand{\Apprentice}{\textsc{Apprentice}\xspace}
\newcommand{\DPagent}{\textsc{DP-Agent}\xspace}
\newcommand{\glucose}{\textsc{GlucoseBench}\xspace}
\newcommand{\DP}{\textsc{DiscoverPhysics}\xspace}
\providecommand{\ActiveChem}{\textsc{ActiveSciBench-Chem}\xspace}
\providecommand{\SciLab}{\textsc{LLM-AutoSciLab}\xspace}

\providecommand{\CHEMbench}{\textsc{ChemBench}\xspace}

\providecommand{\ChemBench}{\CHEMbench}
\newcommand{\DPbench}{\textsc{ForceBench}\xspace}
\newcommand{\HHbench}{\textsc{NeuronBench}\xspace}

\newcommand{\boxing}{\textsc{BoxingGym}\xspace}

\newcommand{\nparams}{C}              
\newcommand{\nparticlesModel}{N_m}    
\newcommand{\nparticlesParams}{N_p}   
\newcommand{\nroundsParams}{R_p}    
\newcommand{\nparticlesLatent}{N_z}   

\newcommand{\eat}[1]{} 

\title{\textbf{Model Discovery Agent:\\LLM-assisted Bayesian experiment design
for data-efficient discovery of mechanistic world models}}
\author{Kevin Murphy \\
  Dept. Computer Science\\
  Univ. British Columbia, Canada.
  }
\date{\today}

\begin{document}
\maketitle

\begin{abstract}
  A primary goal of science is to learn mechanistic world models from limited experimental data, both to explain observations and to predict novel interventions. We introduce the Model Discovery Agent (MDA), which combines LLM proposals for $\mathcal M$-open model discovery, experiment design based on Value of Information, and approximate Bayesian inference over model structures, parameters, and stochastic latent trajectories. We apply MDA to learn symbolic reaction rate laws for ChemBench \citep{kabra2026autoscilab},   partially observed ODE models for GlucoseBench  \citep{xie2018simglucose,kovatchev2009insilico},  and partially observed SDE models for a new stochastic single-neuron simulator we create. In the appendix, we also show results on various other domains from BoxingGym \citep{gandhi2025boxinggym}. We show that MDA has improved sample efficiency compared to various baseline methods, and the learned models are good predictors but also provide interpretable abstractions of each domain.
 \end{abstract}

\eat{
\begin{abstract}
  A primary goal of science is to learn mechanistic world models from limited experimental data, both to explain observations and to predict novel interventions. We introduce the Model Discovery Agent (MDA), which combines LLM proposals for $\mathcal M$-open discovery, information-directed experiment design, and approximate inference over model structures, parameters, and stochastic latent trajectories. We evaluate static symbolic rate laws on \CHEMbench\ \citep{kabra2026autoscilab}, partially observed ODE models of glucose--insulin dynamics, and a stochastic single-neuron benchmark modeled by partially observed SDEs. MDA improves held-out prediction relative to domain-specific and in-context baselines, with gains that vary by domain and output channel. Its expandable latent models provide interpretable predictive abstractions without claiming unique identification of the underlying physiology. Additional results on BoxingGym \citep{gandhi2025boxinggym} illustrate the framework's scope.
  \end{abstract}
}

\section{Introduction}
\label{sec:intro}

In this paper, we develop a new algorithm for learning
a mechanistic world model (aka causal world model) from a small
amount of data.\footnote{
We use the term ``mechanistic world model'' following  \citep{Posner2026}.
These are models composed of modular, scientifically meaningful building blocks
or mechanisms, as in a structural causal model \citep{pearl2009}.
In this paper,  we  restrict ourselves to  ``level 2'' causality
\citep{Bareinboim2022};
this can be handled with standard decision-theoretic machinery
\citep{Dawid2015,Mlodozeniec2025},
and does not need the more complex machinery
required for ``level 3'' counterfactual reasoning
\citep{Dawid2000}.
Note that  the philosophy of science literature
makes a  distinction between causal models and mechanistic models
(see e.g., \citep{Machamer2000,Craver2006,Woodward2004,Weber2008,Glennan2017,Batterman2014});
we use the term mechanistic broadly here.
}
Such models can support scientific explanation and prediction
\citep{Salmon1984,Shmueli2010,Krenn2022,Messeri2024,Bajorath2025,
  Serre2025,Kramer2026,Posner2026}.
In addition, they can be used to answer  \emph{interventional questions}
 \citep{pearl2009,richens2024robust}:
not ``what will happen?'' but ``what \emph{would} happen if I did
action $a$?'' (e.g., predicting the effect of administering a new drug to a patient).

Unfortunately, a model with latent variables/ mechanisms is typically \emph{unidentifiable}
  from observation alone: the passive data underdetermines it, and
  interventions --- perturbing the system and watching how it
  responds --- can distinguish explanations that passive data cannot. But experiments are expensive (a
  lab assay, a clinical trial), which makes the
  operative problem \emph{data efficiency}: identify the model,
  well enough to answer future queries, in as few experiments as
  possible. This is the classical remit of \emph{Bayesian experimental
  design} (BED) --- choose the intervention whose outcome is most
  informative \citep{lindley1956,chaloner1995,rainforth2024} --- increasingly combined with hypothesis generation in recent active
  theory-learning systems \citep{piriyakulkij2024,Elteto2026,Prystawski2026}.
  Here we study this combination across explicit algebraic models and
  partially observed deterministic and stochastic dynamics.

  We define this problem formally in \cref{sec:problem}.
Then in  \cref{sec:methods}
  we present a solution, which we call the \textbf{Model Discovery Agent
  (MDA)}, which combines three ingredients:
approximate Bayesian inference over a hierarchy of models, parameters, and latent trajectories,
implemented using particle methods or MAP estimation;
a large language model (LLM), which is used as a way to propose
new models  when the current hypothesis space is detected to be
insufficient (this is needed to tackle the
$\mathcal{M}$-\emph{open} regime, where the true model may lie
\emph{outside} the current hypothesis class
\citep{bernardo1994,MacKinlay2016,kelter2020mopen});
and an experiment designer based on maximizing the Value of Information
(see e.g., \citep{rainforth2024}).
The loop fits and weights candidate models, selects an experiment,
and uses predictive discrepancies to guide further proposals.

In \cref{sec:results},
we empirically evaluate  MDA on problems of increasing complexity,
and compare to various baselines. On \CHEMbench
\citep{kabra2026autoscilab}, MDA discovers static symbolic rate laws
for chemical reactions. On
\glucose \citep{xie2018simglucose,kovatchev2009insilico},
MDA learns partially observed deterministic ODEs,
including latent compartments that encode delayed input responses.
Finally, on \HHbench, a novel single-neuron electrophysiology
benchmark we introduce
based on a stochastic version of the Hodgkin-Huxley (HH) model,
MDA learns a partially observed SDE.
In the appendix (\cref{app:boxing}),  we also report results of applying MDA
to  \boxing\ \citep{gandhi2025boxinggym}, which contains a variety of
simple problems from psychology to ecology,
which we model using Generalized Linear Models
and probabilistic programs.
Across these benchmarks, MDA discovers interpretable models
which provide strong predictive performance on novel
inputs and interventions.

\section{Problem statement}
\label{sec:problem}

\paragraph{Agent-environment interface.}
We consider an agent interacting with an unknown ``blackbox'' dynamical system
(c.f., \citep{Ljung1998}).
The outputs or observations are
$y_{1:T}$, for $y_t \in \mathcal{Y} \subset \real^{d_y}$.
The inputs or  control signals
$\inputs_{1:T}$ are chosen by the agent using
$\inputs_t \sim \pi(\cdot \mid \inputs_{1:t-1}, y_{1:t-1}, t)$,
with $\inputs_t\in\inputSpace\subset\real^{d_\inputs}$,
where $\pi$ is the policy.\footnote{
For interventional prediction, we assume actions depend only on recorded
history and context (including state estimates computed from them), with
policy randomization independent of system noise. This excludes unmeasured
confounding of action selection; causal identification additionally requires
adequate action coverage and suitable model/identification assumptions
\citep{Cornish2026}. Without these, same-policy observational prediction may
still be accurate, but need not transfer to a new intervention or policy.
}
The agent  can also optionally specify the initial condition
of the system, $\initial$;
if it is not specified,  its
distribution must be inferred from the training data and context.
Finally, the agent can optionally  apply a  perturbation or intervention
$\perturb$ that modifies the system parameters,
yielding $\theta' = \dopo(\perturb,\theta)$.
(An omitted perturbation leaves the system unchanged.)
We define an experiment design
as the tuple $\design = (\perturb, \initial, \pi)$.
\eat{
In this paper, we focus on open-loop control,
which is a special case where the policy $\pi$
deterministically supplies a fixed sequence of inputs, $\inputs_{1:T}$ (possibly no-op),
so $\pi$ depends only on $t$ within the episode,
and not the previous actions or observations;
this simplifies the design space,
and avoids some issues with closed-loop identifiability
\citep{Ljung1998}.
}

\paragraph{Experimental protocol.}
The agent is presented with some background context $C$,
and an optional initial  dataset
$\data_0 = \{ (\design^i, y_{1:T}^i): i=1:N_0 \}$,
where each sample is drawn from the system using
$y_{1:T}^i \sim p^\ast(\cdot|\design^i)$.
(We also record
the sampled inputs  $\inputs_{1:T}^i$, but omit this from notation, for brevity.)
\eat{
for the open-loop deterministic policies we use in this paper,
they are recoverable from $\design^i$,
so we omit them from the notation.
}
The agent is then given a budget of $\Nexpt$ turns to interact with the system.
At each step, it designs an experiment
$\design^\expt = (\perturb^\expt, \initial^{\expt}, \pi^{\expt})$,
records  one trajectory (episode) $y_{1:T}^{\expt}$ from the environment,
and updates
$\data_{0:\expt}=\data_{0} \cup\{ (\design^i,y_{1:T}^i): i=1:\expt \}$.
The agent can  use this data to update its beliefs about the underlying
model, $p_\expt(m)=p(m\mid C,\data_{0:\expt},\mathcal M_\expt)$,
where the candidate support $\mathcal M_\expt$ may grow or shrink,
and this belief can be used to design the next experiment,
and to make predictions about the future.
The objective is low held-out forecasting loss within the experiment budget;
MDA uses model-discrimination information gain as a tractable acquisition
surrogate (\cref{eq:voi}), not the held-out outcomes themselves.
See \cref{alg:api} for the pseudocode for this experimental protocol,
that is agnostic to the form of the agent.

\label{sec:eval}
\paragraph{Conditional forecasting and evaluation.}
We evaluate predictions on held-out experiments,
not recovery of a unique true mechanism,
since distinct latent models can have
similar interventional predictions \citep{beckers2019,Dyer2023,Cornish2026}.
Given a new design,
$\design\sim\queryDist$,
the oracle rolls out the policy to collect the ground truth
trajectory $(\inputs_{1:T}, y_{1:T})$.
Let $t_c$ be a cut point, through which the agent observes
the initial history $\mathcal H_{t_c}=(\inputs_{1:t_c},y_{1:t_c}^{\rm obs})$,
including observation masks when entries are missing.
(If $t_c=0$, the agent does not see any such prefix.)
The agent must then predict a distribution over (functions of)
future observables, given this partial history $\mathcal H_{t_c}$,
the training data $\D=\D_{0:\expt}$,
and any relevant context or metadata $C$,
by computing the  posterior predictive distribution:\footnote{
%
If $\design$ specifies an
open-loop policy $\pi$, then
it defines the whole input sequence $\inputs_{1:T}$ up front
(so the policy is just a function of
time, not the past observations or actions).
If $\pi$ is a closed-loop policy,
then future inputs $\inputs_{t_c+1:T}$
must be generated jointly with future observations $y_{t_c+1:T}$,
when forecasting the future.
(This paper focuses on learning action-conditioned world models,
not on  learning the policy of another agent from observational data.)
}
\begin{equation}
 p\!\left(\target(Y_{t_c+1:T})\mid \design,
       \mathcal H_{t_c},\D,C\right).
 \label{eq:conditional_target}
\end{equation}
The target function $\target$ maps the future observation sequence to a
scalar, feature vector, or trajectory, depending on the benchmark.
For \CHEMbench it is
the scalar log-rate $\target(y)=\log(1+y)\in\real$, matching nMSLE;
for \glucose it is
the future observed trajectory
$\target(y_{t_c+1:T})=y_{t_c+1:T}\in\real^{(T-t_c)d_y}$,
where $d_y=2$ is the number of observed variables per step;
and for \HHbench,
it is  a vector of summary features,
$\target(y_{1:T}) \in \real^6$
(e.g., neuron spike count --- see \cref{eq:hhstochfeatures} for details).
We condition on the observed history
so far $\mathcal H_{t_c}$ for this unit (if $t_c>0$),
plus the training episodes $\D_{0:\expt}$ (collected by the agent),
plus any supplied context or metadata $C$.

We mostly focus on
predicting the mean of \cref{eq:conditional_target},
and measuring performance with squared error,
although we do consider distributional forecasting metrics
in \cref{app:CRPS}.
In the simplest setting, where the prediction target
is the raw observation sequence,
$\target(y_{1:T}) = y_{1:T}$,
there is no missing data,
and $t_c=0$ (so there is no history prefix),
the normalized mean squared error metric we use is
\begin{equation}
 L_{\rm nMSE}=\frac1{d_y}\sum_{i=1}^{d_y}
 \frac{\sum_{q=1}^Q \sum_{t=1}^T(\hat y_{qti}-y_{qti})^2}{N s_i^2}
 \label{eq:nmse}
\end{equation}
Here $q$ indexes held-out queries (novel experiments/designs),
$i$ observation channels,
$t$ future time points,
$y_{qti}$ is the observation.
$\hat{y}_{qti}$ is the agent's prediction,
$s_i$ is a scaling factor that is benchmark specific,
and $N=Q T$ is the number of observed time points in the test set.
We give the general case in \cref{app:eval}.
\eat{
This measures held-out interventional prediction. Novel test designs need
not be out of distribution; transfer beyond the training design distribution
is a stronger requirement (cf.\ \citealp{richens2024robust}).
}

\section{Method}
\label{sec:methods}

\paragraph{Overview.}
The MDA method is visualized in  \cref{fig:loop};
see \cref{alg:mda} for detailed pseudocode.
At each step, the agent updates its belief state $p_\expt=p(m|\data_{0:\expt})$,
which is a posterior distribution over models or hypotheses $m$,
represented as a weighted finite set.
\eat{
In some cases we use nested particle methods
to represent uncertainty over the
parameters and latent states \citep{chopin2013smc2}.
In other cases, we use a simpler MAP estimate,
combined with a BIC approximation to the marginal likelihood;
in this case,  the belief state then denotes
approximate model weights rather than an exact Bayesian posterior.
}
MDA then  chooses the next experiment by
maximizing the value of information,
$\design_{\expt+1}=\arg\max_{\design \in \designSpace} \VOI(\design, p_\expt)$.
It runs the experiment and updates its dataset
by appending  $\data_{\expt+1}$.
After $\Nexpt$ rounds, the agent is asked to forecast the outcomes
to some novel experimental conditions.
We describe this in more detail below, but for full details,
see \cref{app:method}.

\paragraph{The three-level hierarchy.}
\begin{figure}[t]
\centering
\resizebox{0.4\linewidth}{!}{%
\begin{tikzpicture}[>=Stealth, node distance=13mm,
    lat/.style={circle,draw,minimum size=8.5mm,inner sep=0pt},
    obs/.style={circle,draw,fill=gray!18,minimum size=8.5mm,inner sep=0pt},
    ctrl/.style={rectangle,draw,rounded corners,minimum size=6.5mm,inner sep=2pt,fill=blue!6},
    par/.style={rectangle,draw,thick,minimum size=8mm,inner sep=3pt,fill=orange!16},
    mod/.style={rectangle,draw,thick,rounded corners,minimum size=8mm,inner sep=3pt,fill=green!12}]
  \node[lat] (z0) {$z_0$};
  \node[lat,right=of z0] (z1) {$z_1$};
  \node[lat,right=of z1] (z2) {$z_2$};
  \node[right=9mm of z2] (zd) {$\cdots$};
  \node[lat,right=9mm of zd] (zT) {$z_T$};
  \draw[->] (z0)--(z1); \draw[->] (z1)--(z2); \draw[->] (z2)--(zd); \draw[->] (zd)--(zT);
  \node[obs,below=of z1] (o1) {$y_1$};
  \node[obs,below=of z2] (o2) {$y_2$};
  \node[obs,below=of zT] (oT) {$y_T$};
  \draw[->] (z1)--(o1); \draw[->] (z2)--(o2); \draw[->] (zT)--(oT);
  \node[ctrl,above=9mm of z1] (x1) {$\inputs_1$};
  \node[ctrl,above=9mm of z2] (x2) {$\inputs_2$};
  \node[ctrl,above=9mm of zT] (xT) {$\inputs_T$};
  \draw[->,dashed,gray] (x1)--(z1);
  \draw[->,dashed,gray] (x2)--(z2); \draw[->,dashed,gray] (xT)--(zT);
  \node[ctrl,above=12mm of x2] (policy) {$\pi$};
  \draw[->,blue!60!black] (policy)--(x1);
  \draw[->,blue!60!black] (policy)--(x2);
  \draw[->,blue!60!black] (policy)--(xT);
  \node[par,below=12mm of o2] (th) {$\theta$};
  \node[mod,left=18mm of th] (m) {$m$};
  \draw[->,green!50!black,thick] (m)--(th);
  \draw[->,orange!70!black] (th) to[out=140,in=-80] (z1.-70);
  \draw[->,orange!70!black] (th) to[out=90,in=-90] (z2.-90);
  \draw[->,orange!70!black] (th) to[out=30,in=-110] (zT.-150);
  \node[right=20mm of th] (a) {$\perturb$};
  \draw[->,red!75!black,line width=0.9pt,decorate,
        decoration={zigzag,segment length=2.6mm,amplitude=1.1mm,post length=2mm}]
        (a)--(th);
  \node[red!75!black,font=\small] at ($(a)!0.5!(th)+(0,3.2mm)$) {$\dopo(\perturb)$};
\end{tikzpicture}
}
\caption{\textbf{The three-level hierarchy} (Eq.~\eqref{eq:ssm}). A model structure $m$
(green) selects a parameterized mechanism $\theta$ (orange), which governs the latent trajectory
$z_{1:T}$ (white); the trajectory emits the noisy observations $y_{1:T}$ (grey).
The experiment may specify $z_0$ directly (otherwise it is inferred) and specifies a policy $\pi$ that selects inputs
$\inputs_{1:T}$ (blue). For closed-loop control, each $\inputs_t$ also depends on
past actions and observations (feedback edges omitted); our experiments use open-loop policies.
An intervention $\dopo(\perturb)$ (lightning bolt) changes the parameters from $\theta$ to $\theta'$.
}
\label{fig:ssm}
\end{figure}
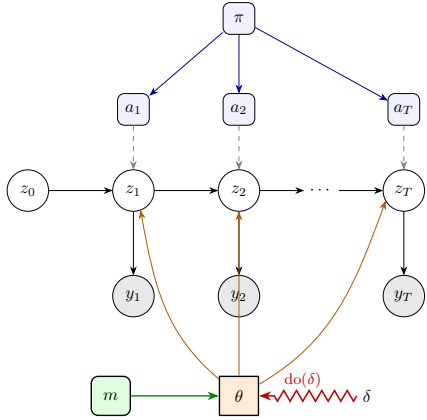

In the general case,
MDA represents the unknown world by the 3 level hierarchy
corresponding to the State Space Model shown in \cref{fig:ssm}.
In equations, this corresponds to
\begin{align}
  m\sim p(m),\qquad \theta\sim p(\theta\mid m), \qquad
  \theta' = \dopo(\perturb,\theta)\\
  z_t\sim p_m\!\left(z_t\mid z_{t-1},\inputs_t;\theta'\right),\qquad
  y_t\sim p_m(y_t\mid z_t;\theta'),
  \label{eq:ssm}
\end{align}
for $t=1{:}T$, with $z_0$ specified by $\design$ when available (otherwise
assigned an initial-state distribution), and $\inputs_t$ selected by $\pi$.
The discrete structure $m$ specifies
what the latent variables $z_t$ are, and how they depend on each other,
as well as the functional forms for all the distributions;
$\theta$ specifies the numerical  parameters;
$z_{1:T}$ is the latent trajectory; and
$y_{1:T}$ is what is measured.
MDA must perform inference over structures, parameters, and
---when dynamics or initial states are uncertain---latent paths.

\paragraph{Computing the posterior  over models.}
The belief state over models, $p_\expt=p(m|\data_{0:\expt}) \propto p(m) p(\data_{0:\expt} \mid m)$, is
represented as a discrete distribution over a finite candidate pool.
This is computed using Bayes' rule over the models in the pool,
and then keeping the top $\nparticlesModel$, if a cap on pool size is imposed.

\paragraph{Expanding the hypothesis space.}
Bayes rule can move probability mass between different models
in its current hypothesis class.
However,
in the $\mathcal M$-open case
\citep{bernardo1994,kelter2020mopen},
we acknowledge that our initial hypothesis class may be insufficient to capture the truth,
so we need  a way to grow it on demand.
To do this, we compute
a \emph{predictive check}, i.e., we evaluate
the performance of the current best model on a novel (input,output) pair
(c.f., \citep{kelter2020mopen}),
and/or check the fit on all the currently collected data.
If the error is too large,  MDA  \emph{expands} the
hypothesis space by prompting the LLM to suggest one or more novel models
conditioned on the observations and context. Where available, the prompt
also includes the current model pool and residual diagnostics.
Note that this hypothesis expansion step is the only part of MDA
that uses an LLM.
After  expanding the pool of models,
we compute the evidence for each new model and renormalize,
keeping the top $\nparticlesModel$ models.
Crucially, we do not require the LLM to propose the right model in a single shot;
instead,  it can suggest local mutations to previous hypotheses,
and thus can
iteratively refine the candidate library
by performing $R_m$ iterations of expansion, scoring and pruning.
 See \cref{alg:mda} for details.

\paragraph{Shrinking the hypothesis space.}
If the highest model weight is sufficiently high, and if its error is sufficiently low,
then we can optionally shrink the pool down to a minimal size,
by adapting the value of $\nparticlesModel$.
This  limits computational cost and candidate proliferation.
Shrinkage is gated by confidence and fit quality, but can still discard a model
that becomes useful later. Maintaining a diverse library
of hypotheses is important, even if the posterior is very concentrated on just a single model,
since it improves the search process, similar to other evolutionary optimization algorithms
such as FunSearch \citep{romeraparedes2024funsearch}.

\paragraph{Computing the marginal likelihood (evidence).}
 To compute the posterior over models,
 we need to estimate the marginal likelihood or evidence for each model,
 $Z_m^\expt = p(\data_{0:\expt}|m) = \int p(\data_{0:\expt}|m,\theta)\, p(\theta|m)\, d\theta$,
where the likelihood  factorizes over the trajectories:
$p(\data_{0:\expt}|m,\theta)=\prod_{i \in \data_{0:\expt}}
p(y_{1:T}^i|\design^i,m,\theta)$.
We estimate $Z_m^\expt$ using tempered SMC
in \cref{app:tempered} or the BIC approximation
in \cref{app:BIC}.
The Bayesian evidence averages the likelihood over the declared parameter prior
\citep{MacKay1991}, which provides an automatic
Occam penalty that depends on both model flexibility and the parameter prior.
However, for some problems
we also add an explicit complexity penalty by using the
model prior $p(m)\propto e^{-\lambda \nparams_m}$,
where $\nparams_m$ is the number of free parameters in $m$
and $\lambda$ is tuned on a validation set.
(By default we use $\lambda=0$, corresponding to a uniform prior.)

\paragraph{Computing the likelihood.}
For latent variable models,
computing the  likelihood
of a single trajectory,
$p(y_{1:T}\mid \design, m,\theta)
= \int \prod_{t=1}^T p(y_t\mid z_t,m,\theta)\,
p(z_t\mid z_{t-1},\inputs_t,m,\theta)\, d z_{1:T}$,
may be intractable.
For this we mostly use another layer of SMC, namely the bootstrap
particle filter in \cref{app:PF},
although we also compare to the  Bayesian Synthetic Likelihood
method in \cref{app:BSL}.

\eat{
\paragraph{Special cases.}
The 3 level hierarchy does not require one universal inference backend,
and we use different methods for different problem domains.
For \ChemBench, we solve a static symbolic
regression problem using MAP parameter estimation and BIC evidence approximation;
there is no latent trajectory in this case.
For \DPbench, we learn the parameters of an ODE using tempered SMC for the parameters
and evidence; the likelihood is tractable because the latent dynamics are deterministic.
For \HHbench, we learn the parameters of an SDE using particle filtering,
 BIC weights, and a MAP prediction.
}

\paragraph{Experiment design.}
We choose the experiment whose outcome is most informative about
which hypothesis is true:
$\design^\star=\arg\max_{\design \in \designSpace} I(M;Y_\design\mid\D)$
\citep{lindley1956,Box1967,chaloner1995,rainforth2024}.
This is called the Expected Information Gain (EIG),
which is a special case of Value of Information (VoI).
In practice we use the model-disagreement surrogate in \cref{eq:voi},
which is an approximation to the mutual information.
\eat{
  \footnote{
We  also considered a more  task-driven VoI
objective,
following \citep{rainforth2024,bickfordsmith2023}.
This objective prefers  the design which maximally
distinguishes the models in terms of their predictive performance,
rather than trying to identify which model or parameters are the true ones.
Preliminary results showed that this approach gave similar results to EIG
on our benchmarks, so we omit the details for brevity.
  }
  }
We optimize the VoI surrogate using CMA-ES \citep{hansen2016cma}
for \ChemBench, which has 7 real-valued inputs,
and exact enumeration of finite action menus for \glucose and \HHbench.
See \cref{app:VOI} for details.

\label{sec:prediction}
\paragraph{Prediction.}
After each step, we convert the belief state,
which is a distribution over models,
$p_\expt = p(m|\data_{0:\expt})$,
to a predictive distribution in observation space,
$\rho_\expt(\design)=p(Y|\design,\data_{0:\expt})$,
which is computed by Bayes model averaging
(c.f.,  \citep{Self1987}).
From this, we derive the posterior predicted mean of the target,
which is optimal under the $\ell_2$ loss we use in \cref{eq:nmse}:
\begin{align}
  \hat{\target}_\expt(\design)
  = \E[\target(Y)|\design,\data_{0:\expt}] =
  \sum_m  \int \Big[\int \target(y)\, p(y|m,\theta,\design)\, dy\Big]\, p(m,\theta|\data_{0:\expt})\, d\theta
  \label{eq:BMA}
\end{align}
For conditional forecasting (where $t_c>0$),
the inner predictive distribution also
conditions on the query prefix in \cref{eq:conditional_target}; this can
be implemented by estimating the latent state at the forecast cut point $t_c$.
We also consider simplifications of this, where we just plugin the MAP
model $\hat{m}_\expt$ and its corresponding parameters, $\hat{\theta}_\expt$,
instead of averaging over them. For $\target(y)=y$,
the prediction becomes
 the plugin posterior predictive mean
$\hat{\target}_\expt(\design) = \E[Y\mid\hat{m}_\expt,\hat{\theta}_\expt, \design]$.
For general target functions, we can sample from the posterior predictive,
and then approximate $E[\target(Y)]$ by Monte Carlo.

\section{Experimental results}
\label{sec:results}

We organize the results by order of
increasing model complexity. \CHEMbench\ requires fitting static symbolic laws;
glucose forecasting requires learning partially observed deterministic ODEs;
and \HHbench\ requires prediction from partially observed SDEs.
In each case we evaluate held-out interventional prediction.
We plot test performance vs the total number of experiments at each step $\expt$,
which is $B_\expt=N_0+\expt$, where $N_0$ are the number of
initial experiments in $\D_0$. The final MDA budgets
are $3+22$, $2+4$, and $2+8$ in chemistry, glucose, and HH, respectively.
Additional
results, on \textsc{BoxingGym},  can be found in \cref{app:boxing}.

\paragraph{Baselines.}
We compare MDA to two main classes of baseline.
The first class are methods based on \emph{induction},
which fit a model
$(\hat{m},\hat{\theta})$ given $\D$,
and then predict using this model.
For \CHEMbench, we use the  method from their paper,
and for \glucose, we fit ARX and sparse polynomial dynamics (SINDYc).
We also consider two baseline approaches based on
\emph{transduction},
which use an LLM to directly  predict the output
$Y$ given the test input $\design$ and all the training data $\D$,
i.e., they ask the LLM to generate
$\arg\max_y p(y \mid \design, \D)$ using in-context learning (ICL).
Surprisingly,
\citep{Li2024induction} showed that the transductive approach
can outperform the inductive approach on some ARC AGI problems,
so this is not a trivial baseline.\footnote{
Induction and transduction use
compute in different ways.
For induction, most of the time is spent in numerical model fitting,
and the LLM calls (for proposing new models) are only a small
fraction of the overall time.
(The model fitting uses CPUs, which we run locally on a laptop
or sometimes on Modal; the LLM calls (which we access via OpenRouter)
use hardware accelerators of various forms, such as GPUs.)
For transduction, all of the time is spent inside the LLM.
It is therefore hard to match compute budgets.
However, our focus in this paper is on sample efficiency,
not compute effiency.
}
to 
The experiments used to create the data $\D$ can either be chosen by the LLM
itself (which we call ``ICL (+LLM design)'', c.f., \citep{Gupta2025}),
or by the MDA agent (which we call ``ICL (+MDA design)'');
this distinction isolates the effects of experiment design
from prediction.

\paragraph{LLMs.}
We use  GPT-5.6-Luna at medium reasoning effort as our LLM
for all the agents
used in the main paper
(except the LLM-free ARX baseline for \glucose).
However, some older results in the appendix
(in particular, the \boxing results in \cref{app:boxing}),
used Claude-Opus-4.7.
We have not systematically studied the effects of LLM choice
due to cost, and since this is orthogonal to the concerns of this paper.
None of the agents have internet access, or the ability
to read source code or data files.
\eat{
The MDA agent just use the LLM to propose models, given a context;
everything else
(designing or executing an experiment,
or fitting or predicting with a model)
is done via function calls as part of the harness
(see \cref{alg:api} for the API).
The \SciLab agent for \CHEMbench has a similar
harness, but differs in the details
(e.g., it uses PySR to perform symbolic regression).
The ICL baselines are pure LLM-based approaches that don't
build explicit models and don't use tools.
}
Prompts are in \cref{app:prompts}.

\subsection{\CHEMbench: discovering enzyme-kinetic rate laws}
\label{sec:chem}

\paragraph{Benchmark.}
\CHEMbench is derived from \citep{kabra2026autoscilab}.
The task is to learn a
static algebraic law mapping seven controllable inputs---substrate, inhibitor, second-substrate and product
concentrations, enzyme loading, temperature, and pH---to the initial rate of a chemical
reaction:
\[
 r=f(C_A,C_I,C_B,C_P,E,T,\mathrm{pH};\theta).
\]
The complete benchmark contains 57 worlds: nine canonical mechanisms and 48 compound mechanisms.
We use the medium difficulty level for all worlds.
We evaluate held-out prediction error
using RMSLE,
as defined in \cref{eq:rmsle},
and its normalized nMSLE version,
as defined in \cref{eq:chemnmse}.
An agent is said to ``pass'' a given world if
$\Ind[\mathrm{RMSLE}<0.05]$.
See \cref{app:chem} for further details.

\paragraph{Methods.}
MDA is initialized with three scalar rate observations in $\D_0$,
which are created by 
varying $C_A\in\{0.03,0.3,3\}$ and holding the other terms constant.
It then actively selects $\Nexpt=22$ additional experiments.
We initialize the hypothesis space with nine canonical mechanisms plus
up to 12 valid, nonduplicate LLM proposals; later triggered calls request
six candidates, with a live-pool cap of 24.
We  fit each model by bounded,
multistart nonlinear least squares,
and approximate the marginal likelihood using the BIC approximation
in \cref{eq:BIC}.
We approximate the model posterior using
$p(m\mid\D)\propto e^{-\mathrm{BIC}_m/2}$.
The predictor uses the highest-weight model and its
point estimate $\hat\theta_m$.
Experiment design
uses the derivative-free CMA-ES algorithm of \citep{hansen2016cma}
to  maximize the weighted between-model disagreement in
\cref{eq:chem-model-disagreement},
which is an approximation to EIG.
(See \cref{fig:chem-controls} for some evidence 
that these EIG designs outperform random designs.)
For the main baseline, we use
the  \SciLab agent from  \citep{kabra2026autoscilab},
which combines an LLM with the evolutionary symbolic
regression method PySR \citep{cranmer2023pysr},
independently
fitted at each budget $B \in \{5,10,15,20,25\}$.
For the ICL+MDA baseline, we reuse MDA's data,
and for ICL+LLM, we start with $\D_0$ but then use the LLM
to choose the next 22 experiments.

 \eat{
 \paragraph{Baseline.}
 As our baseline we use the agent from
 \citep{kabra2026autoscilab}, which they call
 \SciLab.
 This is an LLM-driven agent that maintains competing hypotheses and proposes batches of bounded
experimental regions. Within each region, \SciLab fits the free constants of the LLM-proposed rate laws and
selects concrete conditions from numerical candidates using between-hypothesis predictive disagreement, with
Gaussian-process uncertainty as a secondary criterion. Its built-in deterministic warm sweeps override the
LLM-proposed regions for the sole batch at $B=5$, both batches at $B=10$, the first two batches at $B=15$, and
the first three batches at $B=20$, although the LLM's candidate laws still inform within-region selection once
data are available. After data collection, \textsc{PySR} \citep{cranmer2023pysr} performs evolutionary symbolic regression---discovering
functional form as well as numerical constants---and its candidates are considered alongside directly fitted
LLM-proposed forms when constructing the final submission.
 }

\begin{figure}[t]
\centering
\begin{subfigure}[t]{0.45\textwidth}\centering
  \includegraphics[width=\textwidth]{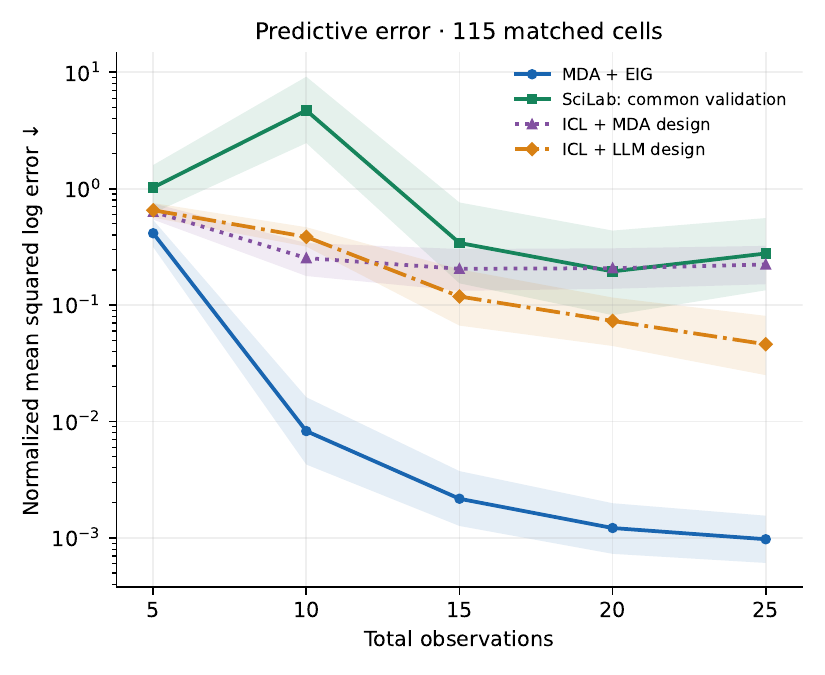}
  \caption{Held-out predictive error.}\label{fig:mainchem-curve}
\end{subfigure}\hfill
\begin{subfigure}[t]{0.45\textwidth}\centering
  \includegraphics[width=1.0\textwidth]{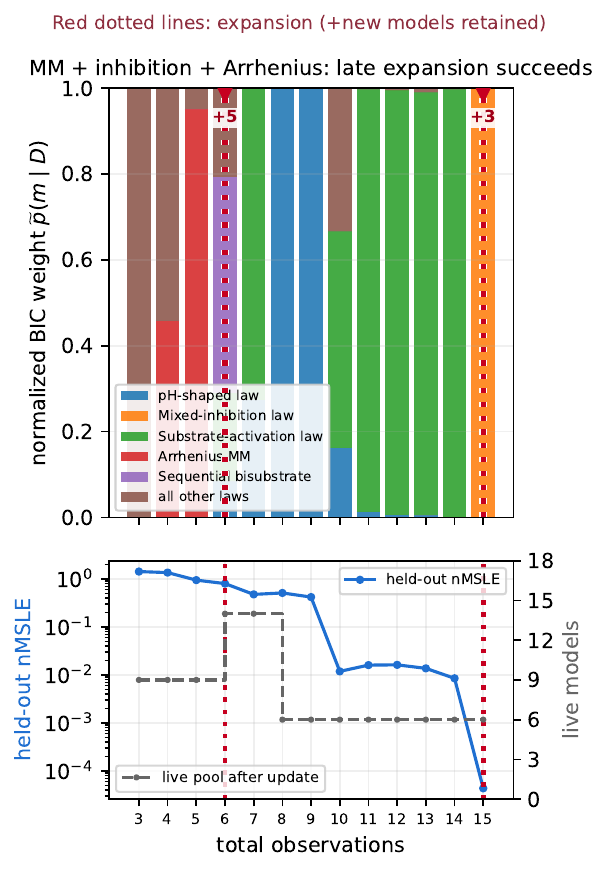}
  \caption{One $\mathcal M$-open discovery trajectory.}\label{fig:mainchem-storyboard}
\end{subfigure}
\caption{\textbf{\CHEMbench: data efficiency and model discovery.}
(\subref{fig:mainchem-curve}) Held-out predictive nMSLE.
  Errors are averaged arithmetically over seeds within world, then geometrically
  over worlds; bands are 95\% world-bootstrap intervals. All methods use the same
  115 world--seed cells spanning 54 worlds where the common-validation \SciLab
  variant is evaluable at every budget (of 171 attempted; \cref{app:chemBaseline}).
(\subref{fig:mainchem-storyboard})
A representative MM $+$ competitive-inhibition $+$ Arrhenius world.
Stacked bars show normalized BIC model weights,
colors track the five laws attaining the most posterior mass (whether seeded or proposed),
red dotted lines mark $\mathcal M$-open expansion
(labels count new models retained after pruning, not net pool growth). The lower panel shows held-out nMSLE
and live-pool size after each update, including adaptive shrinkage at step $B=8$,
when the pool reduced from 14 to 6,
and at $B=15$, when 3 new laws
replace 3 old ones, leaving 6 in total.
}
\label{fig:mainchem}
\end{figure}

\paragraph{Results.}
\Cref{fig:mainchem-curve} shows that MDA is substantially more sample
efficient than the other methods.
(The small error spike at $B=10$ for \SciLab is explained in \cref{app:chemBaseline}.)
\Cref{fig:mainchem-storyboard} visualizes the posterior over models
over time created by MDA on a specific example,
along with the corresponding test error.
At steps 6 and 15, an inadequate fit triggers an expansion
of the hypothesis space;
the second such expansion adds the orange model,
and its selection sharply reduces error.
\eat{
(Note that at $B=15$, three new laws
replace three old ones, leaving six in total;
at $B=8$, adaptive score-based pruning
reduced the pool from 14 to 6.)
}
Illustrative laws from MDA and \SciLab  are shown in
\cref{tab:chemlaws}.

\begin{table}[t]
\centering\scriptsize
\setlength{\tabcolsep}{2.2pt}
\renewcommand{\arraystretch}{0.90}
\resizebox{\textwidth}{!}{%
\begin{tabular}{@{}l p{0.255\textwidth}p{0.355\textwidth}p{0.285\textwidth}@{}}
\toprule
& \multicolumn{1}{c}{\textbf{Michaelis--Menten}}
& \multicolumn{1}{c}{\textbf{MM $+$ competitive inhibition $+$ Arrhenius}}
& \multicolumn{1}{c}{\textbf{Fractal kinetics}} \\
\midrule
\textbf{Truth}
& $\displaystyle \frac{kEC_A}{K_m+C_A}$
& $\displaystyle \frac{kEC_A e^{E_a(1/310-1/T)}}{K_m(1+C_I/K_i)+C_A}$
& $kEC_A^\alpha$ \\ \hline
\addlinespace[1pt]
\textbf{MDA}
& $\displaystyle \frac{kEC_A}{K_m+C_A}$ \quad (.002)
& $f^\star s_B(C_B)q_P(C_P)q_{\rm pH}(\mathrm{pH})$ \quad (.033)
& $\displaystyle \frac{kEC_A^n}{K_h^n+C_A^n}$ \quad \textcolor{red}{(.115)} \\ \hline
\addlinespace[1pt]
\textbf{\SciLab}
& $\displaystyle \frac{5.114EC_A}{.102+C_A}$ \quad (.011)
& \(\begin{aligned}
  &1.49\!\times\!10^{-69}C_A^{.673}C_I^{-.481}C_P^{-.007}\\[-2pt]
  &\qquad E^{.065}T^{25.804}\mathrm{pH}^{4.059}\quad\textcolor{red}{(.750)}
\end{aligned}\)
& \(\begin{aligned}
  &1.90C_A^{.353}C_I^{.002}C_B^{-.001}C_P^{-.002}\\[-2pt]
  &\qquad E^{.998}T^{.275}\mathrm{pH}^{-.055}\quad(.018)
\end{aligned}\) \\
\bottomrule
\end{tabular}
}
\caption{\textbf{Illustrative \ChemBench laws at $B=15$.}
  Parentheses: RMSLE (red exceeds the $0.05$ pass threshold).
  $f^\star$ denotes the true law above; $s_B,q_P,q_{\rm pH}$ are extra fitted factors.}
\label{tab:chemlaws}
\end{table}

\subsection{\glucose: discovering latent ODE models}
\label{sec:glucose}

\paragraph{Benchmark.}
\glucose modifies \texttt{simglucose} \citep{xie2018simglucose}, a Python
implementation of the UVA/Padova type-1 diabetes model \citep{kovatchev2009insilico}.
The original simulator was FDA-accepted for preclinical insulin-control testing
in place of animal trials \citep{Cobelli2023}.
Its 30 virtual patients (ten per age group) have hidden 13-dimensional nonlinear
ODE states. Agents observe noisy glucose every five minutes and insulin every
15 minutes over six hours; simulator variables and parameters remain hidden.
After two initial episodes, agents select four more from 48 meal amount/time
and bolus amount/delay templates. We evaluate twelve unseen interventions for
the \emph{same patient}, conditioned on a 45-minute prefix; other cuts and
details appear in \cref{app:glucose}.

\paragraph{Methods.}
MDA starts with four compartment ODEs and can add mechanisms and latent states
(up to six per model). It fits parameters numerically, weights models using
BIC (\cref{eq:BIC}), and selects experiments with noise-normalized disagreement
(\cref{eq:glucose-model-disagreement}). An EKF conditions states on the query
prefix before BIC-weighted prediction (\cref{eq:glucose_full_predictive}).
Baselines are the two ICL methods, an ARX model
for each observation channel (\cref{app:ARX}),
and SINDYc sparse polynomial dynamics \citep{brunton2016sindy,brunton2016sindyc}.
ARX and SINDYc are fitted to MDA's data,
since they cannot desgin their own experiments.
Each patient is fitted separately (a single shared model is future work).

\begin{figure}[t]
\centering
\begin{subfigure}{0.48\textwidth}\centering
 \includegraphics[width=1.0\linewidth]{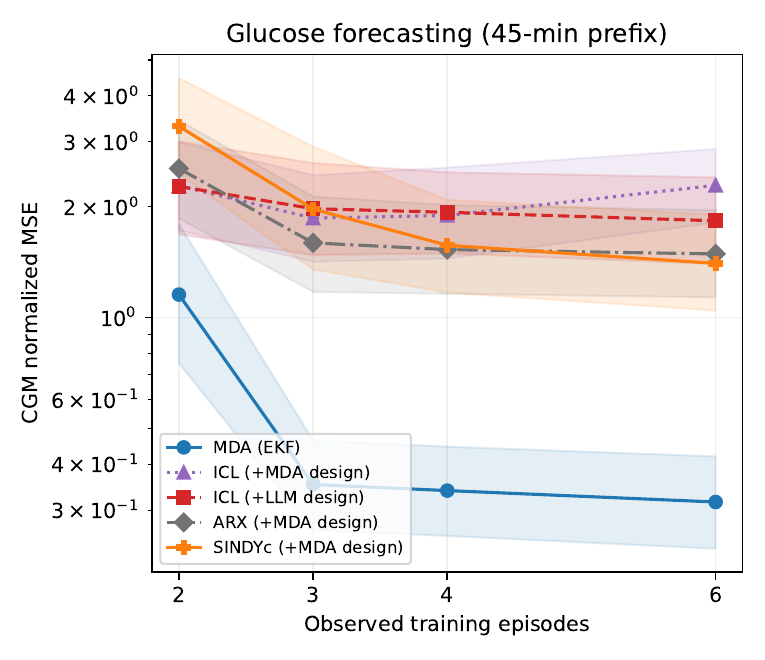}
 \caption{Held-out CGM prediction.}\label{fig:glucose_main_curve}
\end{subfigure}\hfill
\begin{subfigure}{0.50\textwidth}\centering
 \includegraphics[width=1.0\linewidth]{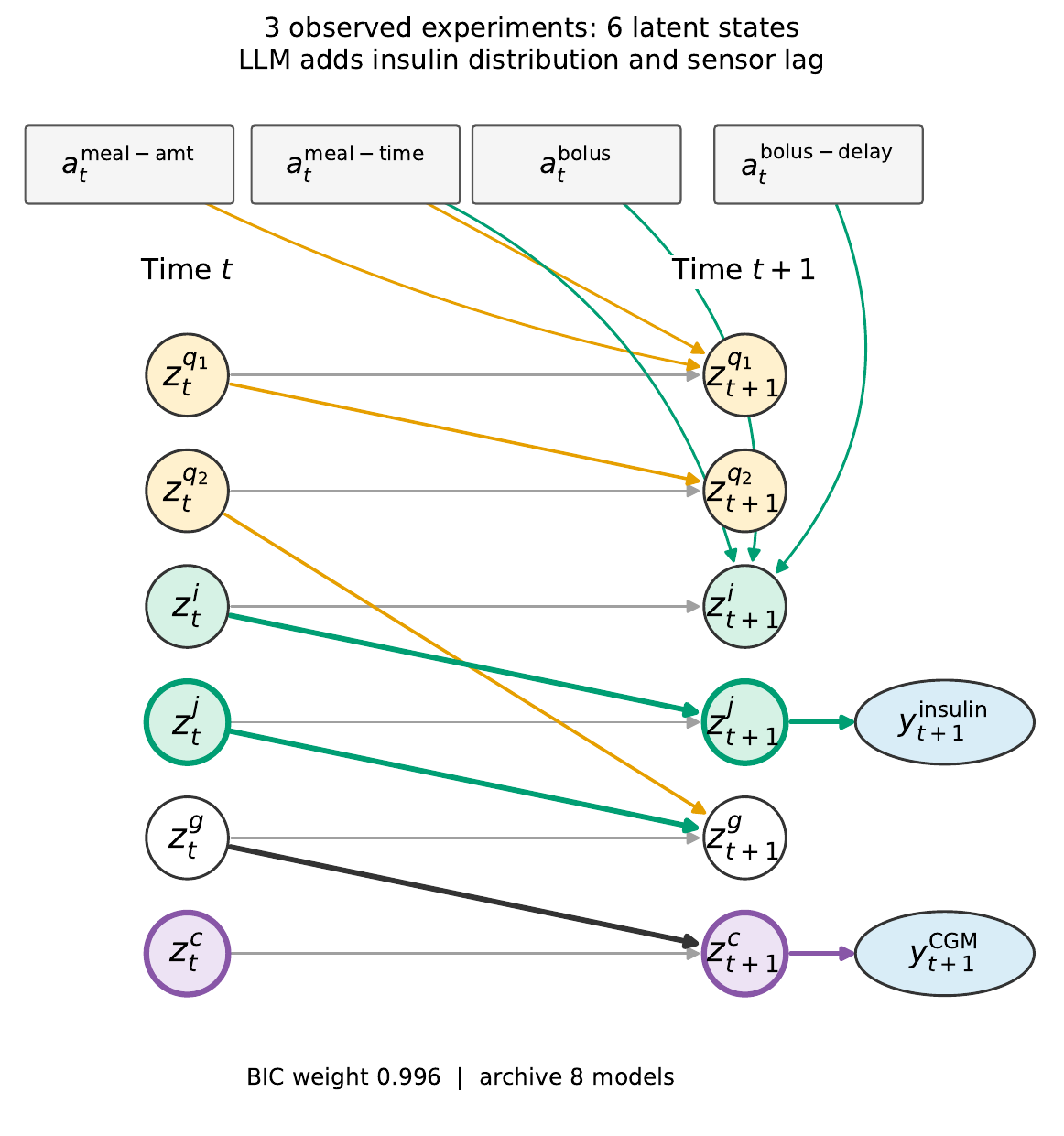}
 \caption{A learned sparse SSM.}\label{fig:glucose_main_model}
\end{subfigure}
\caption{\textbf{\glucose: prediction and learned latent structure.}
Left: initial-training-variance-normalized CGM MSE after a 45-minute prefix,
geometric means across all 30
patients with 95\% patient-bootstrap intervals; GPT-5.6 Luna (medium).
Right: a six-state model proposed and selected after three episodes.
Bold nodes/edges mark added insulin-memory and sensor-lag states.
}
\label{fig:glucose_main}
\end{figure}

\paragraph{Results.}
\Cref{fig:glucose_main}(a) shows that MDA is much more sample
efficient than the other methods in terms of predicting future GCM
given a prefix of $t_c=45$ minutes
 (see \cref{fig:glucose_multicut} for the effect of changing the prefix length).
\Cref{fig:glucose_main}(b) illustrates a 6-state model learned
by MDA after 3 trajectories. It shows the sparse latent connectivity and
added memory nodes, which retain the effects of transient inputs.

\subsection{\HHbench: discovering mechanisms with stochastic latent dynamics}
\label{sec:bio}
\label{sec:neuron}

\paragraph{Benchmark.}
\HHbench is a new benchmark we created based on
generalized stochastic Hodgkin--Huxley models.
Each world uses one of six
``mystery neurons'', five with nonstandard mechanisms and one with a textbook
M-current comparator. Finite populations of ion
channels introduce intrinsic gating noise, turning the hidden voltage-and-gate dynamics into an SDE.
The agent observes 20 noisy voltage samples per experiment.
The design space consists of 90 structured electrical
pulse protocols which specify the input current over time. We score
held-out predictions using a six-dimensional functional $\target(y_{1:T})$ containing spike-count,
adaptation, run-down, and sub-threshold voltage features,
defined in \cref{eq:hhstochfeatures}.
See \cref{app:HH} for further details.

\paragraph{Methods.}
A model $m$ specifies which ion-channel mechanisms are present, parameters $\theta$ include their
conductances and kinetics, and the latent path $z_{1:T}$ contains voltage and unobserved channel gates.
We consider 4 MDA variants:
stochastic dynamics (SDE) vs deterministic dynamics (ODE),
crossed with
recursive SMC estimation of the parameter posterior and evidence
(\cref{app:recursiveEvidence}) vs MAP parameter estimation
plus the BIC evidence approximation (\cref{app:BIC}).
The ODE arms use the Gaussian likelihood in \cref{eq:likelihoodDet},
and
the SDE arms estimate the likelihood with a bootstrap particle filter (\cref{app:PF});
the likelihood is then used as input to SMC parameter posterior estimation
or parameter point estimation.
We also compare to the two ICL baselines.

\eat{
Unlike the deterministic ODE setting, $z_{1:T}$ is not fixed by $(m,\theta)$, and the nonlinear SDE has
no tractable path integral. MDA uses a bootstrap particle filter for the
likelihood (\cref{app:PF});
an initial tempered-SMC fit
(\cref{app:tempered}) followed by recursive parameter-weight updates
(\cref{app:recursiveEvidence}) for the evidence;
and an evidence-weighted archive initialized by the LLM from two trajectories.
The initial archive need not contain the true mechanism.
To isolate inference,
we hold it fixed and replay an SDE--EIG parent's experimental history across
all four model-based arms.
%
As baselines, we compare to the two ICL variants.
We cross two choices: deterministic ODE versus stochastic SDE dynamics,
and recursive SMC versus point fitting with BIC evidence.
The ODE arms use the Gaussian likelihood in \cref{eq:likelihoodDet};
the SDE arms estimate the likelihood with a PF.
BIC arms optimize their respective likelihood directly, without SMC.
}

\begin{figure}[t]
\centering
\begin{subfigure}[t]{0.49\textwidth}\centering
  \includegraphics[width=\textwidth]{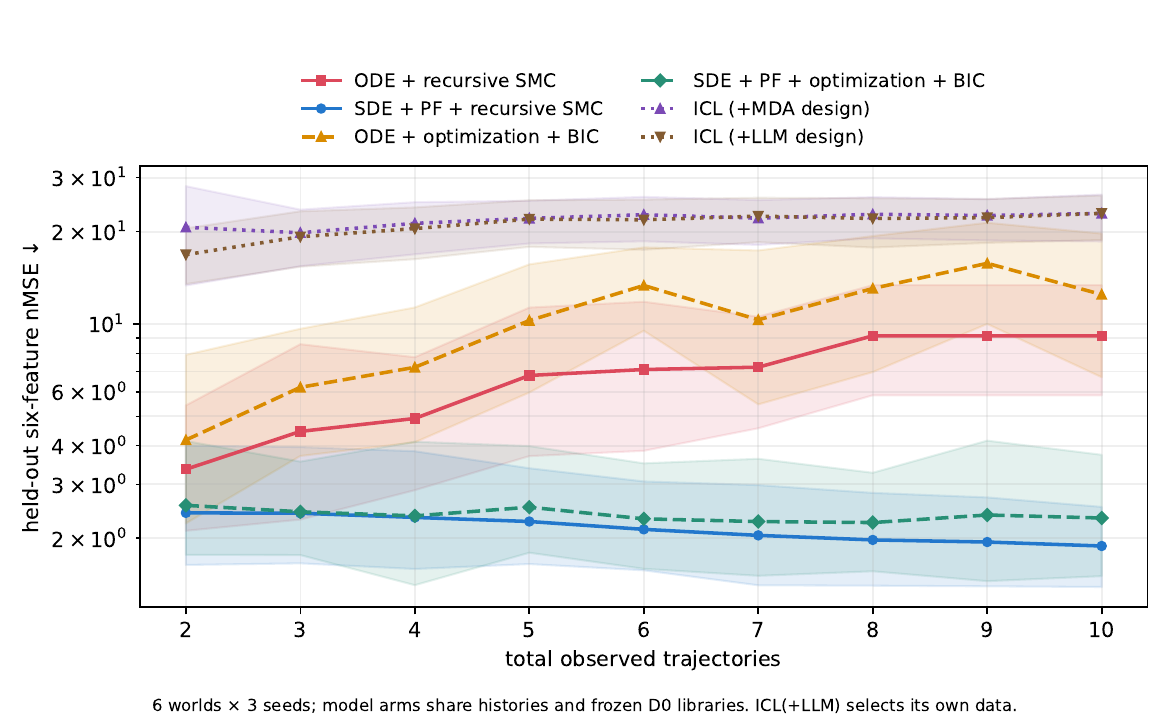}
  \caption{Controlled inference comparison.}\label{fig:hhmopenmain}\label{fig:hhmopenladder}
\end{subfigure}\hfill
\begin{subfigure}[t]{0.49\textwidth}\centering
  \includegraphics[width=\textwidth]{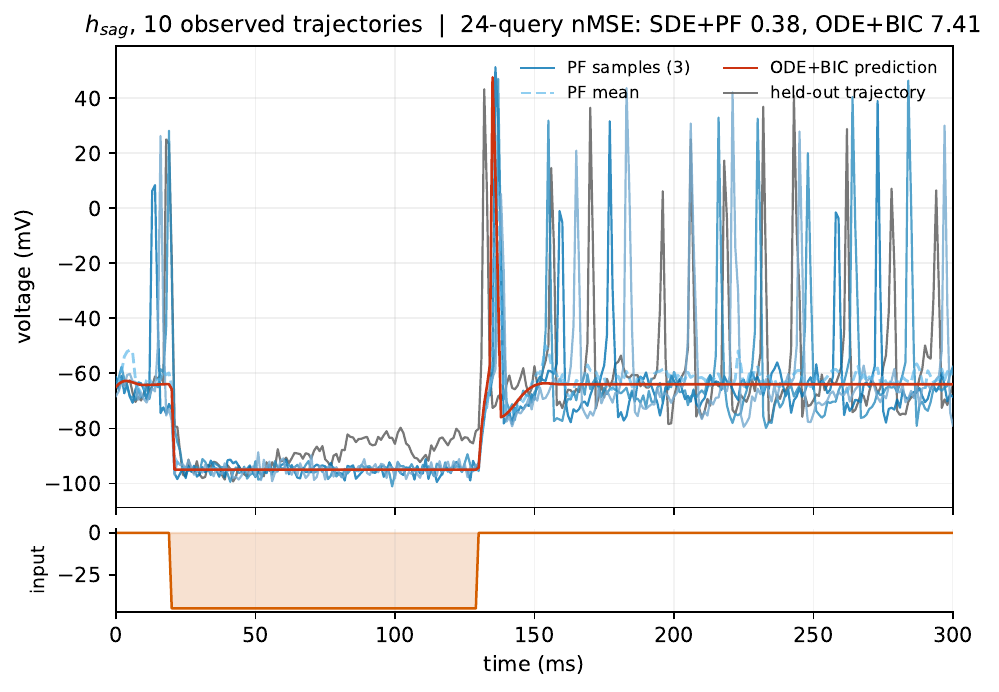}
  \caption{Posterior-predictive trajectories.}\label{fig:hhpredoverlaymain}
\end{subfigure}
\caption{\textbf{\HHbench: stochastic inference and its predictive consequences.}
  (\subref{fig:hhmopenmain}) Held-out six-feature nMSE for
  four MDA variants (ODE/SDE $\times$ recursive SMC/optimization+BIC)
  and two ICL baselines. To isolate inference, all four MDA variants use the
  same LLM-generated  model archive and experimental history collected
  by the SDE--SMC parent.
  ICL (+MDA design) uses the same history; ICL (+LLM design) chooses its own actions.
  All curves use the same six worlds and three seeds; lines are geometric
  means and shading gives pointwise 95\% hierarchical-bootstrap intervals.
  SMC uses 128 parameter particles; SDE likelihoods use 128 PF particles.
  Predictions sample the full model--parameter mixture, without top-component truncation.
  (\subref{fig:hhpredoverlaymain})
  An illustrative trajectory from the $I_h$-sag world
  for a particular input stimulus, along with predictions
after observing 10 trajectories: PF sample paths (blue), PF mean
(dashed), ODE+BIC prediction (red), held-out trajectory (gray),
and injected current below. The title reports aggregate error across 24 queries,
not the error of this selected illustrative trajectory.
}
\label{fig:hhmainresults}
\end{figure}

\paragraph{Results.}
\Cref{fig:hhmopenmain} shows that MDA with stochastic dynamics
(and either SMC or BIC estimate for marginal likelihood)
works much better than deterministic ODE approximations,
and both work better than the two ICL baselines.
\eat{
\Cref{fig:hhmopenmain} shows that stochastic dynamics improve prediction
under both inference stacks, and both SDE arms outperform the ICL baselines.
The SDE advantage therefore does not hinge on comparing SMC against BIC.
SDE+SMC improves modestly as data accumulate; its difference from SDE+BIC
is statistically unresolved in this cohort.
Recursive ODE-SMC suffers parameter-particle collapse, but directly optimized
ODE+BIC also performs poorly. This comparison tests inference on shared data,
not the benefit of later model expansion or independently selected designs.
%
}
\Cref{fig:hhpredoverlaymain} shows that  stochastic sample
paths retain spiking variability that the fitted ODE misses.
These predictions do not exactly match the precise spike times of the single
sample ground truth rollout, but they suffice to get low nMSE
score of 0.38 for the feature summary (aggregated over 24 queries)
for the SDE+SMC method, compared to 7.41 for the ODE+BIC method.

\section{Related work\protect\footnote{For more related work, see \cref{app:related}.}}
\label{sec:related}
\vspace{-0.7\baselineskip}

\begin{itemize}[leftmargin=1.1em,itemsep=0pt,topsep=0pt,parsep=0pt]
\item \textbf{Causal models for interventional prediction.}
Predicting ``what if'' questions using causal models is discussed at
length in \citet{pearl2009}.  Recently \citet{richens2024robust}
established a theoretical connection between robust prediction over a rich intervention
class and causal models; however, our finite held-out tests
do not establish causal identification.
There is also much recent work on causality and LLMs
(see, e.g., \citealp{kiciman2024,ban2025}).
We choose to use an LLM to
explicitly represent causal mechanisms,
rather than performing implicit reasoning,
since a key goal is to discover an interpretable model.
Our model-based pipelines outperform the tested ICL forecasters, but also
use numerical fitting and simulation; this is not a compute-matched
isolation of induction versus transduction.

\item \textbf{Bayesian experimental design.}
Choosing the most informative experiment is the classical model-discrimination objective of
\citep{lindley1956}, reviewed in \citep{chaloner1995,rainforth2024,Huan2024}.
Recently \citep{bedllm} proposed to combine BED with LLMs,
but our approach is very different, since we use the LLM
to propose an explicit probabilistic model,
whereas they perform all probability calculations implicitly
using the LLM itself.

\eat{
\item \textbf{Simulation-based inference.}
When the likelihood is intractable, SBI learns it or the posterior from simulations
\citep{cranmer2020frontier}, and learned/embedding summary statistics are a whole subfield of their own
\citep{fearnhead2012semiauto,chen2021neural,radev2020bayesflow,Deistler2025}.
}

\item \textbf{LLMs for scientific discovery.}
  LLMs have been used to fit scientific models to static datasets,
  often augmented with literature review,
using blackbox optimization
\citep{romeraparedes2024funsearch,wahl2026modelsmc,Kasenberg2026,ERA,Gottweis2026,Xie2026false,song2025sde}.
In addition there is some work on
actively collecting datasets for model fitting using agent-designed experiments
\citep{piriyakulkij2024,Huang2025,abhyankar2026llmaces,Elteto2026,Prystawski2026,
  Jagadish2026,Bisht2026,Fu2026,Ghareeb2026}.
MDA builds on SMC-S's pool-conditioned symbolic proposals \citep{piriyakulkij2024}
and shares mechanistic population scoring with ModelSMC \citep{wahl2026modelsmc}.
ATLAS \citep{Elteto2026} similarly uses active discrimination among RNN
hypotheses to recover agents in bandit tasks.
Our emphasis is a common executable-model interface across algebraic, ODE, and
stochastic latent-dynamics tasks, with controlled inference and acquisition
comparisons (see \cref{app:related} for SINDy and other numerical approaches).

\item \textbf{Benchmarks for interactive scientific discovery.}
  Various benchmarks evaluate agents that learn scientific laws by \emph{interactive experimentation}:
  we build
on \ActiveChem \citep{kabra2026autoscilab},
and \textsc{BoxingGym} \citep{gandhi2025boxinggym},
and add a glucose study and our own \HHbench.
Other relevant benchmarks include \DP\ \citep{wiemann2026discoverphysics},
\textsc{NewtonBench} \citep{zheng2026newtonbench},
\textsc{Science-Gym} \citep{cerrato2026sciencegym}, and
\textsc{SciGym} \citep{duan2025scigym}.
\end{itemize}

\eat{
which also wraps an LLM
structure-proposer in SMC; we differ in conditioning proposals on the
whole pool's residuals (as in SMC-S \citep{piriyakulkij2024}), in
adding VoI-based experiment \emph{design}, and centrally in the
$\mathcal M$-open \emph{expansion} to novel latents with
identification by design. Concurrently, LLM-ACES
\citep{abhyankar2026llmaces} also closes an
LLM-proposer/experiment-selection loop for dynamical-system discovery,
but drives its search by LLM-guided \emph{symbolic regression} (PySR)
and point fit rather than a two-level Bayesian evidence posterior with
an explicit VoI design objective. Their free-form symbolic-regression
search is well suited to their ODEBench suite (63 arbitrary Strogatz
ODEs), where the target is an unrestricted algebraic expression; MDA
is instead aimed at \emph{structured mechanistic} discovery, where the
hypothesis space is a constrained set of physically meaningful
mechanisms --- channels, force laws, rate laws --- that an LLM can
propose and two-level Bayesian evidence can discriminate under a tight
experiment budget.
}

\section{Discussion}
\label{sec:discussion}

MDA combines LLM-proposed executable libraries with approximate Bayesian inference
and experimental design,
to learn models with predictively useful abstractions.
In the future we would like to improve inference efficiency, using methods such as
amortized SBI
\citep{cranmer2020frontier,radev2020bayesflow,Deistler2025},
as well as tackle harder benchmarks by giving the LLM more tools for
analysing data (rather than putting everything in the prompt)
and for proposing a library of reusable model mechanisms.

\eat{
MDA combines LLM-proposed executable libraries with numerical inference
and experimental design. The evidence favors stochastic dynamics for neural
prediction, but does not establish that full Bayesian integration is always
better than optimization and BIC. Candidate coverage, approximate model weights,
and finite benchmark families limit the conclusions. In particular, the neural
headline initializes models from data but does not test later model births.
Full-history refitting can be expensive; recursive updates reduce this cost
but risk particle impoverishment. Future work should isolate the benefit of
online expansion and improve inference efficiency, including amortized SBI
\citep{cranmer2020frontier,radev2020bayesflow,Deistler2025}.
}

\clearpage

\section{AI use statement}

In this work, we used generative AI tools to help write the code,
launch and analyse experiments, and help with writing and figure preparation.

\clearpage

\bibliographystyle{arxiv_mda}
\bibliography{refs}
\clearpage

\appendix
\clearpage
\section{Method: further details}
\label{app:method}

\subsection{The agent-environment interface}
\label{app:api}

\Cref{alg:api} sketches the agent/environment interface at a high
level.\footnote{
Our API is loosely based on \textsc{BoxingGym}'s Fig.~2
\citep{gandhi2025boxinggym}.
However,
we make several changes:
(i) we evaluate after every step, to get a learning curve,
rather than a single number;
(ii) we add an update-state method to each agent, since MDA
sequentially updates its belief state.
}

\begin{algorithm}[!ht]
  \caption{\textbf{The  agent/environment interface}.
    }
\label{alg:api}
\begin{algorithmic}[1]
  \State $\text{env}\gets\textsc{Env.init}()$
   \State $C\gets\text{env.\textsc{describe}()}$
   \State $\D_0 \gets \text{env.\textsc{init-data}()}$
   \State $\target \gets \text{env.\textsc{target}()}$ \Comment{the evaluation functional}
  \State $\text{agent} \gets \textsc{Agent.init}(\textsc{llm}, C, \D_0, \target)$
  \State $\ell_{\mathrm{pred}}(0)\gets\text{env.\textsc{eval-predictor}}(\text{agent.\textsc{get-predictor}}())$ \Comment{hold-out predictive loss}
\For{$\expt=1:\Nexpt$}
  \State $\design_\expt\gets\text{agent.\textsc{design}}()$
  \State $\text{obs}_\expt\gets\text{env.\textsc{step}}(\design_\expt)$
  \State $\text{agent.\textsc{update-state}}(\design_\expt,\text{obs}_\expt)$
  \State $\ell_{\mathrm{pred}}(\expt)\gets\text{env.\textsc{eval-predictor}}(\text{agent.\textsc{get-predictor}}())$ \Comment{\cref{alg:mda}}
\EndFor
\Statex
\Function{\textsc{eval-predictor}}{$\hat\target$} \Comment{score a predictor on the held-out query set $\mathcal Q$}
  \State $\hat f_q\gets\hat\target(\design_q,\mathcal H_{t_{c,q}})$ for each held-out query $q$
  \State \Return normalized loss against independent references $f_q^{\rm ref}$ \Comment{\cref{eq:nmsegeneral}}
\EndFunction
\end{algorithmic}
\end{algorithm}

\subsection{Evaluation}
\label{app:eval}

Let $\hat P_q$ be the agent's predictive distribution for query $q$, conditional
on $\design_q$, its observed prefix $\mathcal H_{t_{c,q}}$, training data
$\D$, and context $C$. The squared-error point prediction is
\begin{equation}
 \hat f_q=\mathbb E_{\hat P_q}
       [\target(Y_{t_{c,q}+1:T})].
 \label{eq:forecast_mean}
\end{equation}
For nonlinear features, this is $\mathbb E[\target(Y)]$, not
$\target(\mathbb E[Y])$. Write its entries as $\hat f_{qji}$, where
$i=1,\ldots,d_F$ indexes feature components, or observed channels for
trajectory targets (then $d_F=d_y$); $j$ indexes future times and is a
singleton for a scalar or fixed-dimensional feature vector.
In general we use the following
normalized mean squared error metric:
\begin{equation}
 L_{\rm nMSE}=\frac1{d_F}\sum_{i=1}^{d_F}
 \frac{\sum_{q,j}o_{qji}(\hat f_{qji}-f^{\rm ref}_{qji})^2}{N_i s_i^2},
 \qquad N_i=\sum_{q,j}o_{qji}.
 \label{eq:nmsegeneral}
\end{equation}
Here $q$ indexes held-out queries (novel experiments/designs),
$i$ observation channels or feature dimensions,
$j$ future time points (a singleton for scalar or feature targets), $o$ is
an observation mask (all ones when fully observed), with $N_i>0$
for each scored component,
and $s_i$ is a scaling factor that is benchmark specific.
(Chemistry specifies $s_i$ in \cref{eq:chemnmse};
Glucose specifies $s_i$ in
\cref{eq:glucose_nmse}; HH specifies $s_i$ in \cref{eq:hhFixedScaleMSE}.)
$\hat{f}_{qji}$ is the agent's prediction,
and $f_{qji}^{\rm ref}$ is a reference,
corresponding to an observed outcome or an independently
estimated oracle mean, as specified by the benchmark.
For a fully observed fixed-dimensional feature vector, this becomes
\begin{equation}
 L_{\rm features}=\frac1{d_F}\sum_{i=1}^{d_F}
 \frac{\sum_{q} (\hat f_{qi}-f^{\rm ref}_{qi})^2}{Q s_i^2}
\end{equation}
For trajectory targets, this becomes
\begin{align}
 L_{\rm traj}=\frac1{d_y}\sum_{i=1}^{d_y}
 \frac{\sum_q\sum_{t>t_{c,q}}o_{qti}(\hat y_{qti}-y_{qti})^2}
 {s_i^2\sum_q\sum_{t>t_{c,q}}o_{qti}}
  \label{eq:lossTimeAvg}
\end{align}
For univariate fully observed time series of common length $T$, with $t_c=0$
and $s_1=1$, this
simplifies further to the familiar expression
\begin{align}
  L_{\rm traj}=\frac{1}{Q T}
  \sum_{q=1}^Q \sum_{t=1}^T (\hat y_{qt}-y_{qt})^2
\end{align}

\eat{
The agent is evaluated using
\begin{align}
  \loss &= \E_{\design\sim\queryDist}\ \E_{y\sim p^\ast(\cdot\mid\design)}
  \big[\,\ell\big(\target(y),\ \predtarget(\design)\big)\,\big]
  \label{eq:predloss}
\end{align}
where the
held-out test experiments are drawn from the \emph{query distribution}
$\design\sim\queryDist$, disjoint from the agent's own designs,
$p^\ast$ is the true system,
$\target(y)$ are the features of the observable the agent must predict,
and $\predtarget(\design)$ is the agent's  prediction.
If we use squared error
this becomes
\begin{align}
  \loss = \E[(\target(y)-\predtarget(\design))^2]
\end{align}
In this case,  the optimal estimate is the posterior mean
\begin{align}
  \predtarget(\design,\mathcal H_{t_c}^{\rm obs})
  &=\E\!\big[\target(Y_{t_c+1:T})\mid\mathcal H_{t_c}^{\rm obs},\design,\D,C\big],
  \label{eq:predtarget}
\end{align}
For binary targets, where $\target(y)=y\in\{0,1\}$,
the optimal estimate is
$\predtarget(\design) = P(Y{=}1\mid\design)  =p$,
so the loss becomes the Brier score:
\begin{align}
  \loss = \E[(y-p)^2]
\end{align}
}

For probabilistic forecasts, we can go beyond evaluating the mean by scoring
the full predictive distribution $\hat P_q$, which may integrate uncertainty
over states, parameters, and model structures. The reference distribution
$P_q^\star$ is the oracle's distribution for the same target, design, and
observed prefix. CRPS scores a univariate forecast against a held-out
realization (or is averaged over independent oracle draws),
which we discuss in \cref{app:CRPS}.

\subsection{Main MDA loop}
\label{app:meta}
\label{app:mda}
\label{app:algo}

\begin{figure}[t]
\centering
  \includegraphics[width=\textwidth]{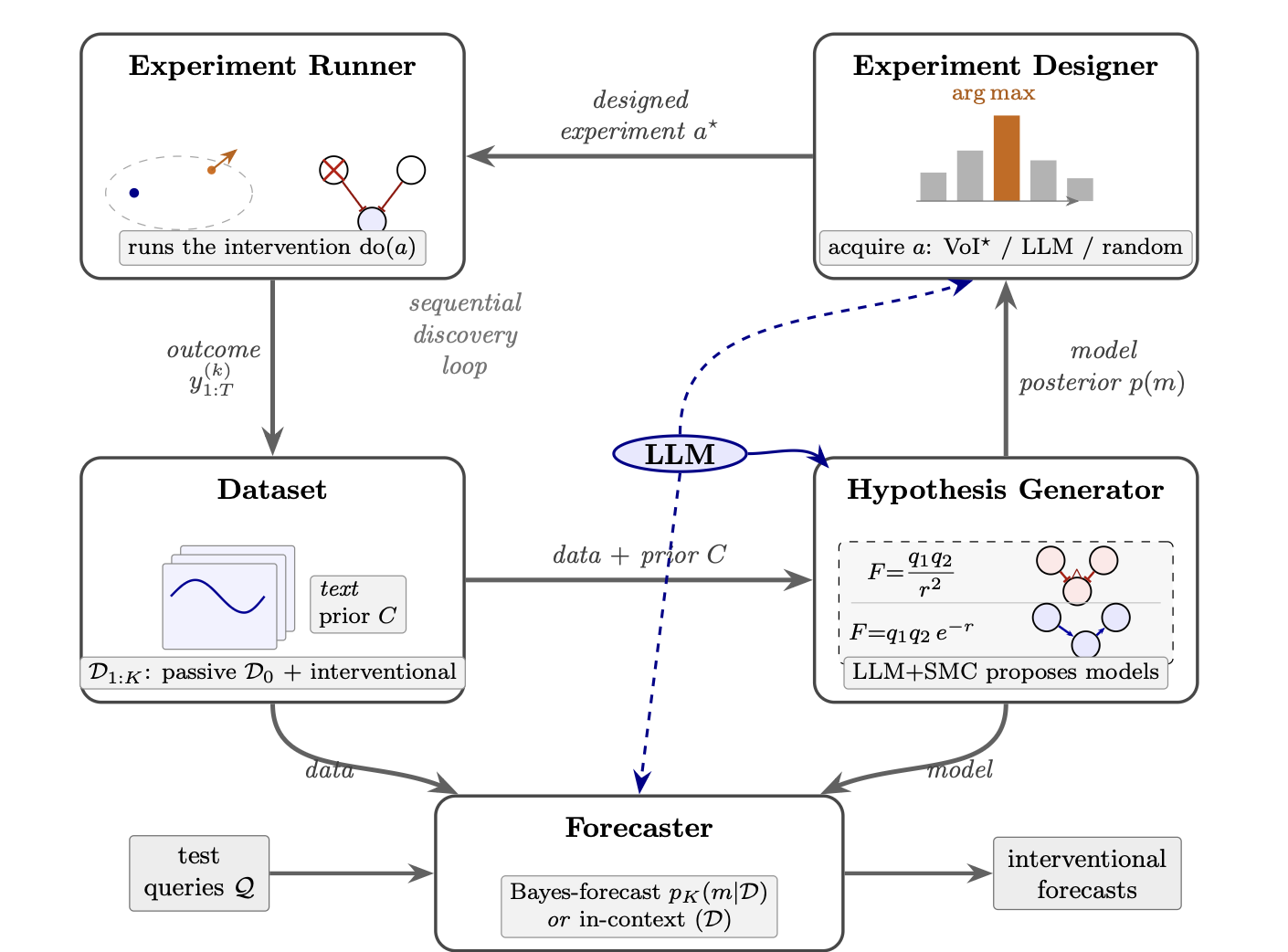}
  \caption{\textbf{The MDA discovery loop.} See text for details.
    (Figure based on \citep[Fig.2]{Elteto2026}.)}
  \label{fig:loop}
\end{figure}

\paragraph{Overview.}
The MDA algorithm is shown schematically in \cref{fig:loop}.
It is a sequential Bayesian experiment-design loop,
where we use an LLM to propose model structures;
but every numeric quantity --- the parameter posterior, the marginal evidence, the model
posterior $p(m\mid\D)$, the VoI design score, and the forecast --- is
computed by numerical fitting, inference, and acquisition routines, never by the LLM.
\eat{
In a nutshell, MDA is
a kind of  \emph{SMC sampler} \citep{delmoral2006smc}
combined with a Value-of-Information Bayesian Experiment Design loop.
(SMC is described  in \citep{Chopin2020,Naesseth2019},
and BED is described in \citep{rainforth2024}.)
}
See \cref{alg:mda} for the high level pseudocode;
we will describe the individual functions below.

\begin{algorithm}[t]
\caption{\textbf{The MDA agent (member functions).} Schematic prequential-check variant; domain-specific checks are described in the benchmark appendices. Defines the agent methods invoked by the interface of
\cref{alg:api}. The agent carries state across calls: the hypothesis space $\mathcal M$, the model posterior
$p(m\mid\D)$, the accumulated data $\D$, the observation model \textsc{loglik}, the \textsc{llm} (the
structure proposer/verbaliser),  and the pool size
$\nparticlesModel$. Config:
$\mathcal M$-open error threshold $\tau_e$, model-search rounds per step $R_m$ (the inner loop of
\textsc{update-state}; $R_m{=}1$ recovers a single gated expansion), and cumulative expansion cap
$N_e^{\max}$ (counted by $n_e$); pool cap $\nparticlesModel$ (sub-routines cross-referenced inline).
\eat{
This pseudocode specifies the full-refit variant. The headline HH SDE+PF arm
instead initializes parameter particles by tempered SMC, then recursively
updates their weights and model evidences using each new trajectory, with
ESS-triggered resampling but no default parameter rejuvenation. Its initialized
archive is frozen; this is not the optimization+BIC ablation. Recursive updates
can suffer particle impoverishment.
}
Optional \textsc{adapt-and-prune} selects
a smaller cap when confidence and fit pass the domain-specific gates, retains
the highest-scoring models, and refits if the pool changes (ChemBench:
\cref{app:chemPoolAdaptation}); it is the identity when disabled.
}
\label{alg:mda}
\begin{algorithmic}[1]
\Function{init}{$\textsc{llm},\ C,\ \D_0,\ \target$}
  \State store $\textsc{llm},\,C,\,\target$;\quad $\D \gets \D_0$;\quad $n_e \gets 0$
  \State $\mathcal M \gets \textsc{init-hyp-space}(\textsc{llm},C,\D_0)$ \Comment{initial pool (\cref{alg:initpool})}
  \State choose $\textsc{loglik}\in\{\textsc{loglik-det},\,\textsc{loglik-pf}, \textsc{loglik-bsl}\}$
  \State choose $\textsc{marglik}\in\{\textsc{marglik-smc},\,\textsc{marglik-pf}, \textsc{marglik-bic}\}$
  \State store optimizer \textsc{opt}
  \State $p(m\mid\D) \gets \textsc{model-posterior}(\D,\mathcal M;\
  \nparticlesModel,\textsc{loglik}, \textsc{marglik})$ \Comment{\cref{alg:model}}
\EndFunction
\Statex
\Function{design}{self}
  \State \Return $\textsc{opt}.\arg\max_{\design\in\designSpace}\VOI(\design;\ p(m\mid\D))$ \Comment{\cref{alg:voi}}
\EndFunction
\Statex
\Function{update-state}{self, $\design,\ \text{obs}$}
  \State $\D^-\gets\D$
  \State $e^*\gets\textsc{predictive-check}(\design,\text{obs},\D^-,p(m\mid\D^-))$ \Comment{pre-update forecast}
  \State $\D \gets \D \cup \{(\design,\text{obs})\}$ \Comment{absorb datum: $\D$ is now $\D_{0:\expt}$}
  \State $p(m\mid\D)\gets\textsc{model-posterior}(\D,\mathcal M)$
  \For{$i = 1 \textbf{ to } R_m$} \Comment{model-search rounds (ModelSMC-style)}
    \State \textbf{if} $e^*\le\tau_e \ \vee\ n_e \ge N_e^{\max}$ \textbf{then break} \Comment{stop if adequate ($n_e$ gate)}
    \State $\mathcal M \gets \textsc{expand-hyp-space}(\textsc{llm},\mathcal M,\D,C;\ N_{\new})$ \Comment{propose local mutations}
    \State $p(m\mid\D) \gets \textsc{model-posterior}(\D,\mathcal M)$ \Comment{full-batch re-judge}
    \State $n_e \mathrel{{+}{=}} 1$
    \State $e^*\gets\textsc{predictive-check}(\design,\text{obs},\D^-,p(m\mid\D))$ \Comment{refitted, in-sample adequacy}
  \EndFor
  \State $(\mathcal M,p(m\mid\D),\nparticlesModel)\gets\textsc{adapt-and-prune}(\mathcal M,p(m\mid\D),\D)$ \Comment{optional pool shrinkage}
\EndFunction
\Statex
\Function{get-model}{self}
  \State \Return $\hat m \gets \arg\max_{m\in\mathcal M}\, p(m\mid\D)$ \Comment{current-data MAP, not maximum over rounds}
\EndFunction
\Statex
\Function{get-predictor}{self, $\text{flavour}{=}\textsc{map}$} \Comment{return a predictor $\hat\target(\cdot)$}
  \If{$\text{flavour}=\textsc{map}$}
    \State \Return $\hat\target\!:(\design,\mathcal H)\mapsto\E[\target(Y)\mid\design,\mathcal H,\hat m,\hat\theta]$ \Comment{plug-in parameters; filter using $\mathcal H$}
  \Else
    \State \Return $\hat\target\!:(\design,\mathcal H)\mapsto\E[\target(Y)\mid\design,\mathcal H,\D]$ \Comment{posterior predictive, \cref{eq:conditional_target}}
  \EndIf
\EndFunction
\end{algorithmic}
\end{algorithm}

\paragraph{SSM.}
\label{app:model}

The agent assumes the unknown environment
can be represented by the state-space model defined in \cref{eq:ssm,fig:ssm}.\footnote{
  The SSM assumption is without loss of generality, since any non-Markovian model can be converted
  to Markov form, as long as the latent state space is allowed to grow.
  }
An experiment \emph{design} $\design$
is the choice of initial state $\initial$,
the policy $\pi$ (a fixed input sequence $\inputs_{1:T}$ in our open-loop experiments)
and the optional perturbations  $\perturb$.
We use the following notation:
$\theta'=\dopo(\perturb,\theta)$ are  the parameters of the system
after applying intervention $\perturb$,
$z_t$ are the hidden states,
$\inputs_t$ are the optional inputs,
and $y_t$ are the observed outputs.
For simpler models,  there might not be any latent variables
--- equivalent to $z=\text{const}$ ---
or there might not be any temporal dynamics --- equivalent to $T=1$.

\paragraph{Latent dynamics.}
For a fixed open-loop input sequence, the distribution over latent paths is given by
\begin{align}
  p(z_{1:T}|\design,\theta,m) = \prod_{t=1}^T
  p(z_t|z_{t-1},\design,\theta',m)
\end{align}
Here the formula conditions on $z_0$ when specified by $\design$;
otherwise integrate over its declared initial-state distribution.
The same convention applies to the compact likelihood algorithms below.
Query prediction additionally conditions on its observed prefix, with
$\mathcal H=\varnothing$ when $t_c=0$.
Note that
$p(z_t|z_{t-1},\inputs_t,\theta)$  may be an implicit distribution,
that we can sample from but cannot evaluate pointwise.
For some cases, the latent dynamics are deterministic,
and are given by an ODE:
\begin{align}
  p(z_t\mid\design,\theta,m,z_0) &= \delta(z_t - z_t(\design,m,\theta)) \\
  z_t(\design,m,\theta) &= \text{ODE-solve}(t, m, \theta', z_0, \inputs_{1:t})
  \label{eq:mt}
\end{align}
where $\theta'$ are the (optionally perturbed) parameters,
$z_0$ is the initial condition specified by $\design$,
and $\inputs_{1:T}$ are the inputs specified by its open-loop policy.

\paragraph{Observation model.}
We usually assume the emission distribution is
\begin{align}
  p(y_t|z_t,m,\theta) = \gauss(y_t|f_o(z_t), \sigma^2)
    \label{eq:obsModel}
\end{align}
where $f_o$ is the observation model.
For scalar latent state space, this is often the identity function;
for multivariate latent state space (e.g., as in HH),
$f_o$ often is a linear projection, which selects the relevant observable components.

\paragraph{Three-level inference.}
MDA has to deal with
3 levels of unknowns:
the latent states $z_{1:T}$ (where we may have $T=1$ for static systems),
the parameters $\theta$,
and the model structure $m \in \mathcal M$.
Bayesian inference at the top level requires that we compute
\begin{align}
  p(m \mid \D, \mathcal M)
  = \frac{p(\D \mid m) p(m)}{p(\D)}
  = \frac{p(\D \mid m) p(m)}{\sum_{m' \in \mathcal M} p(\D \mid m') p(m')}
\end{align}
This is computed by the \textsc{model-posterior} function in \cref{alg:model}.
When we add the hypothesis space expansion procedure
of \cref{alg:expand}, we get a method that is similar
to the SMC-S procedure discussed in 
\cref{app:posterior}.

To compute the marginal likelihood or
evidence $Z_m = p(\D \mid m)$, we have to marginalize
out the parameters:
\begin{align}
  p(\D \mid m) = \int p(\D \mid m, \theta) p(\theta \mid m) d\theta
  \end{align}
By default, this is computed using tempered SMC in \cref{alg:smc}.
See \cref{app:margLik} for details.

Finally, to compute the (observed data) likelihood,
we have to marginalize out the latents:
\begin{align}
  p(\D \mid m,\theta) = \int p(\D \mid z, m, \theta) p(z \mid m,
    \theta) dz
  \end{align}
By default, this is computed by bootstrap particle filtering
(a special case of SMC) in \cref{alg:pf}.
See \cref{app:likelihood} for details.

Using PF to estimate $p(\D \mid m,\theta)$ inside of an outer
SMC algorithm to estimate $p(\theta \mid \D,m)$ follows the nested architecture of SMC$^2$
\citep{chopin2013smc2}.
Our outer level enumerates and weights candidate models rather than running
another SMC sampler. The three-level hierarchy thus combines particle
inference within models with approximate finite-library selection;
LLM proposals and pruning search for useful structures.

\eat{
\paragraph{Target and approximation.}
The per-model evidence target is
\begin{equation}
  Z_m^{(\expt)}=p(\data_{0:\expt}\mid m)=
  \iint p(\data_{0:\expt}\mid m,\theta,z)\,p(z\mid m,\theta)\,p(\theta\mid m)\,dz\,d\theta,
  \label{eq:Zk}
\end{equation}
A bootstrap PF supplies a nonnegative unbiased likelihood estimate, but its
logarithm is biased and normalized finite-particle model weights are not
unbiased posterior probabilities. Pseudo-marginal exactness requires a valid
extended-state construction retaining estimator randomness and invariant
transitions \citep{andrieu2009pseudomarginal}; unbiased inner estimates alone
do not establish it. Fractional powers of a noisy likelihood estimate need
not be unbiased estimates of fractional powers of the likelihood. This does
not rule out a correct terminal extended-space target, but neither
likelihood tempering nor data tempering alone guarantees one. We treat our
finite-particle implementations and heuristic model search as approximate
inference on the discovered candidate set, not as an exact sampler over all
possible programs.
}

\paragraph{Simpler inference backends.}
If the latent variables are missing, or can be computed
deterministically,
then we don't need the bottom-most PF layer to compute the
likelihood. 
For the finite candidate pool, model weights are computed by enumeration;
there is no additional structure-level SMC layer.
We also consider other approximations,
such as replacing particle filtering for computing the likelihood
with Bayesian synthetic likelihood (see \cref{app:BSL}),
and  replacing tempered SMC for computing the marginal likelihood
with  a BIC approximation (see \cref{app:BIC}).

\paragraph{From beliefs to actions.}
Given a posterior $p(m,\theta|\data)$,
we can choose the next experiment $\design$ by maximizing the value of information
provided by $\design$, as explained in \cref{app:VOI}.

\paragraph{From beliefs to predictions.}
Given a posterior $p(m,\theta|\data)$,
we can predict the outcome of a novel experiment using
the posterior mean:
\begin{align}
  \predtarget_{\mathrm{BMA}}(\design) = \int \int \sum_m \target(Y)
  p(Y|\design,m,\theta) p(m,\theta|\data) dY d\theta
  \end{align}
We also consider a simplified prediction, where we use  a plugin estimate
with the MAP model and its parameters,
\begin{align}
  \predtarget_{\mathrm{plugin}}(\design) = \int \target(Y)   p(Y|\design,\hat{m},\hat{\theta}_m) dY 
  \end{align}
where $\hat{\theta}_m$ is the posterior mean or mode
of the parameters for model $\hat{m}$.

\paragraph{Experimental settings.}
\Cref{tab:smc} summarizes the different variants of MDA that we use
in our main experiments. Further details can be found in the following appendices.

\eat{
\subsection{SMC hyper-parameters and computational cost.}
Table~\ref{tab:smc} specifies the parameter settings
for the main experiments in this paper.
Full-history refitting
at every round has $O(\Nexpt^2\nparticlesModel)$ forward-simulation cost when
fit cost is linear in accumulated data, holding inner computation fixed.
The HH recursive update avoids rescoring all past trajectories on each update,
but risks parameter-particle impoverishment. LLM proposal calls and numerical
fitting costs should be accounted for separately; aggregate GPU-hours are
not end-to-end elapsed time.
}

\begin{table}[t]
\centering\footnotesize
\begin{tabular}{@{}p{.17\linewidth}p{.27\linewidth}p{.24\linewidth}p{.22\linewidth}@{}}
\toprule
Arm & Likelihood / parameter inference & Model scoring / prediction & Model library / experiment design \\
\midrule
Chemistry MDA & Weighted residual fit / bounded multistart optimization
 & BIC weights / MAP plug-in prediction
& 9 canonical models plus up to 12 initial proposals,
adaptive cap 6--24, up to 6 proposals per expansion / CMA-ES over 7d space
\\
Glucose MDA & ODE observation likelihood with AR(1) CGM noise / bounded optimization
 & BIC weights / model-averaged plug-in prediction
& 4 initial models, up to 2 proposals per expansion, cap 14
/ enumeration over 48 nominal templates (42 distinct)
\\
HH SDE+PF & Bootstrap PF, $N_z=128$ /
SMC tempering with $N_p=64$, then recursive updates
 & Estimated evidence / BMA prediction
& 3 model proposals plus plain Na/K/leak, online expansion / enumeration over 90 designs
\\
HH ODE+BIC & Deterministic Gaussian likelihood / bounded optimization
 & BIC weights / model-averaged deterministic predictions at fitted parameters
 & Same initial archive / enumeration over 90 designs
\\
\bottomrule
\end{tabular}
\caption{\textbf{Main experimental settings.}
  We specify the key parameter/algorithm configurations for the experiments
  in the main text. All LLMs use  GPT-5.6-Luna, medium effort
  (except in \cref{app:boxing}).
  The two ICL baselines perform direct prediction without these inference backends.
  Chemistry BIC, glucose BIC, and HH ODE+BIC use no additional parameter-count prior;
  HH SDE+PF weights additionally use
  $p(m)\propto\exp(-2\nparams_m)$.
  Experiment design maximizes a between-model disagreement surrogate
  (\cref{eq:voi}, with domain-specific scaling) using CMA-ES
  or enumeration.
  More specific details can be found in the appendix for each experimental domain.
  }
\label{tab:smc}
\end{table}

\subsection{Priors}
\label{app:structure_prior}

Candidate structures may be seeded or LLM-proposed; each
has its own free parameters $\theta$. The joint prior factorises as
\begin{align}
  p(m,\theta)=p(m)\,\prod_{j=1}^{\nparams_m}p(\theta_j\mid m),
\end{align}
with a \emph{uniform} prior on each coefficient over its
declared bounds\footnote{
When the proposer chooses bounds after seeing data, those bounds define a
data-dependent prior. Conditioning on that choice does not remove data reuse:
narrowing bounds around a fit can inflate evidence. These are approximate
predictive model weights, not calibrated Bayes factors under a prespecified prior.
},
and a \emph{structure} prior
\begin{align}
  p(m)\propto e^{-\lambda \nparams_m},
  \label{eq:occamprior}
\end{align}
where $\nparams_m$ is the number of free parameters of $m$ and $\lambda$ is a per-parameter Occam penalty (in
nats; the active per-domain settings and BIC exceptions are in \cref{tab:smc}). The posterior over structures
is then $p(m\mid\D)\propto Z_m\,e^{-\lambda\nparams_m}$, with
$Z_m=\int p(\D\mid\theta,m)\,p(\theta\mid m)\,d\theta$ the marginal likelihood computed by \cref{alg:smc}. When
$\mathcal M$-open expansion adds a new structure it enters with the same unnormalized prior $e^{-\lambda\nparams_m}$ and the
posterior is renormalized over the current finite pool; there is no separate catch-all mass, so the pool is always a
self-normalized distribution on the visited finite support.

\paragraph{Prior choice and predictive weighting.}
Uniform priors are convenient, but not noninformative: their bounds and
parameterization can substantially affect marginal likelihoods and hence model
weights \citep{llorente2022safepriors}. A stronger justification would combine
physical parameter scales with prior-predictive checks that simulated responses
are plausible across relevant interventions. Sensitivity of model weights need
not imply sensitivity of predictions when different mechanisms behave similarly;
our empirical claims concern held-out prediction rather than calibrated
probabilities of recovering the true mechanism. For this predictive objective,
an alternative is to choose ensemble weights using cross-validated or genuinely
pre-update predictive scores, rather than marginal likelihoods. Such weights
would represent predictive utility, not posterior model probabilities; the
refitted residuals used in our adequacy checks are not held-out validation scores.
A prospective safeguard is to fix parameter ranges by mechanism and physical
units before seeing outcomes, or to evaluate newly proposed bounds only on
subsequent data. Finite-bound validation alone does not prevent evidence inflation.
We leave a systematic investigation of prior choice, predictive robustness, and
predictive weighting to future work.

\eat{
This explicit complexity term is an additional finite-pool regularizer.
It does not imply that exact Bayesian evidence generally under-penalizes
flexibility; its empirical effect depends on priors, noise assumptions,
and numerical evidence error.
We found (in some preliminary experiments, not reported here)
that selecting models based on the
\emph{representation length} of their
mechanism (e.g., string length or Halstead complexity \citep{Halstead})
 worked less well, since it ignores the
 flexibility of the underlying ``elementary'' functions
 in the representation.
}

\eat{
An alternative remedy is to be Bayesian about the noise scale $\sigma$: put an inverse-gamma prior on the
residual variance and \emph{marginalize} it, giving a Student-$t$ marginal likelihood
$-(a+M/2)\log\!\big(b+\tfrac12\,\mathrm{SSE}(\theta)\big)$ over the $M$ residuals. Empirically (on \DPbench),
$\sigma$-marginalization \emph{alone} did not cure the overfitting, and can worsen model selection: with the
hundreds of residuals per trajectory, the marginal likelihood over-rewards \emph{any} reduction in SSE by a
factor ${\sim}M/2$ (a Lindley-paradox-like effect \citep{lindley1957}), so a flexible form that absorbs the
noise wins. The explicit complexity prior \cref{eq:occamprior} suffices for robust results.
}

\subsection{Computing the likelihood}
\label{app:likelihood}

All three likelihood routines take $(\D,m,\theta)$ and sum the log-likelihoods
over experiments $(\design_i,y_{i,1:T_i})\in\D$, assumed conditionally
independent given $(m,\theta)$ and their designs. Each experiment has its
own initial state and input sequence; trajectories are not concatenated.
The recursive evidence update passes a singleton dataset for each new experiment.

\begin{algorithm}[t]
\caption{\textsc{loglik-det} --- the exact log-likelihood for \emph{deterministic} latent dynamics
(\cref{eq:likelihoodDet}), summed over the dataset. Each experiment requires one noise-free rollout from its own known initial state; emissions are independent Gaussians.}
\label{alg:loglikdet}
\begin{algorithmic}[1]
\Statex \textbf{def}\ \textsc{loglik-det}$(\D,m,\theta)\to\ell$
\State $\ell\gets0$
\For{each $(\design_i,y_{i,1:T_i})\in\D$}
  \State $z_{i,1:T_i}\gets\mathrm{ODE-solve}(m,\theta,\design_i)$ \Comment{experiment-specific initialization and inputs}
  \State $\ell\gets\ell+\sum_{t=1}^{T_i}\log\gauss(y_{i,t};f_o(z_{i,t}),\sigma^2)$
\EndFor
\State \Return $\ell$
\end{algorithmic}
\end{algorithm}

We  assume the complete-data conditional likelihood
for a single experiment is given by
\begin{align}
  p(y_{1:T}|z_{1:T},\design,m,\theta) = \prod_{t=1}^T \gauss(y_t|f_o(z_t), \sigma^2)
    \label{eq:obsModelTrajectory}
\end{align}
where $f_o$ is the observation model.
Marginalizing out the latent variables gives
the observed-data likelihood:
\begin{align}
  p(y_{1:T}|\design, m,\theta) =
  \int p(y_{1:T} | z_{1:T}, \design, m, \theta)
  p(z_{1:T} | \design, m,  \theta) d z_{1:T}
  \label{eq:likelihood}
\end{align}
If the latent dynamics are deterministic,
the likelihood is tractable:
\begin{align}
  p(y_{1:T}|\design,m,\theta) = \prod_{t=1}^T
  \gauss(y_t|f_o(z_t(\design,m,\theta)), \sigma^2)
  \label{eq:likelihoodDet}
\end{align}
where $z_t(\design,m,\theta)$ is defined in \cref{eq:mt}.
See \cref{alg:loglikdet}.
By contrast, for the stochastic models,
the path integral over $z_{1:T}$ is generally intractable,
so we must use approximations, which we discuss below.

\subsubsection{Particle filtering}
\label{app:PF}
\label{app:pf}

Our default method is to use the bootstrap particle filter algorithm
shown in \cref{alg:pf}.
It propagates weighted filtering particles for $p(z_t\mid y_{1:t},m,\theta)$;
sampling posterior paths additionally requires retained ancestry or smoothing.
The displayed routine returns an
unbiased approximation of the likelihood, given by
\begin{align}
  \hat p(y_{1:T}\mid m,\theta)=  \prod_t
  \big(\tfrac1{\nparticlesLatent}\sum_{j=1}^{\nparticlesLatent} w_t^{(j)}\big)
  \end{align}
  where $w_t^j$ is the weight of particle $j$.
  For details, see e.g.  \citep{Chopin2020,Naesseth2019}.

\begin{algorithm}[t]
\caption{\textsc{loglik-pf}: sum of independently estimated trajectory log-likelihoods. Each experiment starts a fresh bootstrap filter with $\nparticlesLatent$ particles. The product of the independent PF likelihood estimates is unbiased for the dataset likelihood under the discretized model; its logarithm is not unbiased. Transitions use the experiment's inputs and may require solver substeps (Fox--Lu SDE for HH, \cref{eq:foxlu}). Cost is $O(\nparticlesLatent\sum_i T_i)$ at fixed solver resolution.}
\label{alg:pf}
\begin{algorithmic}[1]
\Statex \textbf{def}\ \textsc{loglik-pf}$(\D,m,\theta)\to\hat\ell$
\State $\hat\ell\gets0$
\For{each $(\design_i,y_{i,1:T_i})\in\D$}
  \State $z_{i,0}^j\gets z_0(\design_i)$ for $j=1,\ldots,\nparticlesLatent$ \Comment{reset for this experiment}
  \For{$t=1,\ldots,T_i$}
    \State $z_{i,t}^j\sim p(z_t\mid z_{i,t-1}^j,\design_i,m,\theta)$ for every $j$
    \State $w_{i,t}^j\gets\mathcal N(y_{i,t};f_o(z_{i,t}^j),\sigma^2)$ for every $j$
    \State $\hat\ell\gets\hat\ell+\log\big(\nparticlesLatent^{-1}\sum_j w_{i,t}^j\big)$ \Comment{evaluate in log space}
    \State systematically resample $\{z_{i,t}^j\}$ with probabilities proportional to $\{w_{i,t}^j\}$
  \EndFor
\EndFor
\State \Return $\hat\ell$
\end{algorithmic}
\end{algorithm}

\subsubsection{Bayesian synthetic likelihood}
\label{app:BSL}
\begin{algorithm}[t]
\caption{\textsc{loglik-bsl}: Gaussian synthetic log-likelihood of independent experiment summaries, not of raw trajectories. The summary map, simulation count $R$, shrinkage $\lambda$, and ridge $\varepsilon>0$ are fixed across candidate models.}
\label{alg:loglikbsl}
\begin{algorithmic}[1]
\Statex \textbf{def} \textsc{loglik-bsl}$(\D,m,\theta)\to\ell$
\State $\ell\gets0$
\For{each $(\design,y)\in\D$}
  \State simulate independent $(z^{(r)},y^{(r)})\sim p(z,y\mid\design,m,\theta)$ for $r=1,\ldots,R$
  \State $s_r\gets\summary(y^{(r)})$;\quad $\hat\mu\gets R^{-1}\sum_r s_r$
  \State $S\gets(R-1)^{-1}\sum_r(s_r-\hat\mu)(s_r-\hat\mu)^\top$
  \State $\hat\Sigma\gets(1-\lambda)S+\lambda\operatorname{diag}(S)+\varepsilon I$
  \State $\ell\gets\ell+\log\mathcal N(\summary(y);\hat\mu,\hat\Sigma)$
\EndFor
\State \Return $\ell$
\end{algorithmic}
\end{algorithm}

An alternative to PF
is to replace
the likelihood $p(y_{1:T}|m,\theta)$ with
an approximation
\begin{align}
  p_{\summary}(\summary(y_{1:T}) |m,\theta) \approx  \gauss(\summary(y_{1:T})|\mu_{m,\theta},\Sigma_{m,\theta})
    \label{eq:sphilik}
  \end{align}
where $\summary(y_{1:T})$ is some summary feature vector
derived from the trajectory,
and $\mu$ and $\Sigma$ are computed as described below.
This approximates the density of the summary, not the raw-data density.
This is useful when a per-time step observation model,
$p(y_t|z_t,m,\theta)$, is not available
(since this is a requirement of PF).
In addition, using the summary features can provide robustness
to model misspecification.

To estimate $\mu$ and $\Sigma$, we can use 
the ``Bayesian Synthetic Likelihood''  method of \citep{wood2010,Deistler2025},
which proceeds as follows.
For each candidate
$(m,\theta)$ we draw $R$ traces $(z^{(r)}, y^{(r)})\!\sim p(z_{1:T}, y_{1:T}\mid m,\theta)$,
throw away $z^r$, and then estimate the likelihood using
\begin{align}
\mu_{m,\theta} &=\tfrac1R\textstyle\sum_{r} \summary(y^{(r)}) \\
\Sigma_{m,\theta} &=\widehat{\mathrm{Cov}}_r\!\big[\summary(y^{(r)})\big]+\varepsilon I,
\end{align}
with a small ridge $\varepsilon$ for conditioning.
See \cref{alg:loglikbsl} for the pseudocode.
We can also replace the Gaussian with a normalizing flow model,
a technique called
\emph{neural likelihood estimation}  \citep{papamakarios2019snl}.\footnote{
NLE is
not to be confused with
\emph{neural posterior estimation} \citep{greenberg2019snpe},
which trains an amortized inference network to compute $p(m,\theta|y_{1:T})$.
However, this requires generating $m$ and $\theta$, both of which can be hard
(especially if the model $m$ is code).
See \citep{cranmer2020frontier,Frazier2024} for more discussion.
}

Instead of manually specifying $\summary(y_{1:T})$, we can also learn it from data.
A simple approach, known as \emph{semi-automatic ABC}
\citep{fearnhead2012semiauto,jiang2017learning},
proceeds as follows:
draw prior samples $\{ (\theta^r, m^r, z^r, y^{r}_{1:T}) \}$ from the model,
then train a neural network regression model to predict
$m_r$ and $\theta_r$ given $\summaryLearned(y_{1:T})$.
For example, we can let $\summaryLearned(y_{1:T})$ be a 1D CNN applied
to the time series following by global average pooling,
and then pass this embedding into an MLP $f_w(s)$ with two output heads,
one for classifying $m$ and one for predicting the mean of $\theta_r$:
\begin{align}
  (\hat{m},\hat{\theta})=(f^m_w(\summaryLearned(y_{1:T})),
  f^\theta_w(\summaryLearned(y_{1:T})))
 \end{align}
We then pass $\summaryLearned(y_{1:T})$, corresponding to the final layer
of the network, into the above BSL method
to compute the likelihood, which is passed to SMC
to compute the evidence $Z_m$.

\eat{
Learning the summary changes the statistical information retained and may permit a smaller embedding or a
smaller Monte Carlo sample $R$, but it does not by itself amortize likelihood evaluation: whenever $m$ or
$\theta$ changes, ordinary BSL must still simulate at that candidate and re-estimate
$(\mu_{m,\theta},\Sigma_{m,\theta})$.  Avoiding those online rollouts additionally requires an emulator of
these moments, neural likelihood estimation, neural posterior estimation, or neural ratio estimation.
We have some positive preliminary results with this approach,
but we leave detailed exploration to future work.
}

There are two problems with BSL.
(1) It  is not much faster than PF
(see e.g., \cref{fig:hhbslladder}),
since it still requires
rolling out $R$ stochastic trajectories for every candidate
$(m,\theta)$.
(2) The above approach to learning $\summaryLearned(y)$
with a neural network requires predicting $m$ and $\theta$ on the output,
which is challenging if $\theta$ has variable size
and/or $m$ is a complex structure (such as a graph or program)
and not just a discrete index.
We can avoid both problems by using
\emph{neural ratio estimation}
(see e.g., \citep{hermans2020ratio,durkan2020contrastive}),
  which trains a binary classifier (on forwards-sampled data) to approximate
the \emph{likelihood-to-evidence ratio}
\begin{equation}
  r_\phi(y,m,\theta)\;=\;\frac{d_\phi}{1-d_\phi}\;\approx\;\frac{p(y_{1:T}\mid m,\theta)}{p(y_{1:T})},
  \label{eq:nre}
\end{equation}
Once trained,
we can use $r_\phi(y,m,\theta)$ in lieu of $p(y|m,\theta)$ inside of tempered SMC,
without needing rollouts at run time, solving problem (1).
Also, this is a discriminative model
that conditions on $m$ and $\theta$ and emits a scalar,
avoiding variable-dimensional outputs. It still requires an encoding of
$(m,\theta)$ and simulation training coverage; generalization to newly
proposed structures is not automatic.
We leave exploration of this approach to future work.

\subsection{Computing the marginal likelihood}
\label{app:margLik}

To compare models with different numbers of parameters,
we need to marginalize out their parameters by computing the evidence:
\begin{equation}
  Z_m^\expt=p(\data_{0:\expt}\mid m)=\int
  p(\D_0\mid m,\theta)\left[ \prod_{j=1}^{\expt} p(y_{j,1:T} \mid \design_j,m,\theta) \right] p(\theta\mid m)\,d\theta
\label{eq:margLik}
\end{equation}
where $p(y \mid m, \theta)$ is the observed data likelihood computed
above.

\subsubsection{Tempered SMC}
\label{app:tempered}

Our default method
to compute $Z_m^\expt$ is to use tempered SMC
with an adaptive tempering schedule, as shown in \cref{alg:smc}.
This also returns the posterior over the parameters for each model:
\begin{align}
p(\theta|m,\data_{0:\expt})
\approx \sum_{i=1}^{\nparticlesParams} W_i^{\expt} \delta(\theta-\theta_i)
\end{align}
Refitting on all accumulated data at every experiment makes the total cost
$O(\Nexpt^2 T\nparticlesParams)$ per model, holding the tempering and
rejuvenation budgets fixed and assuming an $O(T)$ trajectory likelihood.
With PF likelihoods, this additionally scales with $\nparticlesLatent$
(and the number of solver substeps per observation).
The recursive alternative below avoids repeatedly processing old trajectories.

With noisy PF likelihoods, generally $\E[\widehat L^\beta]\neq L^\beta$ for $0<\beta<1$.
Unbiased PF estimates alone therefore do not justify the intermediate
tempered targets; exact pseudo-marginal inference requires an appropriate
extended-state construction \citep{andrieu2009pseudomarginal}.
We treat the implemented particle evidence as a numerical approximation.

\begin{algorithm}[t]
\caption{Adaptive-tempering SMC over $\nparticlesParams$ parameter
  particles for each model $m$. Returns particles, weights, and $\log Z_m$.
  The effective sample size is given by
  $\mathrm{ESS}=(\sum W_i)^2/\sum W_i^2$;
  $\eta$ is the target ESS fraction, $\nroundsParams$ the
number of rejuvenation moves per temperature rung, and the annealing schedule runs for at most $J_p$
rungs --- so \textsc{loglik} is called $O(\nparticlesParams \nroundsParams J_p)$ times (LLM-free, but potentially expensive). $\ell_i$ is
a \emph{log}-likelihood; \textsc{loglik}$(\D,m,\theta)$ is a \emph{plug-in} argument --- \textsc{loglik-det}
(\cref{alg:loglikdet}) for deterministic dynamics, or the particle filter \textsc{loglik-pf} (\cref{alg:pf})
for stochastic dynamics --- so this SMC is agnostic to how the likelihood is formed.
Parameter priors are uniform over the proposer's declared bounds, so the
random-walk Metropolis rejuvenation accepts on the tempered-likelihood ratio
alone for a symmetric proposal, with proposals outside the support rejected.
The NumPy fitter instead clips proposals at the bounds, so it
violates detailed balance for this target and can bias posterior and evidence
estimates; its legacy results should not be read as a test of exact Bayesian inference. Its diagonal proposal
scale is half the current per-coordinate particle standard deviation
(floored at $10^{-3}$ before scaling).}
\label{alg:smc}
\begin{algorithmic}[1]
\Statex \textbf{def}\ \textsc{marglik-smc}$(\D, m;\ \textsc{loglik})\to(\{\theta_i\},\{W_i\},\log Z_m)$
\State sample $\theta_i \sim p(\cdot\mid m)$ for $i=1..\nparticlesParams$;\quad $W_i \gets 1/\nparticlesParams$
\State $\beta \gets 0$;\quad $\log Z_m \gets 0$
\State $\ell_i \gets \textsc{loglik}(\D, m,\theta_i)$ 
\While{$\beta < 1$} \Comment{$\le J_p$ annealing rungs}
  \State take $\Delta\beta=1-\beta$ if its ESS meets the target or the rung cap is reached; otherwise bisect to target ESS $\eta\nparticlesParams$
  \State $\log Z_m \mathrel{+}= \log \sum_i W_i\,e^{\Delta\beta\,\ell_i}$ \Comment{evidence increment}
  \State $W_i \gets W_i\,e^{\Delta\beta\,\ell_i} \big/ \textstyle\sum_j W_j\,e^{\Delta\beta\,\ell_j}$;\quad $\beta \mathrel{+}= \Delta\beta$
  \State resample $\{\theta_i\} \propto \{W_i\}$;\quad $W_i \gets 1/\nparticlesParams$
  \For{$s = 1 \dots \nroundsParams$} \Comment{RW-Metropolis at temperature $\beta$}
    \State propose $\theta_i' \sim q(\cdot\mid\theta_i)$; reject immediately if outside the prior bounds
    \State $\ell_i' \gets \textsc{loglik}(\D, m,\theta_i')$
    \State $(\theta_i,\ell_i)\gets(\theta_i',\ell_i')$ w.p.\ $\min\{1,\, e^{\beta(\ell_i'-\ell_i)}\}$
  \EndFor
\EndWhile
\State \Return $(\{\theta_i\},\{W_i\},\log Z_m)$
\end{algorithmic}
\end{algorithm}

\subsubsection{Recursive parameter and evidence updates}
\label{app:recursiveEvidence}

\begin{algorithm}[t]
\caption{\textsc{marglik-pf}: particle filtering over static parameters across experiments (\cref{eq:recursiveEvidenceParticles,eq:recursiveWeights}). This outer recursion is distinct from the latent-state filter \textsc{loglik-pf}; any chosen trajectory likelihood can supply its increments. Cached state is maintained separately for each model.}
\label{alg:marglikpf}
\begin{algorithmic}[1]
\Statex \textbf{def} \textsc{marglik-pf}$(\D_{0:\expt},m;\textsc{loglik})\to(\{\theta_i\},\{W_i\},\log\hat Z_m)$
\State \textbf{if} no valid cached state exists \textbf{then return} \textsc{marglik-smc}$(\D_{0:\expt},m;\textsc{loglik})$ and cache its result
\For{each experiment $(\design_j,y_j)$ not yet assimilated, in order}
  \State $a_i\gets\log W_i+\textsc{loglik}(\{(\design_j,y_j)\},m,\theta_i)$ for every $i$
  \State $c\gets\operatorname{logsumexp}_i(a_i)$
  \State $\log\hat Z_m\gets\log\hat Z_m+c$
  \State $W_i\gets\exp(a_i-c)$
  \If{$1/\sum_i W_i^2<\nparticlesParams/2$}
    \State systematically resample $\{\theta_i\}$ with probabilities $\{W_i\}$;\quad $W_i\gets1/\nparticlesParams$
  \EndIf
\EndFor
\State cache particles, weights, evidence, and assimilated experiment IDs
\State \Return $(\{\theta_i\},\{W_i\},\log\hat Z_m)$
\end{algorithmic}
\end{algorithm}

For many experiments or long trajectories, full-history refitting is expensive.
For a fixed model $m$, define the new-experiment likelihood
$L_{m,\expt}(\theta)=p(y_{\expt,1:T}\mid\design_\expt,m,\theta)$.
Conditioning on the chosen designs and assuming experiments are independent
given $(m,\theta)$, Bayes' rule gives the exact recursion
\begin{align}
 c_{m,\expt}
 &=\int L_{m,\expt}(\theta)\,
       p(\theta\mid m,\D_{0:\expt-1})\,d\theta,
 & Z_m^\expt&=Z_m^{\expt-1}c_{m,\expt},
 \label{eq:recursiveEvidence}\\
 p(\theta\mid m,\D_{0:\expt})
 &=\frac{L_{m,\expt}(\theta)\,
 p(\theta\mid m,\D_{0:\expt-1})}{c_{m,\expt}}.
 \label{eq:recursiveTheta}
\end{align}
Initialize $\widehat Z_m^0$ and the parameter particles by tempered SMC on
$\D_0$. Given normalized retained weights $W_{m,i}^{\expt-1}$, evaluate
only the new trajectory at each retained parameter value:
\begin{align}
 \widehat c_{m,\expt}
 &=\sum_{i=1}^{\nparticlesParams}W_{m,i}^{\expt-1}
       \widehat L_{m,\expt}(\theta_{m,i}^{\expt-1}),
 &\log\widehat Z_m^\expt
 &=\log\widehat Z_m^{\expt-1}+\log\widehat c_{m,\expt},
 \label{eq:recursiveEvidenceParticles}\\
 \widetilde W_{m,i}^{\expt}
 &=\frac{W_{m,i}^{\expt-1}\widehat L_{m,\expt}
 (\theta_{m,i}^{\expt-1})}{\widehat c_{m,\expt}}.
 \label{eq:recursiveWeights}
\end{align}
Here $\widehat L$ is the exact likelihood when tractable, or its PF estimate
for stochastic dynamics; the implementation computes the sums in log space.
The model weights are then proportional to $p(m)\widehat Z_m^\expt$ over the
current archive. If $1/\sum_i(\widetilde W_{m,i}^{\expt})^2$
falls below $\nparticlesParams/2$, we systematically resample parameter
particles and reset their weights to $1/\nparticlesParams$, without changing
the accumulated evidence. Otherwise we retain the weighted particles.
The HH SDE+PF experiment in \cref{app:hhstochmethods} uses this procedure, with
$\nparticlesParams=64$ and $\nparticlesLatent=128$, and no subsequent
parameter rejuvenation.

\paragraph{Cost and limitations.}
Excluding initialization, for fixed particle counts and no full-history
rejuvenation, the recursive cost per model is
$O(\nparticlesParams\sum_{\expt=1}^{\Nexpt} C(T_\expt))$,
where $C(T)$ is the cost of one trajectory likelihood.
For equal lengths and a tractable $O(T)$ likelihood, this is
$O(\Nexpt T\nparticlesParams)$ rather than
$O(\Nexpt^2 T\nparticlesParams)$; suppressing likelihood cost gives the
$O(\Nexpt\nparticlesParams)$ versus $O(\Nexpt^2\nparticlesParams)$ comparison.
For PF, $C(T)=O(T\nparticlesLatent)$ at fixed solver resolution.
Thus recursion avoids revisiting old experiments but does not eliminate
the cost of processing a long new trajectory. For a single growing time
series, analogous within-trajectory updates require retaining each
parameter particle's latent filtering state; the implemented HH outer
update instead assimilates whole, independently initialized experiments.

Repeated resampling can collapse parameter diversity: it duplicates existing
values rather than discovering new ones. The recursion is therefore a
finite-particle approximation, not a guarantee of accurate long-run evidence.
Weight ESS after resampling does not measure distinct parameter support;
particle impoverishment can understate parameter uncertainty and distort
both prediction intervals and acquisition scores.
(See \citep{chopin2002ibis,chopin2013smc2} for details.)
Posterior-invariant MCMC rejuvenation or occasional batch refits can help,
but may require evaluating the full history and invalidate the simple
linear-cost bound. A newly proposed structure has no inherited parameter
posterior or evidence and must be fitted on all accumulated data; the current
HH implementation refits when the archive changes.

\subsubsection{BIC approximation}
\label{app:BIC}

\begin{algorithm}[t]
\caption{\textsc{marglik-bic}: optimization and BIC evidence approximation. The Gaussian profiled-scale branch matches \cref{eq:BIC}; the direct-likelihood branch covers PF/BSL optimization. $N$ is the declared observation count, not the number of Monte Carlo particles. Comparisons require a common observation space and likelihood convention.}
\label{alg:marglikbic}
\begin{algorithmic}[1]
\Statex \textbf{def} \textsc{marglik-bic}$(\D,m;\textsc{loglik})\to(\{\hat\theta_m\},\{1\},\log\hat Z_m)$
\If{using Gaussian residuals with a profiled common variance}
  \State $\hat\theta_m\gets\arg\min_{\theta\in\Theta_m}\mathrm{RSS}_m(\theta)$ \Comment{bounded multi-start optimization}
  \State $b_m\gets N\log(\mathrm{RSS}_m(\hat\theta_m)/N)+\nparams_m\log N$
\Else
  \State $\hat\theta_m\gets\arg\max_{\theta\in\Theta_m}\textsc{loglik}(\D,m,\theta)$ \Comment{declared simulation budget for noisy likelihoods}
  \State $b_m\gets-2\,\textsc{loglik}(\D,m,\hat\theta_m)+\nparams_m\log N$
\EndIf
\State \Return $(\{\hat\theta_m\},\{1\},-b_m/2)$ \Comment{log evidence up to a common constant}
\end{algorithmic}
\end{algorithm}

As a faster alternative to tempered SMC, we use the Bayesian Information
Criterion (BIC), which we now explain.

Write $\ell_m(\theta)=\log p(\data\mid m,\theta)$ and
$\nparams_m=\dim(\theta_m)$. For an interior maximum and a smooth prior,
the large-sample Laplace approximation is
\begin{equation}
\log Z_m \approx \ell_m(\hat\theta_m)+\log p(\hat\theta_m\mid m)
 +\frac{\nparams_m}{2}\log(2\pi)-\frac12\log|H_m|,
\label{eq:laplace-bic}
\end{equation}
where $H_m=-\nabla^2\ell_m(\hat\theta_m)$. Under regularity conditions,
$H_m$ grows proportionally to $N$, so retaining the leading terms gives
\begin{equation}
 -2\log Z_m \approx \mathrm{BIC}_m
 =-2\ell_m(\hat\theta_m)+\nparams_m\log N.
\end{equation}
BIC drops bounded, generally model-dependent prior-density and curvature
terms. We normalize $p(m)\exp(-\mathrm{BIC}_m/2)$ over the candidate pool.
Below we discuss how to compute BIC,
with different noise assumptions.
See \cref{alg:marglikbic} for the pseudocode.

\paragraph{Fixed residual scale ($\tau=1$).}
For $N$ scalar observations, assume fixed positive whitening scales
$\sigma_n$ shared across candidates, and define
\begin{equation}
 \mathrm{RSS}_m(\theta)=\sum_{n=1}^N
 \left(\frac{y_n-\hat y_n(\theta)}{\sigma_n}\right)^2.
\end{equation}
With independent Gaussian errors and $\hat\theta_m$ minimizing this RSS,
\begin{equation}
 -2\log p(\data\mid m,\hat\theta_m)
 =\mathrm{RSS}_m+\sum_{n=1}^N\log(2\pi\sigma_n^2).
\end{equation}
For a common $\sigma_n=\sigma$, the last term is $N\log(2\pi\sigma^2)$.
This is the negative twice-log-likelihood, not yet BIC: adding the parameter
penalty and dropping the shared normalization gives
\begin{equation}
  \mathrm{BIC}^{\mathrm{fixed}}_m=\mathrm{RSS}_m+\nparams_m\log N.
  \label{eq:BICfixed}
\end{equation}
Here $\mathrm{RSS}_m=\mathrm{RSS}_m(\hat\theta_m)$ is already noise-whitened.

\paragraph{Fitted (profiled) residual scale.}
Alternatively, retain the same fixed whitening scales
$\sigma_n$, shared across candidate models, and an unknown common
residual-scale multiplier $\tau$,
giving the log likelihood
\begin{align}
  y_n\mid m,\theta_m,\tau
  &\sim \mathcal N\!\left(\hat y_n(\theta_m),\tau^2\sigma_n^2\right),\\
  -2\log p(\data\mid m,\theta_m,\tau)
  &= \frac{\mathrm{RSS}_m(\theta_m)}{\tau^2}+N\log\tau^2
  +\sum_{n=1}^N\log(2\pi\sigma_n^2).
\end{align}
The maximum-likelihood estimate $\hat\theta_m$ minimizes this weighted RSS.
Writing $\mathrm{RSS}_m=\mathrm{RSS}_m(\hat\theta_m)$, maximizing over $\tau$
gives $\hat\tau_m^2=\mathrm{RSS}_m/N$. Substitution yields
\begin{equation}
 -2\log p(\data\mid m,\hat\theta_m,\hat\tau_m)
 =N\log(\mathrm{RSS}_m/N)+N+\sum_{n=1}^N\log(2\pi\sigma_n^2).
\end{equation}
The full BIC adds $(\nparams_m+1)\log N$, since the fitted scale $\tau$
is one additional parameter.
The terms $N$ and $\sum_n\log(2\pi\sigma_n^2)$
and this additional $\log N$
are identical for every candidate on the same
dataset, so we drop them.
This gives the comparison score
\begin{align}
  \mathrm{BIC}_m &:=N \log(\mathrm{RSS}_m/N)+\nparams_m\log N.
 \label{eq:BIC}
\end{align}
where $\nparams_m=\dim(\theta_m)$ is the number of fitted model
parameters (excluding the common fitted residual scale).
Lower is better. We use $\log\widehat Z_m=-\mathrm{BIC}_m/2$ as an
approximate evidence score, and normalize $p(m)\exp(-\mathrm{BIC}_m/2)$ over the pool.
\eat{
This is an approximation, not an exact equality up to a universal constant:
under regular large-sample assumptions, the full BIC approximates
$-2\log Z_m$ while discarding bounded, generally \emph{model-dependent}
terms involving the parameter prior and local likelihood curvature.
Those terms are distinct from the exactly model-independent Gaussian
normalization terms dropped above. For singular models, noisy PF likelihoods,
or approximate synthetic likelihoods, BIC is used here as a scoring heuristic.
}

\paragraph{Noise terms $\sigma_n$.}
The term  $\sigma_n$ is a heteroscedastic whitening scale:
\begin{align}
  \sigma_n &= \max\!\left(\rho |y_n|,\sigma_{\mathrm{floor}}\right)
  \end{align}
In \CHEMbench we use
$\rho=0.03$ and $\sigma_{\mathrm{floor}}=0.01$.  For the stochastic
HH experiment, the deterministic ODE+BIC control instead uses a constant
$\sigma_n=2\,\mathrm{mV}$ (equivalently, $\rho=0$ and
$\sigma_{\mathrm{floor}}=2\,\mathrm{mV}$), matching the declared voltage
observation noise.  Each trajectory contributes 20 voltage observations, so
$N=20(N_0+\expt)$ after $\expt$ active experiments and $N_0$ initial trajectories.

\paragraph{Comparison of BIC estimators.}
In the end-to-end \CHEMbench control in \cref{fig:chem-controls}, fixed-scale
BIC gives better early-budget predictions than profiled-scale BIC under
CMA-EIG, although their performance is similar at 15 observations.

\subsection{Prior sensitivity}
The model evidence averages the likelihood over the parameter prior, rather than only
rewarding its maximum. A diffuse prior can therefore strongly penalize a
model even when its fitted predictions barely change
\citep{llorente2022safepriors}. For example, enlarging a uniform prior's volume
by a factor $c$, while leaving the likelihood integral essentially unchanged,
reduces $\log Z_m$ by $\log c$; an improper prior leaves ordinary Bayes factors
undefined.

In \cref{eq:laplace-bic}, BIC retains the leading terms
$\ell_m(\hat\theta_m)-\tfrac12\nparams_m\log N$, dropping prior-density and
curvature terms that can matter at small $N$. It avoids explicit prior-volume
normalization, but fitting bounds still matter, and unidentified parameters
violate its regularity assumptions. Thus a predictive advantage for BIC need
not mean that it estimates evidence more accurately.
\Cref{fig:chem-controls} compares predictive performance of complete inference
configurations; it does not isolate evidence-estimation accuracy.

\subsection{Computing the model posterior}
\label{app:posterior}

\begin{algorithm}[t]
\caption{\textsc{model-posterior}: normalize estimated log evidences plus log priors over the current pool (no LLM). Each model uses the selected evidence backend, including its cached state for recursive updates. The routine updates the stored pool and parameter states in place and returns approximate, pool-conditional model weights.}
\label{alg:model}
\begin{algorithmic}[1]
  \Statex \textbf{def}\ \textsc{model-posterior}$(\D,\ \mathcal M;\
  \nparticlesModel,\ \textsc{loglik}, \textsc{marglik})
  \to p(m\mid\D)$
\For{$m_i \in \mathcal M$}
  \State $(\{\theta_{ij}\},\{W_{ij}\},\log\hat Z_i) \gets\textsc{marglik}(\D,\ m_i;\ \textsc{loglik})$
\EndFor
\State $p(m_i\mid\D)\gets\mathrm{softmax}_i\big(\log\hat Z_i+\log\pi(m_i)\big)$ \Comment{structure prior $\pi$}
\State $\mathcal M\gets\textsc{top-}\nparticlesModel\big(\mathcal M;\ p(m\mid\D)\big)$ \Comment{evidence-prune pool to $\nparticlesModel$}
\State renormalize $p(m\mid\D)$ on the retained pool
\State store $\hat p(\D)=\sum_{m_i\in\mathcal M}\pi(m_i)\hat Z_i$ \Comment{finite-pool normalizer only}
\State \Return $p(m\mid\D)$
\end{algorithmic}
\end{algorithm}

The main inference procedure over models, given the current hypothesis
class $\mathcal M$, is shown in \cref{alg:model}.
This just applies Bayes rule to compute $p(m \mid \D)$ for each
$m \in \mathcal M$, and then keeps the top $\nparticlesModel$ candidates.

\eat{
\paragraph{Helper interfaces.}
Algorithm~\ref{alg:mda} stores backend choices and per-model inference state;
subsequent calls omit those stored arguments. Each marginal-likelihood routine
returns parameter support, weights, and log evidence; \textsc{model-posterior}
stores all three for prediction and design. Pool truncation updates the agent's
stored $\mathcal M$. \textsc{init-hyp-space} is defined in \cref{alg:initpool};
the likelihoods in \cref{alg:loglikdet,alg:pf,alg:loglikbsl}; the evidence routines
in \cref{alg:smc,alg:marglikpf,alg:marglikbic}; and expansion, checking, pruning,
and acquisition in \cref{alg:expand,alg:check,alg:adaptprune,alg:voi}.
The optimizer is CMA-ES for continuous ChemBench designs and enumeration for
the finite DP/HH menus; prediction uses \cref{eq:BMA} or its stated plug-in variant.
}

\begin{algorithm}[t]
\caption{\textsc{adapt-and-prune}: optional score-ranked pool shrinkage. The fit statistic and thresholds are domain-specific; ChemBench uses maximum weight $>0.9$ and median relative fit residual $\le0.05$ to select caps 6 versus 24 (\cref{app:chemPoolAdaptation}). This does not merge predictively similar models or restore previously discarded models.}
\label{alg:adaptprune}
\begin{algorithmic}[1]
\Statex \textbf{def} \textsc{adapt-and-prune}$(\mathcal M,p,\D)\to(\mathcal M,p,\nparticlesModel)$
\State \textbf{if} adaptation disabled \textbf{then return} $(\mathcal M,p,\nparticlesModel)$
\State $m^*\gets\arg\max_m p(m)$
\State $c\gets N_{\max}$
\State \textbf{if} $p(m^*)>\tau_c$ and fit error of $m^*$ on $\D$ is $\le\tau_{\rm fit}$ \textbf{then} $c\gets N_{\min}$
\If{$|\mathcal M|>c$}
  \State retain the $c$ models with largest $p(m)$ in $\mathcal M$
  \State invalidate inference caches if a full refit is required by the backend
  \State $p\gets\textsc{model-posterior}(\D,\mathcal M;c,\textsc{loglik},\textsc{marglik})$
\EndIf
\State \Return $(\mathcal M,p,c)$
\end{algorithmic}
\end{algorithm}

\paragraph{$\mathcal{M}$-open extension.}
When the current best (MAP) model's residual (error) $e^*$
exceeds a threshold $\tau_e$ (\cref{alg:check}), we invoke
\textsc{expand-hyp-space} (\cref{alg:expand}): the LLM proposes a batch of
$N_{\new}$ new structures from the observations and available fit diagnostics.
ChemBench provides residuals for the current pool, inspired by the pool-conditioned
proposals of SMC-S \citep{piriyakulkij2024}; glucose supplies observed trajectories
and candidate fit diagnostics.
The new
structures are scored by evidence and the pool is evidence-pruned in the following \textsc{model-posterior}
call. If $e^*\le\tau_e$ the pool is left unchanged and \textsc{model-posterior} merely re-scores the existing
structures on the new datum --- no LLM call.
\eat{
\HHbench, \CHEMbench, and \textsc{BoxingGym} all expand on
demand this way, including \DPbench, whose LLM-proposed force-law pool is initialised from the seed orbit and
then expanded on the same residual-triggered schedule.
}

\paragraph{The predictive check.}
\label{app:predictiveCheck}
The predictive check measures how well the current model explains observations;
a large discrepancy triggers $\mathcal M$-open expansion.
\Cref{alg:check} gives a prequential variant: score the new observation before
fitting it, optionally combining this with residuals on stored data.
The executed ChemBench and DP agents instead gate their first expansion on
residuals cached from the preceding fit. They then refit on all available
data and, if expanding again, recheck that fit. A newly surprising observation
can therefore affect expansion one update later. Neither version retains a
running archive of historical forecasts. Relative residuals balance component
scales, and taking their median limits the influence of outliers, but can
also delay expansion after a single informative failure.
More precisely, let the relative error on experiment $j$ be
the median relative error over target components $i$:
\begin{align}
  r_j(m,\theta)
  &= \underset{i}{\operatorname{median}}
    \frac{\left|\target_i(y_j)-
    \E[\target_i(Y)\mid \design_j,m,\theta]\right|}
         {\max\!\left(\left|\target_i(y_j)\right|,c_i\right)}
\end{align}
Then the estimated  error at step $k$ is
\begin{align}
  e^*&=\operatorname{median}
  \left(
    \{r_\expt(m^*,\hat\theta^{m^*})\}
    \cup
    \{r_j(m^*,\hat\theta^{m^*}):j<\expt\}_{\mathtt{backtest}}
  \right),
  \label{eq:predictiveCheck}
\end{align}
where
the first set is the prequential error on the new observation and the optional second set contains the
back-test errors.  Here $m^*$ and $\hat\theta^{m^*}$ are fixed at their pre-observation values when the new
datum is scored.  Expansion is triggered when $e^*>\tau_e$.

\paragraph{Finite-pool interpretation and limitations.}
The outer (model inference) loop, at step $\expt$, is
approximating the following target distribution
\begin{equation}
  p_\expt(m)\ \propto\ p(m)\,Z_m^{(\expt)}
  \label{eq:smc3target}
\end{equation}
On a deduplicated finite pool $\mathcal M$, exact evidences would give
the restricted posterior $w_i\propto p(m_i)Z_{m_i}^{(\expt)}$.
In practice the evidences, or their BIC substitutes, are approximate.
When enabled, residual-conditioned LLM proposals enlarge this support
\citep{piriyakulkij2024}; top-$\nparticlesModel$ pruning is deterministic
hypothesis search, not a posterior-invariant resampling kernel.
Normalizing over enumerated unique candidates does not require treating
them as importance samples from the LLM, but it also does not account for
unproposed model mass.

\eat{
For an exact finite-pool posterior, truncation and renormalization change
the distribution by its discarded mass in total variation. This says
nothing about models never proposed, and noisy evidence prevents treating
the observed discarded weight as an exact error bound. Neither full refitting
nor resampling guarantees that model or parameter weights remain diffuse.

Full-history refitting is one implementation, not a necessity of parameter
marginalization. Recursive parameter/state inference is possible
\citep{chopin2002ibis,chopin2013smc2}; the HH implementation retains parameter
particles and updates with the newest likelihood when the archive is unchanged.
This saves computation but may lose particle diversity. Retaining low-weight
models in an archive and retaining diverse parameter particles are distinct issues.
}

\begin{algorithm}[t]
\caption{\textsc{init-hyp-space}: domain-configured initialization. ChemBench hybrid mode combines canonical laws with LLM proposals; truth-blind initialization supplies no privileged true model. Validation checks executability and parameter bounds, not held-out performance.}
\label{alg:initpool}
\begin{algorithmic}[1]
\Statex \textbf{def} \textsc{init-hyp-space}$(\textsc{llm},C,\D_0)\to\mathcal M$
\State $\mathcal M\gets$ configured seed library (possibly empty)
\If{LLM initialization enabled}
  \State propose a batch of $N_{\rm init}$ models from $\textsc{llm}(\cdot\mid C,\D_0)$
  \State validate candidates; merge valid candidates into $\mathcal M$, removing exact duplicates
\EndIf
\State \textbf{if} $\mathcal M=\varnothing$ \textbf{then} retry within the declared limit or report initialization failure
\State \Return $\mathcal M$ \Comment{scoring and capping follow in \textsc{model-posterior}}
\end{algorithmic}
\end{algorithm}

\begin{algorithm}[t]
\caption{\textsc{expand-hyp-space} --- grow the pool with a batch of $N_{\new}$ LLM-proposed structures. Pool residuals are optional context (used in ChemBench; DP supplies observation summaries). Scoring
and pruning are deferred to the next \textsc{model-posterior} (\cref{alg:model}); this is the sole LLM call.}
\label{alg:expand}
\begin{algorithmic}[1]
\Statex \textbf{def}\ \textsc{expand-hyp-space}$(\textsc{llm},\ \mathcal M,\ \D,\ C;\ N_{\new})\to\mathcal M'$
\State optionally compute $\rho_j \gets \textsc{residuals}(m_j,\D)$ for $m_j\in\mathcal M$ \Comment{backend-specific context}
\State propose $N_{\new}$ candidates using $C$, observation summaries, and any available pool diagnostics
\State \Return $\mathcal M\cup\{m'_l\}$
\end{algorithmic}
\end{algorithm}

\begin{algorithm}[t]
\caption{\textsc{predictive-check} --- score how well the current
  belief predicts the target $\target$, returning the median residual
  $e$ that drives the $\mathcal M$-open trigger of \cref{alg:mda}.
  The freshly observed datum is scored
  \emph{prequentially} (the current MAP model $m^\ast$ has not yet
  seen $y$); with the \textsc{backtest} flag (default on), the stored
  data $\D_{0:\expt-1}$ are also re-scored against $m^\ast$,
  and the check returns the
  median error over the combined set. The per-datum residual
  $\tilde{\ell}(a,b){=}\operatorname{median}_i|a_i{-}b_i|/\max(|a_i|,c_i)$ is
  per-component \emph{relative} error, where $a=\target(y)$ is the observed
  target and $b=\hat\target(\design)$ its forecast,
  with a small floor of $c_i$ for numerical stability.
  This is useful for a target with
  heterogeneous scales (e.g.\ \HHbench's
  $[n_\test,V_{\min},V_{\mathrm{end}}]$), so the error is  not dominated by its
  largest component.
  For the identity-target rungs ($\target(y_{1:T})=y_{1:T}$),
  the median is computed over time steps $t=1:T$.
  The first call uses the pre-observation fit; subsequent calls following
  expansion use a refitted model and are in-sample adequacy checks, not new
  prequential evidence. The HH stochastic gate uses a different statistic.
  }
\label{alg:check}
\begin{algorithmic}[1]
  \Statex \textbf{def}\ \textsc{predictive-check}$\big(
  \design,\,y,\ \D_{0:\expt-1},\ p(m\mid\D),\ \textsc{backtest}=T\big)\to e$
\State $m^\ast\gets\arg\max_m p(m\mid\D)$ \Comment{MAP under the supplied fit}
\State $\hat\target(\design)\gets\E[\target(Y)\mid\design,\,m^\ast,\hat\theta^{m^\ast}]$ \Comment{MAP model's target forecast}
\State $E \gets \big\{\,\tilde{\ell}\big(\target(y),\,\hat\target(\design)\big)\,\big\}$
\Comment{prequential only if the supplied fit excludes $y$}
\If{\textsc{backtest}}
\State $E \gets E \cup \big\{\, \tilde{\ell}\big(\target(y_j),\,\hat\target(\design_j)\big)\
:\ (\design_j,y_j)\in\D_{0:\expt-1} \,\big\}$ \Comment{in-sample residuals}
\EndIf
\State \Return $\operatorname{median}(E)$
\end{algorithmic}
\end{algorithm}

\subsection{Algorithms for experiment design}
\label{app:VOI}

\paragraph{VoI API.}
The design step maximises a \textsc{VoI} score over the design space.
\Cref{alg:voi} defines the interface,
which reads off the  pooled two-level
posterior $p(m\mid\D)\,W_i^m$.
The objectives
only differ in \emph{which} uncertainty they reduce ---
the model index (model-discrimination EIG, the default),
or the whole latent $(m,\theta)$ (joint EIG).
We derive these objectives below.

\begin{algorithm}[t]
\caption{\textsc{VoI} --- the experiment-design score at $\design$ under the two-level posterior, with a
\textsc{flavour} selector. All flavours use only the per-particle predictions
$\mu_i(\design)=\E[Y_\design\mid m_i,\theta_i,\design]$ (weights $W_i^m$ within structure $m$),
noise-whitened by the observation covariance $\Sigma_\varepsilon$. \textsc{model} is Lindley's
model-disagreement surrogate (default); \textsc{joint} keeps within-structure
parameter uncertainty;
}
\label{alg:voi}
\begin{algorithmic}[1]
\Statex \textbf{def}\ \textsc{VoI}$(\design,\ p(m,\theta\mid\D);\ \textsc{flavour})\to \real_{\ge 0}$
\State $\bar\mu_m(\design)\gets\textstyle\sum_i W_i^m\,\mu_i(\design)$;\quad $\bar\mu(\design)\gets\textstyle\sum_m p(m\mid\D)\,\bar\mu_m(\design)$
\State \Return
$\begin{cases}
  \textstyle\sum_m p(m\mid\D)\,\big\lVert\bar\mu_m(\design)-\bar\mu(\design)\big\rVert^2_{\Sigma_\varepsilon^{-1}}
    & \textsc{model}\ \ \text{(\cref{eq:voi})}\\[3pt]
  \textstyle\sum_{m,i} p(m\mid\D)\,W_i^m\,\big\lVert\mu_i(\design)-\bar\mu(\design)\big\rVert^2_{\Sigma_\varepsilon^{-1}}
    & \textsc{joint}\ \ \text{(\cref{eq:voiFull})}\\[3pt]
\end{cases}$
\end{algorithmic}
\end{algorithm}

\paragraph{From Value of information to expected information gain.}
The VoI,
first proposed in \citep{howard1966},
is a \emph{decision-theoretic} concept defined as follows:
for a terminal decision $\terminal \in \terminalSpace$ with
utility $U(a,h)$ over the unknown state of nature $h$,
the value of running experiment $\design$ is the expected gain in
attainable utility from observing its outcome \emph{before} deciding,
\begin{equation}
  \mathrm{VoI}_U(\design)
  = \E_{y\sim p(\cdot\mid\design,\D)}\Big[\,
    \max_{\terminal \in \terminalSpace}\ \E_{p(h\mid\D,y,\design)}\,U(\terminal,h)\,\Big]
  \;-\; \max_{\terminal \in \terminalSpace}\ \E_{p(h\mid\D)}\,U(\terminal,h),
  \label{eq:voidecision}
\end{equation}
measured in the \emph{units of the task utility} $U$.

Now let the decision be model
\emph{identification}, i.e.\ report a distribution $q(\cdot)$ over the
model index
$h=m$ under the log score
$U(q,m)=\log q(m)$. The inner maximiser is the posterior, $q^\star=p(M\mid\D,y,\design)$, and
$\max_q\E_{p(M\mid\cdot)}\log q(M)=-H[p(M\mid\cdot)]$, so \cref{eq:voidecision}
becomes
\begin{equation}
  \mathrm{VoI}_{\log}(\design)
  = \Big(-\,\E_y\,H[p(M\mid\D,y,\design)]\Big) - \Big(-H[p(M\mid\D)]\Big)
  = I(M;Y_\design\mid\D)
  = \mathrm{EIG}(\design).
  \label{eq:voieig}
\end{equation}
This is known as the \emph{Expected information gain} (EIG),
and was first proposed by \citep{lindley1956}.
(We discuss other forms of VoI below.)

\paragraph{Estimating the EIG.} We estimate the EIG
as follows.
First we rewrite the mutual information of \cref{eq:voieig} in its
\emph{outcome-entropy} form, $I(M;Y_\design\mid\D)=H[Y_\design\mid\D]-H[Y_\design\mid M,\D]$,
in terms of each structure's \emph{posterior predictive} of the outcome,
\begin{equation}
  P_m(\cdot)\;:=\;p\big(Y_\design\mid m,\D,\design\big)
  \;=\;\int p\big(Y_\design\mid m,\theta,\design \big)\,p(\theta\mid m,\D)\,d\theta,
  \label{eq:predm}
\end{equation}
i.e.\ the distribution of the outcome we would observe under design $\design$ if structure $m$ were true,
with its parameters marginalised over the within-structure posterior. Writing $p(m):=p(m\mid\D)$, the marginal
predictive is the mixture $p(Y_\design\mid\D)=\sum_m p(m)\,P_m$, so $H[Y_\design\mid\D]=H\!\big(\sum_m
p(m)P_m\big)$; and conditioning on $M{=}m$ makes $Y_\design\sim P_m$, so $H[Y_\design\mid
M,\D]=\sum_m p(m)\,H(P_m)$. Hence
\begin{equation}
  \EIG(\design)\;=\;I(M;Y_\design\mid\D)
  \;=\; H\!\Big(\sum_m p(m)\,P_m\Big)\;-\;\sum_m p(m)\,H(P_m).
  \label{eq:eigmix}
\end{equation}
The first term is the entropy of the \emph{mixture} predictive (the total uncertainty about the outcome); the
second is the average entropy \emph{within} a structure (the irreducible outcome noise that cannot help
discriminate $M$). Their difference is the outcome uncertainty attributable to not knowing which structure is
true --- equivalently $I(M;Y_\design\mid\D)=\sum_m p(m)\,\operatorname{KL}\!\big(P_m\,\big\|\,\sum_{m'}p(m')P_{m'}\big)$,
the mean KL from each structure's predictive to the mixture (a generalised Jensen--Shannon divergence), which
is manifestly maximised by designs $\design$ that drive the $P_m$
apart.

\paragraph{Gaussian approximation.}
The MI is intractable for general mixture, so we specialise to a
Gaussian likelihood model and use a deterministic moment-matching approximation.
Let $Y_\design\in\real^d$ be the quantity the likelihood conditions
on, which could be  the raw trace $y_{1:T}$ or some summary.
Model its outcome, conditional on the model $m$,
as $Y^m_\design= \bar\mu_m(\design)+\varepsilon$,
 where
 \begin{align}
   \bar \mu_{m}(\design) & =\E_{\theta \mid m, \D}\!\left[\E[Y_\design\mid m,\theta,\design]\right]
\end{align}
 and $\varepsilon\sim\gauss(0,\Sigma_\varepsilon)$ is an assumed common
 within-model noise covariance ($\sigma^2 I$ in the simplest case).
 Parameter and process uncertainty generally make the actual within-model
 predictive covariance model-dependent; the following bound applies to
 this common-covariance surrogate model, not automatically to the original mixture.
 The unconditional distribution is thus the Gaussian mixture
 \begin{align}
 Y_\design\sim\sum_m
 p(m\mid\D)\,\gauss(\bar\mu_m(\design),\Sigma_\varepsilon)
 \end{align}
 with covariance $\Sigma_\varepsilon+\Sigma_\mu(\design)$, where
 $\Sigma_\mu(\design)=\operatorname{Var}_{p(m \mid\D)}[\bar\mu_m(\design)]$
 is the $d\times d$ between-class
covariance of the mean predictions. The conditional entropy
$H[Y_\design\mid M]=\tfrac12\ln\!\big((2\pi
e)^d\det\Sigma_\varepsilon\big)$ is exact,
but a Gaussian
mixture has \emph{no closed-form} differential entropy, so we \emph{approximate} $H[Y_\design]$ by
the entropy of a single Gaussian of the same covariance (moment
matching).
The $(2\pi e)^d\det\Sigma_\varepsilon$ cancels in the difference,
giving
\begin{equation}
  I(M;Y_\design\mid\D)=H[Y_\design]-H[Y_\design\mid M]
  \approx\tfrac12\ln\det\!\Big(I+\Sigma_\varepsilon^{-1}\Sigma_\mu(\design)\Big),
  \label{eq:migauss}
\end{equation}
In  the scalar case $d{=}1$, this becomes
\begin{equation}
  I(M;Y_\design\mid \D) \approx \tfrac12\ln(1+\operatorname{Var}[\mu(\design)]/\sigma^2)
\end{equation}
This is
an \emph{upper bound} on the true mutual information,
since a Gaussian maximises entropy at fixed
covariance.
Furthermore, it is monotone (in the Loewner order\footnote{
The \emph{Loewner (partial) order} on symmetric matrices:
$A\succeq B$ iff $A-B$ is positive semidefinite \citep{pukelsheim2006}. With the same noise covariance at both designs, $\Sigma_\mu(\design)\succeq
\Sigma_\mu(\design')$ implies \cref{eq:migauss} is at least as large
at $\design$,
so a design that raises the
between-class covariance in \emph{every} direction has a no-smaller surrogate score.
For design-dependent noise, compare noise-whitened covariances instead.})
in the between-class covariance $\Sigma_\mu(\design)$.
This makes it a principled surrogate --- a design that raises the between-class covariance can only raise the
score --- although, since different designs generally induce \emph{incomparable} covariances under the Loewner
order, we do not claim its $\arg\max$ exactly coincides with that of the true mixture mutual information.

\paragraph{Weighted model disagreement.}
A cheaper surrogate replaces the log-determinant by the trace of
$\Sigma_\varepsilon^{-1}\Sigma_\mu(\design)$.
This resembles information-matrix criteria in optimal design
\citep{chaloner1995,pukelsheim2006}, but is not itself a posterior
ellipsoid volume or exact mutual information for a categorical model index.
Unlike EIG, which is bounded by $H(M\mid\D)$, this trace can be arbitrarily large.
In practice, we therefore use the following
\begin{align}
  \design^\star
  &= \arg\max_{\design\in\designSpace}\ \operatorname{tr}\!\big(\Sigma_\varepsilon^{-1}\Sigma_\mu(\design)\big) \\
  &= \arg\max_{\design}\ \textstyle\sum_m p(m\mid\D)\,\big\lVert\bar\mu_m(\design)-\bar\mu(\design)\big\rVert^2_{\Sigma_\varepsilon^{-1}},
  \label{eq:voi} \\
   \bar \mu_{m}(\design) & =\E_{\theta \mid m, \D}\!\left[\E[Y_\design\mid m,\theta,\design]\right]
  \approx \sum_i W_i^{m}\mu_{m,\theta_m^i}(\design)  \\
\bar\mu(\design)&=\sum_m p(m\mid\D)\,\bar\mu_m(\design)
\end{align}
where $\lVert v\rVert^2_{A}{=}v^\top A v$,
and  we have assumed the parameter posterior
 is represented as a set of weighted samples,
 $p(\theta|m,\D) = \sum_i W_i^{m} \delta(\theta-\theta_m^i)$.
(Note that using per-class \emph{means} $\bar\mu_m(\design)$
rather than noisy draws keeps EIG genuine epistemic disagreement
--- Lindley's intuition \citep{lindley1956} that the best experiment is the one whose outcome current
beliefs least agree on.)

\paragraph{Why moment-matching rather than nested Monte Carlo.}
With a finite model pool and exact conditional predictive densities,
an unbiased outer Monte Carlo estimator of EIG is:
\begin{equation}
  \widehat{\mathrm{EIG}}(\design)
  = \frac{1}{N}\sum_{n=1}^{N}\log\frac{p(y_n\mid m_n,\design)}
         {\sum_{m'}p(m'\mid\D)\,p(y_n\mid m',\design)},
  \qquad m_n\sim p(m\mid\D),\ \ y_n\sim p(Y_\design\mid m_n,\D),
  \label{eq:nestedmc}
\end{equation}
where the inner sum is the marginal predictive $p(y_n\mid\design,\D)$.
(A further inner loop over the
particles $\theta$ appears when parameters are not marginalised
analytically).
This is the estimator
\textsc{BoxingGym}'s \verb|info_gain| evaluates.
The displayed estimator is unbiased under those assumptions. Approximating
the conditional densities inside the logarithm introduces additional
finite-inner-sample bias; that nested approximation is distinct from the
outer Monte Carlo error. The finite-model sum costs
$O(N\lvert\mathcal M\rvert)$ per candidate design
\citep{rainforth2024,foster2021boed}.
This yields a noisy, slow, non-smooth objective for the design
search, which is why we maximise the fast, closed-form surrogate
\cref{eq:voi} instead.

\eat{
(When an exact estimator of the MI is cheap
(e.g., a discrete model index with a tractable likelihood),
we compute $I(M;Y_\design)$ exactly instead of moment-matching.
We reserve the MC estimator for spot-checking these fast surrogates.
Note that
\textsc{BoxingGym}
\citep{gandhi2025boxinggym} uses  this nested MC estimator
to {\em evaluate} the EIG of an LLM-proposed action under the
ground-truth model, but they do not optimize it.
}

\paragraph{EIG for model and parameters.}
An alternative acquisition keeps the within-class
parameter spread instead of averaging it out. By the law of total variance the \emph{full}
two-level predictive variance splits into the between-class term of Eq.~\eqref{eq:voi} plus a
within-class one:
\begin{align}
  \operatorname{tr}\!\Big(\Sigma_\varepsilon^{-1}\operatorname{Var}_{p(m,\theta\mid\D)}\!\big[\,\E(Y_\design\mid m,\theta,\dopo(\design))\,\big]\Big)
  &=\underbrace{\sum_m p(m\mid\D)\big\lVert\bar\mu_m(\design)-\bar\mu(\design)\big\rVert^2_{\Sigma_\varepsilon^{-1}}}_{\text{between classes (mechanism disagreement)}} \nonumber\\
  &\quad+\underbrace{\sum_m p(m\mid\D)\sum_i W_i^{m}\big\lVert\mu_{m,i}(\design)-\bar\mu_m(\design)\big\rVert^2_{\Sigma_\varepsilon^{-1}}}_{\text{within class (parameter uncertainty)}},
  \label{eq:voiFull}
\end{align}
estimated empirically by the pooled particle variance
$\sum_{m,i} w_{m,i}\big\lVert\mu_{m,i}(\design)-\bar\mu(\design)\big\rVert^2_{\Sigma_\varepsilon^{-1}}$ with
$w_{m,i}\!\propto\! p(m\mid\D)\,W_i^{m}$. Only the between-class term is collapsed by identifying
the class.
The within-class term is residual parameter uncertainty.
If the per-structure parameter posteriors have concentrated,
this latter term becomes negligible, so  this full variance coincides with
Eq.~\eqref{eq:voi}.
However, optimizing \cref{eq:voiFull} can be useful if the parameter
values contain more useful information than just the model index.
(We tried this version in \cref{app:locationResults},
but found it to be less numerically stable than \cref{eq:voi}.)

\paragraph{Optimising over a large design space.}
When the design space is large, we can use various gradient free optimizers to pick the design.
For continuous spaces a common choice is CMA-ES \citep{hansen2016cma}.
For discrete spaces,  we can use  LLM-driven evolutionary search methods
such as FunSearch \citep{romeraparedes2024funsearch}.

\paragraph{From model-discrimination to task-aware design.}
So far we have assumed the goal is to identify the true model,
so we have optimized EIG $I(M;Y_\design)$,
spending  budget to resolve which \emph{structure} generated the
data. But since we ultimately evaluate performance in terms of
prediction accuracy of a target variable,
we only need to resolve the model
\emph{to the extent that it changes the forecast of the
target} $\target$ on the query distribution $\queryDist$. The general
\emph{task-aware} value of information is the decision-theoretic
\cref{eq:voidecision} with the terminal action set to \emph{forecasting
the target} rather than identifying the model: at a future
experimental query
$\design_q \sim\queryDist$,
nature returns observation $Y_{\design_q}$,
from which we get feature $\target_q = \target(Y_{\design_q})$;
the agent reports a forecast $\hat \target_q$ scored
by a loss $\ell\big(\target_q,\hat \target_q\big)$;
the Bayes risk (expected loss for the optimal estimator) for this query is given by
\begin{equation}
  \mathcal{R}_\ell(\design_q \mid\D)\;=\;
  \min_{\hat \target_q}\ \E_{p(Y_q \mid\design_q,\D)}\,\ell\big(\target(Y_q),\hat \target_q\big),
  \label{eq:taskrisk}
\end{equation}
So the value of  running experiment $\design$ now
is the expected reduction in that
risk once the outcome $Y_\design$ is
observed, averaged over the query distribution:
\begin{equation}
  \mathrm{VoI}^{\queryDist}_\ell(\design)
  \;=\; \E_{\design_q\sim\queryDist}\Big[\, \mathcal{R}_\ell(\design_q\mid\D)\;-\;
    \E_{y_\design \sim p(\cdot\mid\design,\D)}\
    \mathcal{R}_\ell(\design_q\mid\D,y_{\design},\design)\,\Big].
  \label{eq:taskvoigeneral}
\end{equation}

This is exactly \cref{eq:voidecision} with decision $\terminal=\hat \target$ and
utility $U=-\ell$ evaluated on the target: it
designs $\design$ to most improve the forecast of $\target$ on future
queries, marginalising over which structure is true rather than trying
to identify it. Two structures that agree on $\queryDist$ need never
be told apart. This is \emph{task-driven} (goal-oriented) experimental
design proposed in \citep{rainforth2024,bickfordsmith2023}.
We leave systematic invesigation of this objective to future work.

\clearpage
\section{Chemistry}
\label{app:chem}

\subsection{Benchmark and metrics}
\label{app:chem_benchmark}

ActiveSciBench-Chem \citep{kabra2026autoscilab}, which we call \CHEMbench, asks an agent to discover a static
rate law
\begin{equation}
  y=f(C_A,C_I,C_B,C_P,\mathrm{Enz},T,\mathrm{pH};\theta).
  \label{eq:ratelaw-app}
\end{equation}
The seven inputs control substrate, inhibitor, second-substrate and product concentrations, enzyme loading,
temperature, and pH. The 57 supported worlds comprise nine canonical mechanisms and 48 compound mechanisms.
We evaluate every world at medium difficulty,
and $1\%$ multiplicative observation noise.
We use three replicate IDs per world;
these are algorithmic seeds, not three redraws of the observation noise at a fixed design.
The oracle deterministically hashes the world and seven-dimensional condition
to its noise draw, so repeating
exactly the same condition returns the same measured rate.
The seed changes MDA's particle fitting and design
search, \SciLab's controller trajectory,
and the held-out query conditions; different selected conditions
therefore generally encounter different deterministic noise values.

The benchmark evaluates a submitted law on novel conditions using
the root mean square log error:
\begin{equation}
\mathrm{RMSLE}=\left[\frac1N\sum_{i=1}^N
  \left(\log(1+\hat y_i)-\log(1+y_i)\right)^2\right]^{1/2}.
\label{eq:rmsle}
\end{equation}
Our held-out predictive score specializes \cref{eq:nmse} to a scalar
log-rate target, with $s_1^2$ equal to its test-set variance:
\begin{equation}
  \mathrm{nMSLE}=
  \frac{N^{-1}\sum_i\left(\log(1+\hat y_i)-\log(1+y_i)\right)^2}
       {\operatorname{Var}[\log(1+y_i)]}.
  \label{eq:chemnmse}
\end{equation}
We report the \emph{benchmark pass rate}, the fraction of runs satisfying
$\Ind[\mathrm{RMSLE}<0.05]$. Although the paper text of
\citet{kabra2026autoscilab} mentions $0.01$, its code uses $0.05$, which we follow. A law that
cannot be evaluated over the declared benchmark test range counts as a failure.

\subsection{MDA model and experimental protocol}
\label{app:chem_priors}

Each LLM-proposed law has free constants $\theta$ with positive bounds.
The primary ChemBench implementation
uses bounded parameter point estimation
and BIC weights in \cref{eq:BIC}; its predictor selects the highest-weight
structure and uses $f_m(a;\hat\theta_m)$.
The SMC control instead uses independent bounded priors for each
parameter, an explicit complexity prior $p(m)\propto\exp(-2\nparams_m)$, and a particle approximation to
$p(\theta\mid m,\D)$ and the marginal evidence. Both fit residuals in log-rate space.

The primary hybrid $\mathcal M$-open
arm begins with nine canonical executable laws and makes one LLM
initialization call requesting 12 additional laws, giving a nominal initial pool
of up to 21 before pruning; screening and deduplication can reduce this number.
When the resulting pool fails the predictive-adequacy check,
each subsequent expansion call requests a batch of six residual-directed proposals;
the live pool is score-pruned to at most 24 models, and can shrink to six
after a confident, adequate fit as described below.
The primary arm uses BIC weights and a point estimate, so the expectation
in \cref{eq:chem-model-disagreement} is evaluated at $\hat\theta_m$.
\eat{
Comparisons against the fixed nine-law control change both the initial
library and permission for later expansion. They do not isolate the causal
benefit of later births; that would require freezing the identical initial
hybrid archive in a matched control.
}

\paragraph{Initialization.}
The MDA agent is initialized with  three fixed passive
observations followed by a sequence of active experiments of batch size 1.
The passive set holds $(C_I,C_B,C_P,\mathrm{Enz},T,\mathrm{pH})=(0,1,0,1,310,7)$
fixed and uses $C_A\in\{0.03,0.3,3\}$.
It is a law-agnostic substrate sweep, shared in form across every MDA cell.

\paragraph{Experiment design.}
After
fitting the hypothesis pool to these three points, each active round scores a condition by the posterior
between-model variance
\begin{equation}
  V(a)=\sum_m p(m\mid\D)\left(\bar\mu_m(a)-\sum_{m'}p(m'\mid\D)\bar\mu_{m'}(a)\right)^2,
  \qquad
  \bar\mu_m(a)=\E_{p(\theta\mid m,\D)}[f_m(a;\theta)].
  \label{eq:chem-model-disagreement}
\end{equation}
This is the same as \cref{eq:voi} without the 
$\Sigma_\epsilon(a)^{-1}$ factor.
Under the benchmark's multiplicative noise this makes it a
model-disagreement proxy rather than the exact implemented EIG objective.

CMA-ES maximizes the objective
in \cref{eq:chem-model-disagreement}
over the continuous normalized seven-dimensional box, using 48
objective evaluations per round; concentration and enzyme axes are mapped logarithmically and temperature and
pH linearly. The oracle evaluates the maximizing condition, then MDA refits before choosing the next one.
Thus the active designs are adaptive and generally differ across arms, worlds, and replicate IDs.

\paragraph{Adaptive pool size and pruning.}
\label{app:chemPoolAdaptation}
After fitting each update, MDA checks whether the largest
normalized model weight exceeds $0.9$ and its residual ratio is at most $1$
(median relative fit residual at most $0.05$).
If both hold, it retains the six highest-scoring models and refits;
otherwise the cap is 24. For the BIC arm, ranking uses BIC weights.
Raising the cap does not itself restore discarded candidates: growth requires
new proposals. Exact-expression matching prevents re-adding an expression
already in the pool, but this is not a similarity-aware merge of predictions
or a guarantee against splitting weight among near-equivalent laws.
Pruning can discard a law that would become useful later; the live pool is
not an indefinitely retained archive of every proposal.

The pool sizes in \cref{fig:mainchem-storyboard} are measured after
proposal, fitting, and pruning. At $B=6$, five new retained expressions take
the pool from nine to 14; at $B=8$, adaptive shrinkage removes eight, leaving
six. At $B=15$, the proposal transcript requests six candidates, and three
new expressions survive while three old ones leave: the final size remains
six. The red markers indicate calls that enlarged the candidate set before
pruning, not a positive net change in the final live pool.

\subsection{Baseline}
\label{app:chemBaseline}

We run the complete \SciLab pipeline independently at each budget
$B\in\{5,10,15,20,25\}$, using GPT-5.6-Luna with medium reasoning effort.
Each run collects exactly $B$ observations and receives one final 800-iteration
\textsc{PySR} fit; intermediate symbolic refinement is disabled, including at
$B=20$ where the upstream default would otherwise add another fit.

In our initial experiments we noticed a very large error spike
at $B=10$ (a smaller spike is still visible in \cref{fig:mainchem-curve}).
The cause was \SciLab's model selector,
which favored nearly saturated seven-variable  power laws
because it compared their \emph{training scores} against other candidates’
\emph{validation scores},
 allowing severe overfitting.
 (This effect does not appear with the larger experimental  budgets
 used in their paper.)
 By slightly modifying their code to 
 rank all candidates on the same validation split,
 we reduced the error by about $9 \times$ on the matched cohort
 (although a smaller spike remains),
 without changing results at other sample sizes.

\subsection{Results}
\label{app:chem_results}

\paragraph{Sample efficiency.}
The primary sample efficiency results are shown in
\cref{fig:mainchem-curve}.
Although there are 171 (world, seed) combinations, the baseline
\SciLab did not submit a valid law for each combination at all budgets.
We therefore restrict all four methods to the same 115
(world, seed) cells evaluable by corrected \SciLab at every budget.
These span 54 worlds, with unequal numbers of retained replicates.
We average seed errors arithmetically within each world, then take the geometric
mean across worlds; 95\% intervals use 5,000 world-bootstrap replicates.
The curve measures prediction quality among evaluable runs;
invalid submissions count as failures in the benchmark pass rate.
At $B=20$ and $25$, MDA's paired nMSLE ratios to corrected \SciLab are
$0.00624$ (95\% CI $0.00263$--$0.0158$) and
$0.00351$ ($0.00152$--$0.00830$), respectively.
The all-cell pass rates are $74.9\%$ and $78.4\%$ for MDA versus $15.8\%$
at both budgets for \SciLab, retaining all 171 cells per budget.
We see that MDA is significantly more sample efficient
than \SciLab and the ICL baselines.
ICL with its own LLM-selected designs slightly outperforms ICL on MDA's
history: designs that distinguish executable models need not be the most
useful examples for an in-context predictor. This does not contradict the
active-versus-random benefit for MDA itself.

\begin{figure}[t]
\centering
\includegraphics[width=\textwidth]{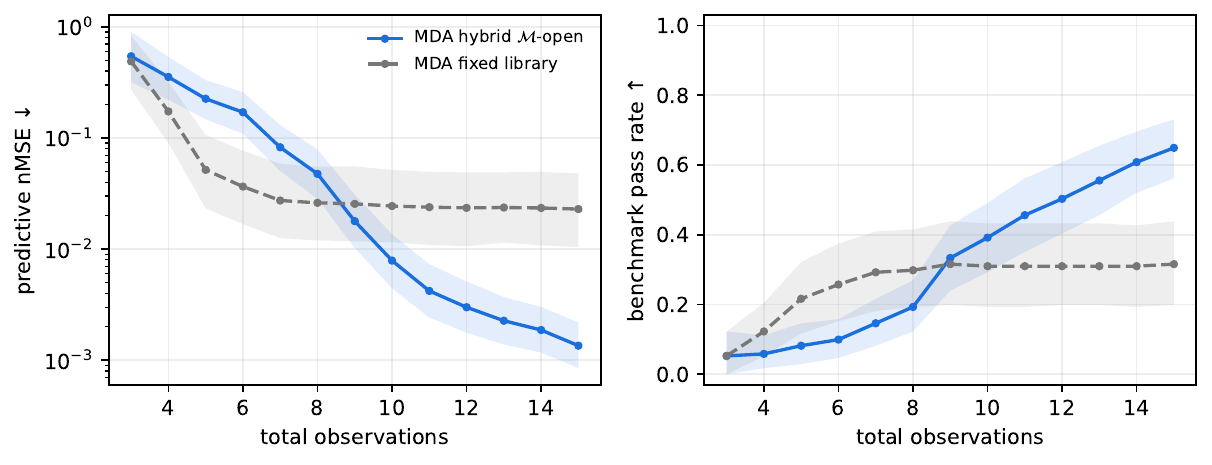}
\caption{\textbf{Benefits of LLM-augmented model search.}
Both arms use the BIC backend, CMA-ES maximization of $V(a)$, identical three-point passive design, and the
same 57 worlds and three replicate IDs; the grey arm is restricted to the nine executable seed laws and makes
no LLM calls, whereas the blue arm permits LLM initialization and residual-directed expansion. At $B=15$,
hybrid minus fixed pass rate is $+33.3$ points over all worlds (95\% CI $+22.8$ to $+43.9$) and $+39.6$ points
over the 48 out-of-library worlds ($+27.8$ to $+51.4$). The corresponding hybrid/fixed nMSE ratios are $0.059$
($0.032$--$0.109$) and $0.0356$ ($0.0185$--$0.0681$). On the nine in-library worlds both pass every cell.
Intervals average the three replicates within world and bootstrap worlds.}
\label{fig:chem-bic-fixed}
\end{figure}

\paragraph{Benefits of $\mathcal M$-open.}
\Cref{fig:chem-bic-fixed} shows the benefits of the $\mathcal M$-open setting:
we compare using a fixed set of 9 possible mechanisms vs
allowing the agent to grow the hypothesis space using the LLM.
On the left we plot test nMSE, and on the right, we plot
pass rate.
We see the $\mathcal M$-open setting is much better.
This measures the combined benefit of LLM initialization and later expansion,
rather than isolating later expansion alone.

\begin{figure}[t]
\centering
\includegraphics[width=\textwidth]{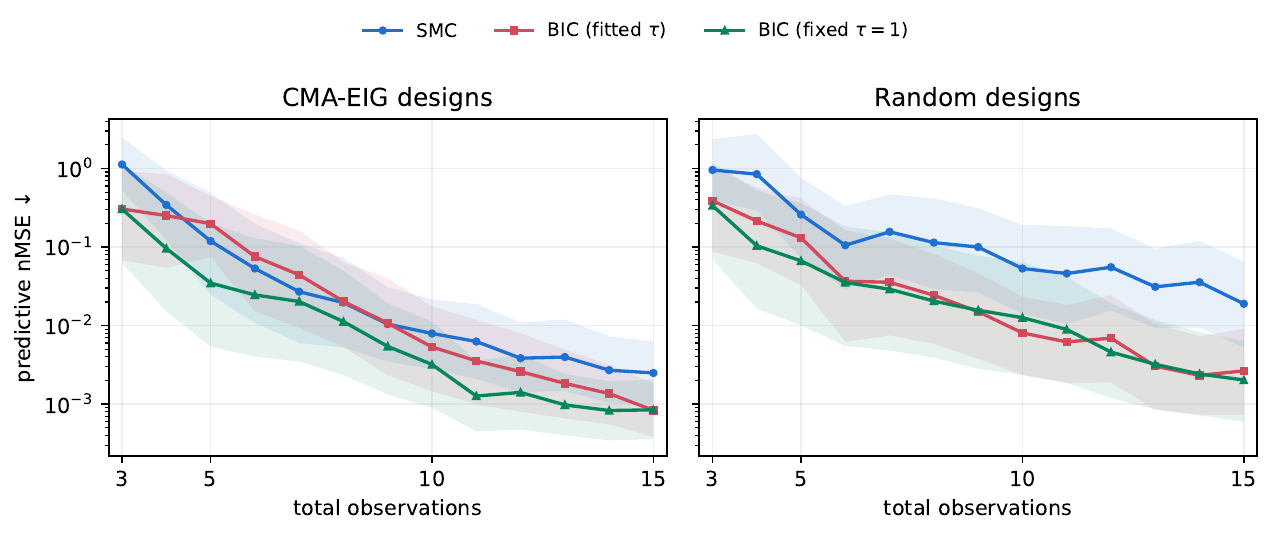}
\caption{\textbf{ChemBench inference configurations $\times$ experiment design.}
Twelve predeclared worlds and three seeds per world; all six arms receive the same three passive observations
and then collect 12 active observations. Both panels show held-out predictive nMSE:
left, designs optimize $V(a)$ with CMA-ES; right, random designs.
Colors identify the same three inference methods in both panels, with shared axes.
BIC denotes
bounded multistart point fitting with weights proportional to $e^{-\mathrm{BIC}/2}$; SMC uses 100 parameter
particles per structure. Lines are geometric means over worlds and replicates;
bands are 95\% world-bootstrap intervals. At 15 observations, CMA-ES versus random
reduces nMSE by factors $0.314$ ($0.127$--$0.698$) under BIC and $0.132$ ($0.061$--$0.287$) under SMC; the
BIC+CMA versus SMC+CMA nMSE ratio is $0.334$ ($0.163$--$0.694$).
These BIC contrasts use the original fitted-scale
arms. The added green arms fix $\tau=1$, replay their paired fitted-scale arm's
initial LLM proposal, and subsequently run their own inference, expansion, and
design loops. Random arms share the same design histories. The budget-geometric
comparison integrates log nMSE over budgets with the trapezoidal rule before
forming paired world-bootstrap ratios.}
\label{fig:chem-controls}
\end{figure}

\paragraph{Benefits of  EIG vs random designs.}
\Cref{fig:chem-controls} shows the benefits of active experiment design
(left panel, optimizing $V(a)$ using CMA) vs using random experiment design
(right panel).
We also compare using tempered SMC to estimate the evidence
(blue lines) vs BIC with a fitted residual scale (red) or fixed scale (green).
These configurations also differ in parameter treatment and model prior:
the SMC arm uses $p(m)\propto e^{-2C_m}$. Thus the curves do not isolate
evidence approximation alone. The BIC configurations predict better here,
with the fixed $\tau$ variant in \cref{eq:BICfixed}
slightly better (in the EIG case) than the profiled $\tau$ variant in \cref{eq:BIC}
in this cohort.

\eat{
Based on these experiments, we make
CMA+BIC our default method for \CHEMbench.
Fixing the scale improves the early CMA learning curve, but the two BIC
variants reach similar final performance. Across budgets 3--15, the
fixed/fitted-scale ratio of budget-geometric nMSE is $0.466$
(95\% world-bootstrap interval $0.284$--$0.746$); at 15 observations it is
$1.015$ ($0.721$--$1.459$). Under random design neither contrast is resolved.
The headline experiment retains its original fitted-scale configuration.
}

\clearpage
\section{\glucose: a partially observed ODE domain}
\label{app:glucose}

\subsection{Domain and experimental protocol}
\glucose,
which is available at \redacted{\url{https://github.com/murphyk/glucosebench}},
is our modification of
\texttt{simglucose}\footnote{
\url{https://github.com/jxx123/simglucose},
simulator revision \texttt{2e897f1}.
},
a Python implementation of the UVA/Padova type-1 diabetes model
from \citep{kovatchev2009insilico},
whose original simulator was accepted by the FDA for preclinical testing of insulin-
control algorithms in place of animal trials
\citep{Cobelli2023}.
We modified the model to allow partial noisy observations of both
glucose (every 5 minutes) and insulin (every 15 minutes),
for up to 6 hours.
Meals and injected insulin drive hidden physiological compartments, while a
continuous glucose monitor (CGM) gives noisy glucose measurements.
The 30 virtual profiles comprise ten children, ten adolescents, and ten adults.
Physiology is deterministic conditional on the profile and inputs; uncertainty
comes from observation noise and model misspecification, not stochastic
physiological transitions.

\paragraph{Patient-specific training and held-out interventions.}
Each profile has its own training episodes, candidate archive, and fitted
parameters. Its test set contains new intervention episodes from that same
profile, not different patients. No data are pooled across patients. Learners
receive observations, masks, measurement units, and delivered inputs, including
basal insulin, but not age, patient ID, body weight, simulator parameters, or
latent states.
Thus the task
measures personalized intervention generalization, not unseen-patient transfer.

Each episode lasts six hours. CGM is sampled every five minutes (73 readings,
including time zero), with temporally correlated native sensor noise.
We add a plasma-insulin assay every 15 minutes (25 readings): the oracle
converts its private insulin mass to concentration and adds independent
Gaussian noise with standard deviation $10$\,pmol/L. This is an added
observation model, not a native sensor. Missing assays have an explicit mask
and contribute nothing to fitting or evaluation.

\paragraph{True data-generating process.}
The generator uses a private
13-dimensional patient state:
\(
x=(Q_{sto1},Q_{sto2},Q_{gut},G_p,G_t,I_p,X,I_1,I_d,I_l,I_{sc1},I_{sc2},G_{sc})
\). Conditional on a virtual-patient parameter vector and the known meal and
insulin inputs, its continuous-time dynamics are deterministic (the equations
below describe the nonnegative physiological domain):
\begin{equation}
\begin{aligned}
\dot Q_{sto1}&=-k_{\max}Q_{sto1}+D,&
\dot Q_{sto2}&=k_{\max}Q_{sto1}-k_{gut}Q_{sto2},\\
\dot Q_{gut}&=k_{gut}Q_{sto2}-k_{abs}Q_{gut},&
\dot G_p&=EGP+R_a-F_{snc}-E-k_1G_p+k_2G_t,\\
\dot G_t&=-U_{id}+k_1G_p-k_2G_t,&
\dot I_p&=-(m_2+m_4)I_p+m_1I_l+k_{a1}I_{sc1}+k_{a2}I_{sc2},\\
\dot X&=-p_{2U}X+p_{2U}(I-I_b),&
\dot I_1&=-k_i(I_1-I),\\
\dot I_d&=-k_i(I_d-I_1),&
\dot I_l&=-(m_1+m_{30})I_l+m_2I_p,\\
\dot I_{sc1}&=U_I-(k_{a1}+k_d)I_{sc1},&
\dot I_{sc2}&=k_dI_{sc1}-k_{a2}I_{sc2},\\
\dot G_{sc}&=-k_{sc}G_{sc}+k_{sc}G_p.&
\end{aligned}
\label{eq:glucose-true-ode}
\end{equation}
All coefficients in Eq.~\eqref{eq:glucose-true-ode} are fixed by the private
virtual-patient profile. The nonlinear fluxes are
\begin{equation}
\begin{gathered}
Q_{sto}=Q_{sto1}+Q_{sto2},\\
k_{gut}=k_{\min}+\frac{k_{\max}-k_{\min}}{2}
 \left[\tanh\!\{\alpha(Q_{sto}-b\bar D)\}
 -\tanh\!\{\gamma(Q_{sto}-d\bar D)\}+2\right],\\
\alpha=\frac{5}{2\bar D(1-b)},\qquad
\gamma=\frac{5}{2\bar Dd},\qquad
R_a=\frac{f k_{abs}Q_{gut}}{BW},\\
EGP=[k_{p1}-k_{p2}G_p-k_{p3}I_d]_+,\qquad
E=k_{e1}[G_p-k_{e2}]_+,\\
I=I_p/V_i,\qquad
U_{id}=\frac{(V_{m0}+V_{mx}X)G_t}{K_{m0}+G_t}.
\end{gathered}
\label{eq:glucose-true-fluxes}
\end{equation}
When no meal is active and $\bar D=0$, the implementation uses
$k_{gut}=k_{\max}$ rather than evaluating the displayed ratios.
Here $D(\tau)=1000m(\tau)$ converts grams of ingested carbohydrate to mg,
$\bar D$ is the current meal's cumulative consumed load plus pre-meal stomach
content, and $U_I(\tau)=6000\{u_b(\tau)+a(\tau)\}/BW$ converts basal and bolus
insulin to the simulator's per-kilogram units. The corpus adapter delivers a
meal at at most 5 g/min and spreads each bolus over five minutes. The
pinned implementation uses adaptive Dormand--Prince integration with inputs
updated every minute, and we sample every $\Delta=5$ minutes.

\paragraph{Our extension: the two-channel observation model.}
At sample time $t_j=j\Delta$, our public observation process is
\begin{equation}
\begin{aligned}
y^C_j&=\operatorname{clip}_{[L,U]}
       \left(G_{sc}(t_j)/V_g+\epsilon^C_j\right),
&M^C_j&=1,\\
y^I_j&=I_p(t_j)/V_i+\epsilon^I_j,
&M^I_j&=\mathbb I\{t_j\bmod15\text{ min}=0\},\\
\epsilon^I_j&\stackrel{\mathrm{iid}}{\sim}\mathcal N(0,10^2)
 &&\text{pmol/L}.
\end{aligned}
\label{eq:glucose-true-observations}
\end{equation}
The GuardianRT CGM error $\epsilon^C_j$ is the simulator's colored,
Johnson--SU-transformed process: it evolves at 15-minute knots and is cubically
interpolated to the five-minute grid before sensor-range clipping. Plasma
insulin is our added noisy assay, without rounding, and unobserved
between 15-minute assay times.

\paragraph{Input sequences.}
\label{app:glucose_design}
An experiment selects a template for an entire input sequence, rather than
adapting inputs within the episode. Background basal insulin is fixed at the
profile's prescribed simulator rate and is public to all methods.
The four design controls $a_t^{\text{meal-amt}}$, $a_t^{\text{meal-time}}$,
$a_t^{\text{bolus}}$, and $a_t^{\text{bolus-delay}}$ specify a schedule chosen
once per experiment. A scheduler converts them into delivery rates.
The menu crosses
\begin{align*}
 \text{meal amount}&\in\{20,40,60\}\ {\rm g}, &
 \text{meal time}&\in\{60,120\}\ {\rm min},\\
 \text{bolus}&\in\{0,0.5,1,1.5\}\ {\rm U}, &
 \text{bolus delay}&\in\{0,30\}\ {\rm min}.
\end{align*}
U denotes units of insulin activity; U/min is a delivery rate. Delay is
relative to meal time. Meals are consumed at $5$\,g/min and boluses delivered
over five minutes; we record actual delivery after pump discretization.
There are 48 nominal templates but 42 distinct interventions, since delay
does not matter for zero bolus. Sampling is without replacement over indices,
so equivalent zero-bolus interventions can recur.

\paragraph{Training set.}
All methods start with the same two passive episodes: a 30-g meal at minute
60 with a 1-U bolus at minute 75, and a 45-g meal at minute 90 with a 1.5-U
bolus at minute 90. Four additional experiments are selected; evaluation
budgets are $B_\expt\in\{2,3,4,6\}$ total episodes.

\paragraph{Test set.}
Each patient has twelve frozen held-out interventions, generated by Latin
hypercube sampling over meal amount 20--60\,g, time 60--120\,min, bolus
0--1.5\,U, and delay 0--30\,min (times rounded to five minutes). These include
inputs between training-menu levels.

\paragraph{Evaluation.}
For each patient, we use the trajectory nMSE in \cref{eq:lossTimeAvg},
with one scale per channel, $s_c$,
defined by the empirical variance of observed values in the two shared
initial training episodes $\D_0$:
\begin{equation}
  \begin{split}
     L_c&=\frac1{N_c s_c^2}\sum_q\sum_{t>t_c}
 o_{qtc}(\widehat y_{qtc}-y_{qtc})^2 \\
 N_c&=\sum_q\sum_{t>t_c}o_{qtc},\qquad
 s_c^2=\max\!\left\{\frac1{|\mathcal Y_{0c}|}
 \sum_{y\in\mathcal Y_{0c}}(y-\bar y_{0c})^2,1\right\}.
 \end{split}
 \label{eq:glucose_nmse}
\end{equation}
Here $\mathcal Y_{0c}$ contains all observed channel-$c$ values in $\D_0$,
$\bar y_{0c}$ is their mean, and $o_{qtc}$ is the observation mask.
The scale floor is one squared
native unit per observation. We normalize channels separately rather than
adding errors in incompatible units, and take geometric means over profiles.
This scale is fixed across methods, budgets and forecast cuts and uses no
test outcomes. Our main figure is the CGM, $t_c=45$ slice of the multi-cut
comparison below, with the same fixed paired-query set.

\subsection{MDA}

In this section, we describe how we applied MDA to this domain.

\paragraph{Models.}
\label{app:glucose_models}
The backend supplies four reduced compartmental ODEs before seeing $\D_0$;
their parameters are then fitted to $\D_0$. These comprise one-stage meal/insulin
absorption with linear or glucose-dependent insulin action, two-stage meal
absorption, and a delayed-CGM-readout model. The four-state, two-stage
meal model is shown on the left of \cref{fig:glucose_model_growth},
and corresponds to these equations:
\begin{align}
 \dot z^{q_1} &= u_{\rm meal}-z^{q_1}/\tau_{q_1},\qquad
 \dot z^{q_2}=(z^{q_1}-z^{q_2})/\tau_{q_2}, \nonumber\\
 \dot z^i &= u_{\rm bolus}-z^i/\tau_i,\qquad
 \dot z^g=k_g(g_b c_0-z^g)+\alpha z^{q_2}-\beta z^i z^g/g_b .
 \label{eq:glucose_reduced_ode}
\end{align}
Here $c_0$ is a fitted equilibrium multiplier, while $z^{q_1},z^{q_2},z^i$
are effective input-memory compartments.
These seed ODEs provide mechanistic inductive bias not supplied as executable
models to ICL; the system comparison does not isolate the contribution of LLM births.
$I_b$ is the average insulin concentration over the observed 45-minute
prefixes of the training episodes,
and $g_b$ is the average CGM value.
The initial
glucose mean is $g_b$, and input-memory means are zero;
the hidden states are then updated by Bayesian filtering (see below).

The conditional observation means are $\widehat y_t^{\rm CGM}=z_t^g$ and
$\widehat y_t^{\rm insulin}=I_b+\lambda z_t^i$, with fitted positive gain
$\lambda$.
We fit an AR(1) model to CGM residuals. For EKF prediction we represent it
using an extra real-valued state $e_t$:
\begin{equation}
 e_t=\rho e_{t-1}+\eta_t,\quad \eta_t\sim\mathcal N(0,\sigma_\eta^2),
 \qquad y_t^{\rm CGM}=h_{\rm CGM}(z_t)+e_t.
 \label{eq:glucose_sensor_state}
\end{equation}
The full latent state is thus  $(z_t,e_t)$.
Physiological coordinates follow the deterministic ODE with
no added process noise; only the sensor-error coordinate receives these
Gaussian innovations. Insulin assays retain their separate observation noise.

The LLM edits dynamics and observation maps using restricted arithmetic
expressions, including adding latent compartments. Initial models have three
or four states; proposals may have six states and twelve fitted parameters.
State names sometimes describe a role (e.g., \texttt{qfast}, \texttt{qslow},
\texttt{cgm}), but do not establish physiological identifiability.

\paragraph{Parameter estimation.}
Parameters are fitted by multistart L-BFGS-B using the joint working
likelihood: correlated Gaussian CGM residuals and independent Gaussian
insulin-assay errors. The CGM innovation variance is profiled out; insulin
noise is fixed at its declared value. BIC counts the ODE parameters plus
the fitted CGM correlation and innovation scale, using the number of observed
scalar targets as its sample size.

\paragraph{Predictive uncertainty.}
The fully Bayesian prediction integrates uncertainty in states, parameters,
and model structure, while retaining observation noise in the predictive
likelihood, as follows
(where we drop conditioning
on the known design and context for brevity):
\begin{equation}
 \begin{split}
 p(Y_{t_c+1:T}\mid\mathcal H,\D)=\sum_m\int\!\int
 &
p(\widetilde z_{t_c}\mid\mathcal H,m,\theta,\D) \,
p(Y_{t_c+1:T}\mid \widetilde z_{t_c},m,\theta)\,
 \\[-2pt]
 &\cdot p(\theta\mid m,\mathcal H,\D)\,
p(m\mid\mathcal H,\D)\,d\widetilde z_{t_c}\,d\theta.
 \end{split}
 \label{eq:glucose_full_predictive}
\end{equation}
Here $\mathcal H=\mathcal H_{t_c}$ is the query prefix and
$\widetilde z_t=(z_t,e_t)$ includes the sensor-error state.
For the displayed forecasts, the first factor,
$p(\widetilde z_{t_c}\mid\mathcal H,m,\theta,\D)$,
is obtained by approximate Bayesian filtering,
based on
the Extended Kalman Filter (\citealp{sarkka2023filtering}).
The second factor,
$p(Y_{t_c+1:T}\mid \widetilde z_{t_c},m,\theta)$,
propagates the filtered state into the future
(using the EKF without any evidence updates),
and converts this into predicted observations.
We approximate the third factor,
$p(\theta\mid m,\mathcal H, \D)$  by $p(\theta\mid m,\D)$,
and then further approximate this by
the point estimate $\delta(\theta - \hat{\theta}_m)$.
Finally we approximate the fourth factor, $p(m\mid\mathcal H,\D)$,
by $p(m\mid\D)$,  which is represented by a weighted
histogram, where the weights are derived from the BIC scores
of each model in the library.
We show some sample forecasts in \cref{fig:glucose_pilot_predictions}.

\paragraph{Experiment design.}
MDA maximizes BIC-weighted
between-model mean-prediction disagreement, normalized by observation-noise
variance and averaged over the two channels at their observation times.
Specifically, for a candidate input schedule $a$, let
$\mu_{m,t,c}(a;\hat\theta_m)$ denote model $m$'s ODE prediction for channel
$c$ at time $t$, using its fitted parameters, and let
$w_m\propto\exp(-\mathrm{BIC}_m/2)$ be normalized model weights. We score
\begin{equation}
 V(a)=\frac12
 \sum_{c\in\{\mathrm{CGM},\mathrm{insulin}\}}
 \frac{1}{|\mathcal T_c|}\sum_{t\in\mathcal T_c}
 \frac{\operatorname{Var}_{m\sim w}
       [\mu_{m,t,c}(a;\hat\theta_m)]}{\nu_c^2},
 \label{eq:glucose-model-disagreement}
\end{equation}
where $\mathcal T_c$ contains scheduled observation times after initialization.
The CGM scale is the model-weighted stationary variance of the fitted AR(1)
residuals, $\nu_{\mathrm{CGM}}^2=\max\{1,\sum_m w_m\sigma_m^2/(1-\rho_m^2)\}$;
the insulin scale is the declared assay-noise variance. These acquisition
scales differ from the evaluation normalization $s_c^2$.
This is a noise-normalized, trajectory-valued counterpart of
\cref{eq:chem-model-disagreement}, giving equal weight to the two channels.
Parameters are held at point estimates rather than integrated out, and
we maximize over unused menu indices rather than by CMA-ES (ties are broken
randomly). Note that this score is a model-disagreement surrogate for EIG, not exact EIG;
within-model predictive variance is not included in its numerator.

\subsection{Baselines}

\paragraph{ARX.}
\label{app:ARX}

As a baseline, we consider fitting a
linear autoregressive model with exogenous inputs
\citep{lutkepohl2005multiple}
for each observation channel separately:
\begin{equation}
 \widetilde y_t=b+\sum_{\ell=1}^{P}\alpha_\ell\widetilde y_{t-\ell}
       +\sum_{j=0}^{L-1}\beta_j^\top u_{t-1-j}+\epsilon_t.
 \label{eq:glucose_arx}
\end{equation}
Here $u_t$ contains delivered meal and bolus amounts in sampling interval $t$,
and $\widetilde y_t$ subtracts an episode baseline. The plotted control fits
separate CGM and insulin regressions on their five- and fifteen-minute grids,
centering by their respective 45-minute prefix means. There are no
cross-channel lags.

\paragraph{VARX.}
We also consider a joint VARX (vector autoregressive) model
which predicts
$\widetilde{\mathbf y}_t=
(\widetilde y_t^{\rm CGM},\widetilde y_t^{\rm insulin})^\top$
using
\begin{equation}
 \widetilde{\mathbf y}_t=\mathbf b+
 \sum_{\ell=1}^{P}A_\ell\widetilde{\mathbf y}_{t-\ell}
 +\sum_{j=0}^{L-1}B_j u_{t-1-j}+\boldsymbol\epsilon_t,
 \qquad A_\ell,B_j\in\mathbb R^{2\times2}.
 \label{eq:glucose_varx}
\end{equation}
Off-diagonal $A_\ell$ entries allow cross-channel prediction. Our implementation
uses the five-minute grid, centering each channel by its initial observation.
Missing insulin predictors are carried forward from the last observed assay,
never interpolated from future assays; only observed targets enter the loss:
\begin{equation}
 \sum_{e,t,c}o_{etc}(\widetilde y_{etc}-w_c^\top x_{et})^2
       +\lambda\sum_c\|S w_c\|_2^2.
 \label{eq:glucose_varx_fit}
\end{equation}
The vector $x_{et}$ contains an intercept, channel lags and input lags;
$S$ is diagonal training-feature RMS scaling floored at one. Ridge includes
the intercept.

In both cases, lag order and regularization use whole-training-episode
cross-validation. Forecasts recurse on predicted outputs after the cut.

\paragraph{SINDy.}
\label{app:glucosesindy}
SINDy (Sparse identification of nonlinear dynamics)
\citep{brunton2016sindy} identifies sparse combinations of candidate
functions; we use its controlled form, SINDYc \citep{brunton2016sindyc},
implemented with PySINDy's sequentially thresholded least squares.
The two state coordinates are observed CGM and insulin, with known meal and
bolus inputs; an optional pair of fixed 30-minute exponential input memories
represents delayed effects. No simulator states or fitted MDA mechanisms are
provided. This control reuses MDA's training episodes, not its acquisition rule.

We select degree-one or degree-two polynomial libraries, smoothing windows
of five or nine samples, input memories on/off, and thresholds
$\{0,0.005,0.02\}$ (ridge penalty $0.1$): 24 configurations.
Missing training insulin is linearly interpolated between assays; quadratic
Savitzky--Golay smoothing estimates derivatives on the five-minute grid.
Scaling and leave-one-episode-out selection use training data only, scoring
the two-channel suffix after a 45-minute prefix.
Before query evaluation, full-data refits are screened on training inputs
at all four cuts and prefixes perturbed by one training standard deviation.
We reject divergent predictions, excursions beyond the observed training
range plus six standard deviations, or scaled RMS changes above $0.02$
when halving the RK4 step from one minute. The first passing CV-ranked model
is retained. This is a finite-horizon safeguard, not a global stability proof.

At prediction time, available prefix observations reset their corresponding
coordinates; missing assays are propagated, not filled using future observations.
Cut zero uses training-only initial values. After the cut, forecasts are fully
recursive with no parameter refitting, clipping, or fallback.
All 120 fits and 5,760 forecasts completed; independent replay reproduced
the fits and predictions. Screening changed seven CV winners; 117 selected
models are linear and three quadratic.

\subsection{Results}

\begin{figure}[!ht]
 \centering
 \includegraphics[width=\linewidth]{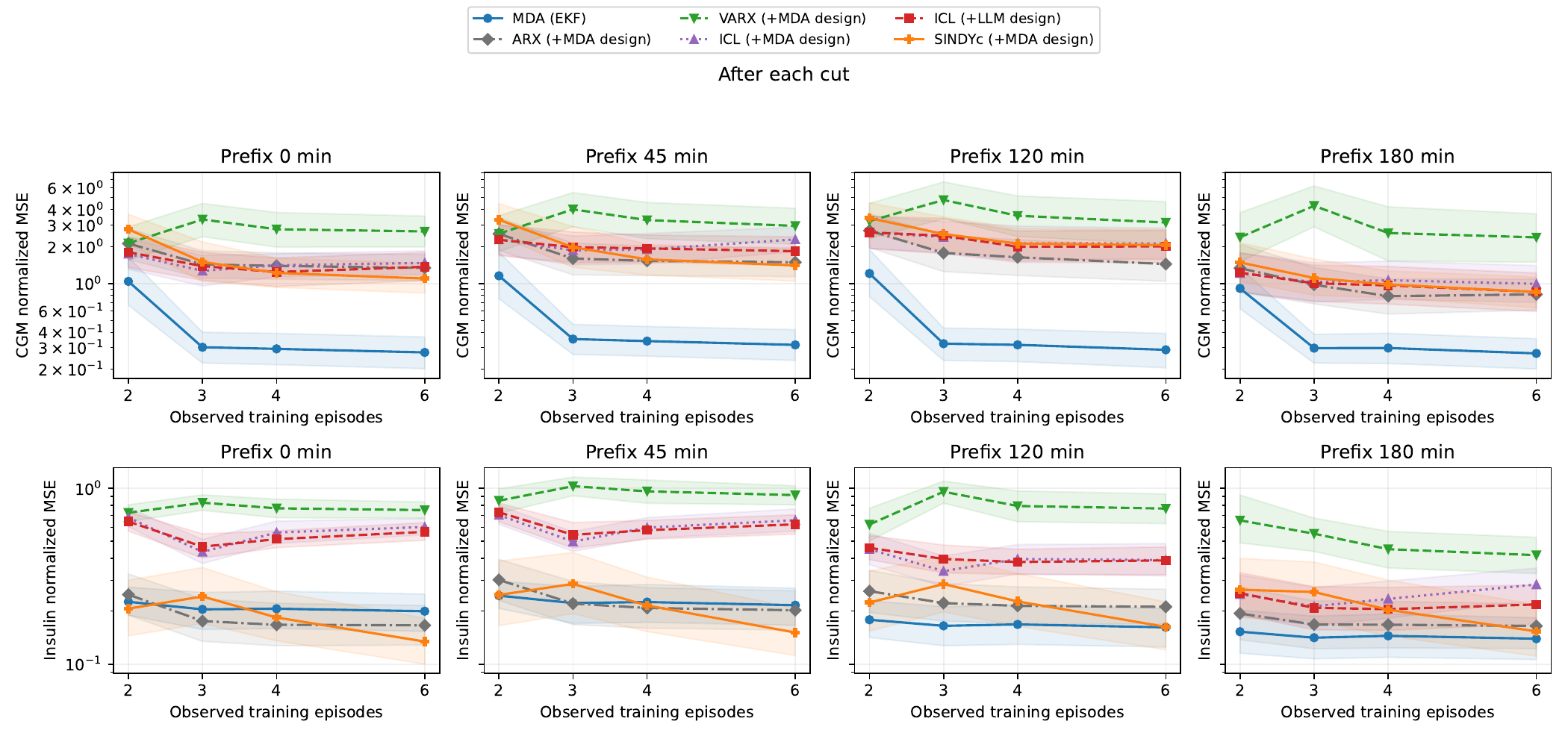}
 \caption{\textbf{Conditional forecasting at four prefix lengths.}
   CGM (top) and insulin (bottom), across all 30 patient profiles.
   (Each patient has one algorithmic replicate; bootstrap
intervals reflect variation across patients, not repeated LLM runs.)
 MDA uses EKF state estimation and BIC model averaging at fixed fitted
 parameters; at cut zero it propagates training-derived initial states
 without query measurements. ARX, VARX, SINDYc and ICL (+MDA design) share MDA's
 histories; ICL (+LLM design) uses its own archived histories. Curves are
 patient-level geometric means of training-variance-normalized MSE;
 bands are 95\% patient-bootstrap intervals. All curves use the same
 273/360 patient--query pairs valid for every method, budget and cut.
 Each column scores only observations after its cut, so columns change
 both conditioning information and forecast horizon.}
 \label{fig:glucose_multicut}
\end{figure}

\paragraph{Sample efficiency curves.}
In \cref{fig:glucose_main_curve} we show the prediction loss
for CGM (after $t_c=45$) vs number of training experiments
for the different methods. We see that MDA is substantially more
sample efficient.
\eat{
That figure uses exactly the CGM, $t_c=45$ results from
\cref{fig:glucose_multicut}, omitting VARX for space. All cuts use the same
EKF/BIC prediction procedure, normalization, and paired queries. We reuse
the training fits and acquisition histories across cuts; acquisition is not
rerun for each prefix length.
\eat{
Using the 359 queries valid at $t_c=45$ alone, rather than the 273-query
four-cut intersection, gives similar final CGM nMSE: $0.304$ for MDA,
$1.452$ for ARX, $2.292$ for ICL (+MDA design), and $1.913$ for
ICL (+LLM design). Thus the headline ordering is not an artifact of failures
at the other cuts.
}
}

In \cref{fig:glucose_multicut} we extend the analysis
by varying the cut point,
using  $t_c=0,45,120,180$ minutes,
and showing results for GCM prediction and insulin prediction.
We use the same training data across all runs,
and just change  the initial belief state,
$p(z_{t_c}\mid\mathcal H_{t_c},m,\theta,\D)$.
We plot prediction error for CGM in the top row and insulin in the bottom row,
and vary $t_c$ along the columns.
We see two main effects:
(1) for predicting GCM, MDA has a big lead over all other methods,
whereas for predicting insulin levels, MDA
ties with the simpler ARX and SINDYc baselines,
because this signal is easier to predict;
(2)  the amount of  history conditioning  does not make much difference
to GCM prediction, but for insulin prediction,
conditioning on $\geq 120$ minute prefix for this patient helps
most methods to various degrees.

\eat{
At six episodes and a 45-minute prefix, CGM nMSE is $0.316$ for MDA
versus $1.403$ for SINDYc; the paired SINDYc/MDA ratio is $4.44$
(95\% patient-bootstrap interval $[3.60,5.53]$).
For insulin the ordering reverses: $0.216$ versus $0.151$, with ratio
$0.70$ ($[0.58,0.85]$). SINDYc is therefore a useful numerical comparator,
not uniformly inferior to the mechanistic pipeline. MDA retains lower aggregate
CGM error at every cut; SINDYc's insulin advantage at cuts zero and 45 minutes
does not persist at the longer cuts.
Excluding the three smoke patients used during engineering gives the same
45-minute-cut conclusion on the remaining 27 patients (ratios $4.86$ for CGM
and $0.69$ for insulin). The all-360-query numerical comparison also preserves
this ordering; SINDYc introduces no additional complete-case exclusions.
}

\eat{
\paragraph{Numerical safeguards.}
Proposals are validated for state and parameter dimensions, finite ordered
bounds, and an allowed expression grammar before fitting. Rollouts abort on
nonfinite states or magnitudes above $10^7$; failed objective evaluations
receive a large finite penalty ($10^{20}$), and candidates whose fits are all
invalid are rejected. These checks limit numerical failures, not model
misspecification; they do not guarantee accurate integration of stiff dynamics.
}

\paragraph{Forecast ability.}

\begin{figure}[!ht]
 \centering
 \includegraphics[width=\linewidth]{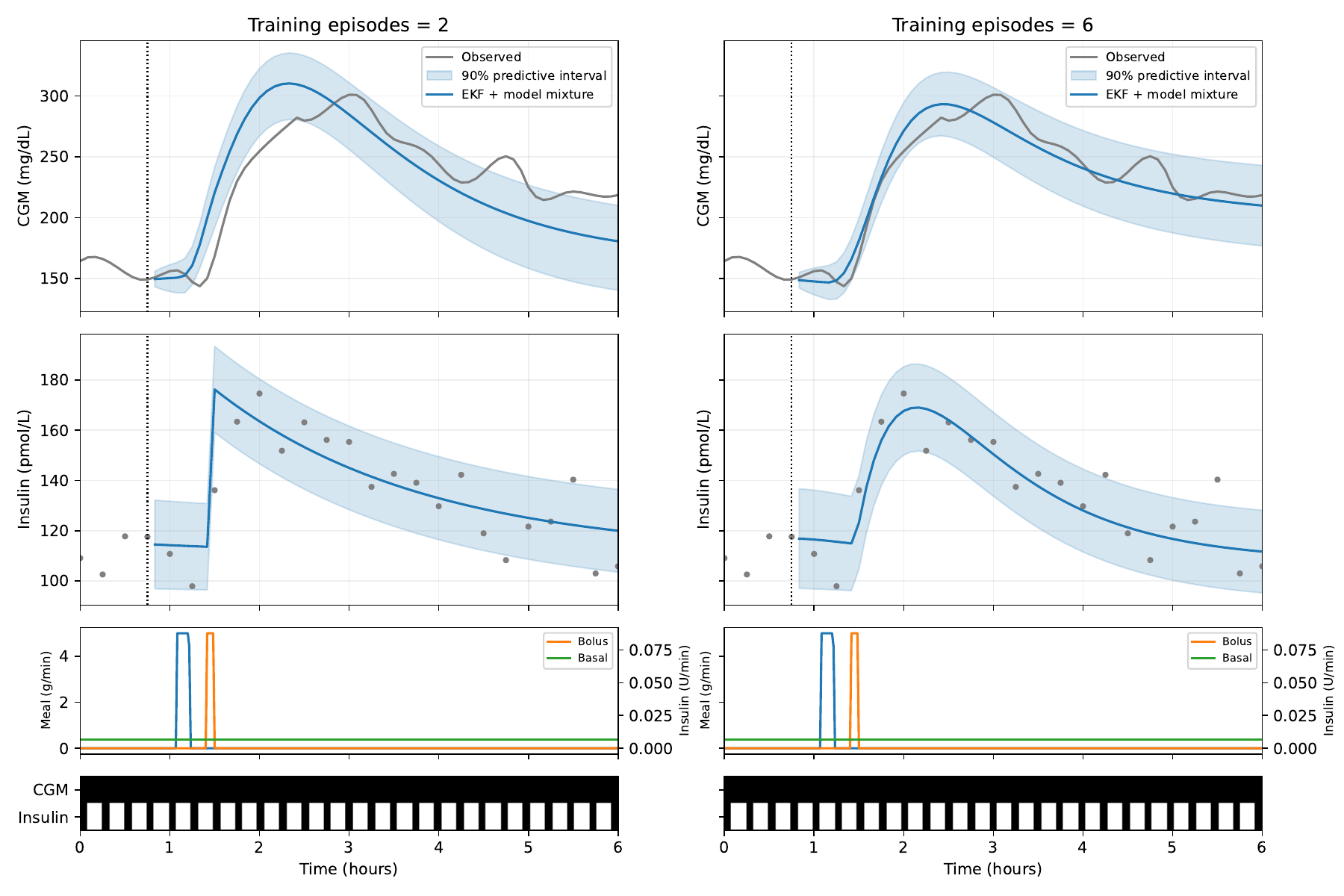}
 \caption{\textbf{Two-channel predictions before and after learning.}
   Sample forecasts for a single patient after  two versus six training episodes.
   Blue curves are
 EKF/BIC-mixture predictive means; gray traces/points are
 held-out CGM/insulin observations. The dotted line marks the 45-minute
 forecast cut. Lower rows show delivered inputs and the observation mask.
 Shading shows marginal 90\% predictive intervals, computed as quantiles
 of the Gaussian-component mixture, not a Gaussian fit to its moments.
 Intervals include filtered-state, observation-noise and between-model
 uncertainty, but \emph{not parameter uncertainty}. The EKF's approximate
 covariance initialization and positivity safeguard apply; nominal 90\%
 coverage is not a calibration claim.
 }
 \label{fig:glucose_pilot_predictions}
\end{figure}

\Cref{fig:glucose_pilot_predictions} gives an example of probabilistic
forecasting,
using an approximation of \cref{eq:glucose_full_predictive},
for a single patient and input sequence (lower rows),
using MDA's model mixture trained with 2 (left column)
or 6 (right column) trajectories.
The example illustrates changes in the forecast mean and uncertainty;
it does not establish improved variance calibration. Model disagreement
accounts for 38\% of the summed CGM predictive variance at two episodes,
but less than 0.001\% at six episodes, when one model has 99.81\% weight.
The remaining variance is within-model state and observation uncertainty.
We do not report calibration or proper-score comparisons here;
\cref{app:CRPS} discusses probabilistic scoring.

\begin{figure}[!ht]
 \centering
 \includegraphics[width=\linewidth]{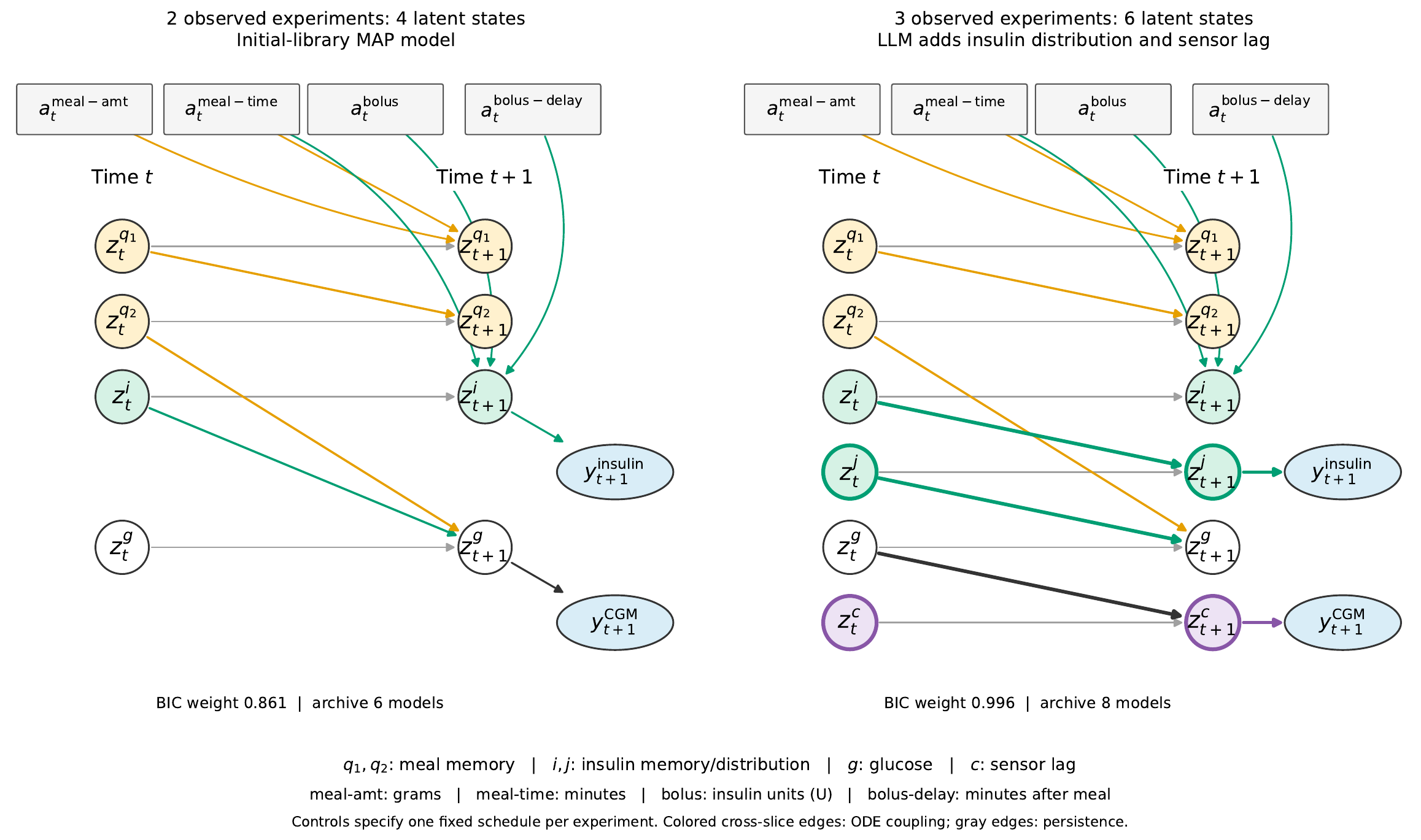}
 \caption{\textbf{A glucose model grows new latent mechanisms.}
   Illustration of the MAP model for a single patient
   after varying number of  training trajectories.
Left: four-state
 initial-library model at two episodes. Right: six-state LLM proposal born
 and selected at episode three. Bold outlines/edges mark added states and
 dependencies. Only next-slice observations are drawn. Colored cross-slice
 edges show ODE couplings in a local Euler rendering; gray edges show
 persistence. Exact finite-time integration can induce indirect dependencies.
 Orange denotes meal memory, green insulin memory, and purple sensor lag.
 Basal delivery is fixed and omitted.
 }
 \label{fig:glucose_model_growth}
\end{figure}

\paragraph{Model structure learning.}
\Cref{fig:glucose_model_growth} shows an example
of model expansion.  After two episodes the MAP model
has four states $(z^g,z^{q_1},z^{q_2},z^i)$. A newly proposed six-state model
selected after episode three adds insulin distribution and sensor lag:
\[
 \dot z^j=k_j(z^i-z^j),\qquad \dot z^c=(z^g-z^c)/\tau_c,\qquad
 \widehat y^{\rm CGM}=z^c,\qquad
 \widehat y^{\rm insulin}=I_b+\lambda z^j.
\]
Insulin affects glucose through $z^j$ instead of directly through $z^i$.

\clearpage
\section{\HHbench: further details}
\label{app:bio}
\label{app:HHdet}
\label{app:HHstoch}
\label{app:HH}

In this section, we introduce our new \HHbench benchmark,
which is available at
\redacted{\url{https://github.com/murphyk/neuronbench}}.
We also give a brief primer on  neuron electrophysiology and
Hodgkin--Huxley models, to make this section comprehensible to
a machine learning audience.

\subsection{Background}
\label{app:ephs}

\paragraph{Primer on neuron electrophysiology.}

\begin{figure}[!ht]
\centering
\includegraphics[width=0.86\textwidth]{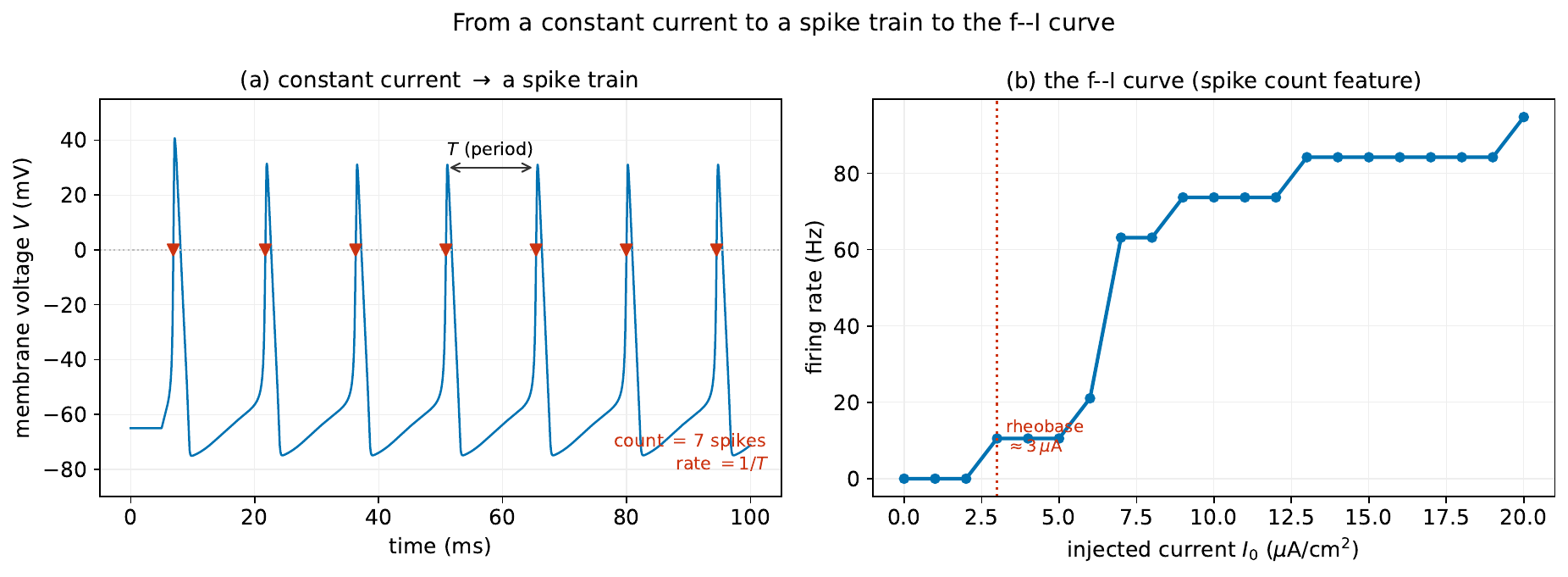}
\caption{\textbf{The f--I curve, and why we count spikes.} \emph{(a)} A constant supra-threshold current makes
the model fire a periodic \emph{spike train}; the readout is simply the \emph{spike count} (red markers) ---
or, per unit time, the firing rate $1/T$. \emph{(b)} Sweeping the injected current traces the \emph{f--I
curve} (firing rate vs.\ current): flat and zero below the \emph{rheobase} (the smallest current that fires,
red), then rising. This matches the intuitive ``how many spikes'' readout.}
\label{fig:hhfi}
\end{figure}

We can view a neuron as a device that turns an injected current into a
voltage trace. At rest the membrane voltage $V$ sits near $-65$\,mV. A
small (\emph{sub-threshold}) injected current depolarises $V$ a little
and it relaxes back --- a passive, RC-like response. A large enough
(\emph{supra-threshold}) current triggers an \emph{action potential}
or \emph{spike}: voltage-gated Na$^+$ channels open regeneratively,
$V$ shoots to ${\sim}{+}40$\,mV in under a millisecond, then K$^+$
channels open and pull it back down. Spikes are the neuron's output;
their \emph{count} (or rate) as a function of the injected-current
amplitude is the \emph{f--I curve} (frequency--current), the standard
input--output summary of a cell: see \cref{fig:hhfi}.

\begin{figure}[t]
\centering
\includegraphics[width=0.62\textwidth]{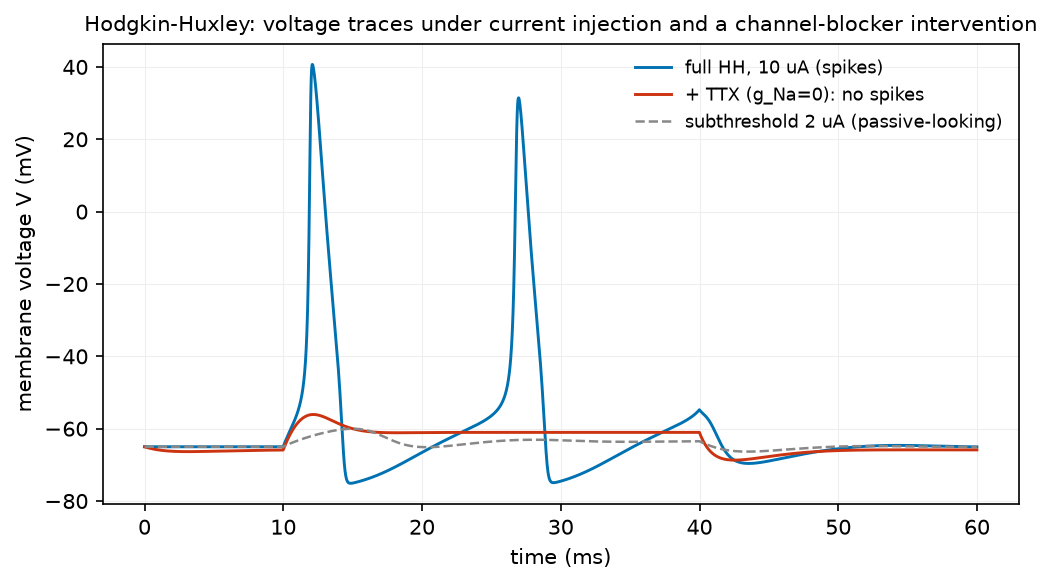}
\caption{\textbf{Example spike traces from a single neuron under different conditions.}
  Membrane voltage under current
injection: a supra-threshold step ($10\,\mu$A) elicits overshooting action potentials (blue); the sodium
blocker TTX ($g_{\mathrm{Na}}{=}0$, a $\dopo$ on the mechanism) abolishes them (red); a sub-threshold current
gives a passive response (grey).
}
\label{fig:hhtraces}
\end{figure}

Crucially, the behavior of the neuron depends on its inputs,
as illustrated
in \cref{fig:hhtraces}. Here we show the voltage over time,
under 3 different experimental conditions: a neuron
stimulated with a 10 $\mu A$ step signal, which generates repeating spikes (blue);
the same neuron
stimulated with a 2 $\mu A$ step signal,
which fails to trigger a response (dotted black);
and the neuron modified by applying TTX blocker and
then stimulated with a 10 $\mu A$ step signal,
which also fails to trigger a response (red line).
This illustrates why experiment design is critical in this domain.

\paragraph{Primer on Hodgkin-Huxley models.}
\label{app:HHprimer}

In 1952, Hodgkin and Huxley \citep{hodgkin1952quantitative} introduced a model to explain the above behavior,
based on their experiments with the giant axon of the squid.
Since then, the model has been generalized and is widely used
to mechanistically explain the spiking behavior of many kinds of neurons.
Hodgkin and Huxley received the 1963 Nobel Prize in Physiology / Medicine for this work.

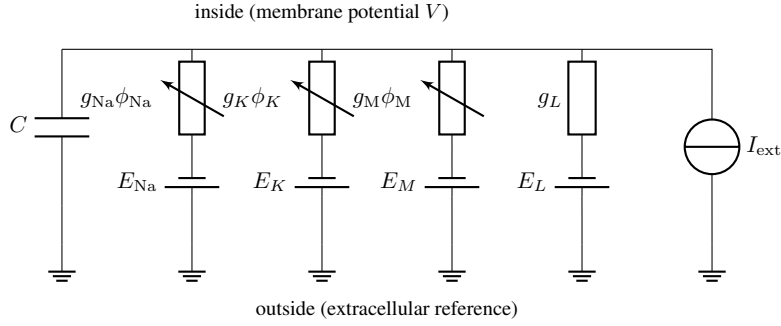
\begin{figure}[t]
\centering
\begin{circuitikz}[scale=0.86, transform shape, european]
  \draw (0,0) node[ground]{} -- (0,0.6);
  \draw (0,0.6) to[C=$C$] (0,3);                                      
  \foreach \x/\g/\e in {2/$g_{\mathrm{Na}}\phi_{\mathrm{Na}}$/$E_{\mathrm{Na}}$,
                        4/$g_K\phi_K$/$E_K$,
                        6/$g_{\mathrm{M}}\phi_{\mathrm{M}}$/$E_M$}{
     \draw (\x,0) node[ground]{} -- (\x,0.4) to[battery1, l=\e] (\x,1.5) to[vR, l=\g] (\x,3);
  }
  \draw (8,0) node[ground]{} -- (8,0.4) to[battery1, l=$E_L$] (8,1.5) to[R, l=$g_L$] (8,3);  
  \draw (0,3) -- (8,3);                                               
  \draw (10,0) node[ground]{} to[I, l_=$I_{\mathrm{ext}}$, invert] (10,3) -- (8,3);  
  \node at (4,3.55) {\footnotesize inside (membrane potential $V$)};
  \node at (5,-1.0) {\footnotesize outside (extracellular reference)};
\end{circuitikz}
\caption{\textbf{The Hodgkin--Huxley equivalent circuit (Eq.~\eqref{eqn:hh}).} The membrane is a capacitor
$C$; each ion channel is a branch with a variable conductance $g_c\phi_c$ (opening/closing gates $\phi_c$) in
series with a battery $E_c$ (the reversal potential). The injected current $I_{\mathrm{ext}}$ charges the
capacitor and flows through the open channels; a \emph{blocker} deletes a branch ($g_c{\to}0$). Which branches
are present is the \emph{structure}; the conductances $g_c$ are the \emph{parameters}.
}
\label{fig:hhcircuit}
\end{figure}

The model they came up with can be represented as an electric circuit, as shown in
\cref{fig:hhcircuit}. This example contains Na, K, M and L ion channels, but the generalized
model can contain different combinations of the 6 channels listed in \cref{tab:hhchan},
each of which have their own
parameters and dynamics. We can write
the generalized model as a set of nonlinear ODEs,
which follow from Kirchoff's current law:
\begin{align}
  C \frac{d V(t)}{dt}
  &= I_{\mathrm{ext}}(t) \;-\; \sum_{c\in\mathcal C} I_c(t) \label{eqn:hh}
  \\
  I_c(t) &= g_c \phi_c(t) (V(t) - E_c) \\
  \phi_c(t) &= m_c^{p_c}(t)  \; n_c^{q_c}(t) \;  h_c^{r_c}(t) \\
  \frac{d x_c(t)}{dt} &= \frac{T_{x,c}^{\infty}(V(t))-x_c(t)}{\tau_{x,c}(V(t))},
  \; x \in \{m,n,h\}
  \label{eq:HH}
  \end{align}
Here $C$ is the capacitance, $V(t)$ is the voltage,  $I_c(t)$ is the current
for channel $c$,  $\mathcal C$ is the set of channels associated with this neuron,
and $\phi_c(t)$ is the fraction of the channel that is open.
Thus
the current in the channel is given by
$I_c=g_c\,\phi_c\,(V-E_c)$: (maximal conductance) $\times$ (fraction open) $\times$ (driving
force).
The fraction open $\phi_c$ (which changes over time) is
based on a product of gating terms --- denoted by $m_c$, $n_c$ and $h_c$ ---
each raised to an integer power ($p_c,q_c,r_c$; how many independent gates the
channel has): see \cref{tab:hhgates} for a list of gates,
and \cref{tab:hhchan} for a list of channels that uses these gates.
Each  such gating term $x_c$
relaxes towards a voltage-dependent target $T_{x,c}^{\infty}(V)$ with its own time constant
$\tau_{x,c}(V)$ (fast for activation, slow for inactivation),
given by
\begin{align}
  T_{x,c}^{\infty}(V) &=\frac{\alpha_x(V)}{\alpha_x(V) + \beta_x(V)} \\
  \tau_{x,c}(V) &=\frac{1}{\alpha_x(V) + \beta_x(V)}
  \label{eq:HHtarget}
\end{align}
where expressions for $\alpha_x$ and $\beta_x$ can be found
at \url{https://en.wikipedia.org/wiki/Hodgkin-Huxley\_model}.
As an example,
the classic spiker is the following three-channel model
\begin{align}
  C\dot V = I_{\mathrm{ext}}
  - g_{\mathrm{Na}}m_{\mathrm{Na}}^3 h_{\mathrm{Na}}\,(V{-}E_{\mathrm{Na}}) - g_K n_K^4\,(V{-}E_K) - g_L\,(V{-}E_L)
\end{align}
Here the Na$^+$ channel carries an activation gate $m_{\mathrm{Na}}$ (cubed) and an inactivation gate
$h_{\mathrm{Na}}$, and the K$^+$ channel a single activation gate $n_K$ (to the fourth).
The names $m,n,h$ are historical: what actually distinguishes a gate is its
target curve $T_{x,c}^{\infty}(V)$ (whether it \emph{opens} or \emph{closes} as $V$ rises) and its time
constant $\tau_{x,c}(V)$. It is the \emph{separation of timescales} --- fast $m_{\mathrm{Na}}$ activation
admitting Na$^+$ for the upstroke, before the slower $h_{\mathrm{Na}}$ inactivation shuts it off and the
slower $n_K$ activation repolarises --- that makes the spike a transient, regenerative event.

\begin{table}[t]
\centering\small
\begin{tabular}{@{}lllll@{}}
\toprule
gate & channel & role & target $T_{x,c}^{\infty}(V)$ & speed \\
\midrule
$m_{\mathrm{Na}}$ & Na$^+$ & activation   & rises with depolarisation          & fast \\
$h_{\mathrm{Na}}$ & Na$^+$ & inactivation & \emph{falls} with depolarisation   & slow \\
$n_K$             & K$^+$  & activation   & rises with depolarisation          & slow \\
\bottomrule
\end{tabular}
\caption{\textbf{The three classic Hodgkin--Huxley gates.} \emph{Activation} gates ($m_{\mathrm{Na}},n_K$)
open as the cell depolarises; the \emph{inactivation} gate ($h_{\mathrm{Na}}$) closes. Each relaxes to a
voltage-dependent target $T_{x,c}^{\infty}(V)$ with its own time constant $\tau_{x,c}(V)$; the fast$/$slow
separation between $m_{\mathrm{Na}}$ and $\{h_{\mathrm{Na}},n_K\}$ is what generates and terminates the spike.
Other channels (Table~\ref{tab:hhchan}) carry their own gates $m_c,n_c,h_c$ with the same form but different
half-voltages and kinetics.}
\label{tab:hhgates}
\end{table}

\paragraph{Stochastic version of Hodgkin-Huxley.}
\label{app:hhstochastic}
To create a stochastic neuron, we add finite-$\Nnoise$ channel noise via the Fox--Lu
diffusion approximation \citep{fox1994}, with the channel count $\Nnoise$ tuning the
intrinsic noise from near-deterministic ($\Nnoise\!\to\!\infty$) to strongly
stochastic.

In more detail,
each gate $x_c$ is really an ensemble of $\Nnoise$ two-state ion channels,
each switching open\,$\leftrightarrow$\,closed
as a continuous-time Markov chain with the voltage-dependent rates $\alpha_x(V),\beta_x(V)$ of
\cref{eq:HHtarget}; the deterministic HH gating ODE is the $\Nnoise\!\to\!\infty$ mean-field limit of the open
fraction. The \emph{Fox--Lu} diffusion approximation \citep{fox1994} keeps finite $\Nnoise$ by replacing that
mean field with a Langevin (stochastic differential) equation --- the deterministic drift plus a Gaussian
channel-noise term whose variance scales as $1/\Nnoise$:
\begin{equation}
d x_c = \big[\alpha_x(V)(1-x_c) - \beta_x(V)\,x_c\big]\,dt
\;+\; \sqrt{\tfrac{\alpha_x(V)(1-x_c) + \beta_x(V)\,x_c}{\Nnoise}}\;\,dW_t,
\qquad x\in\{m,n,h\},
\label{eq:foxlu}
\end{equation}
with $dW_t$ an independent Wiener increment per gate. The diffusion coefficient is the sum of the two
transition fluxes divided by $\Nnoise$ (the system-size / $\Omega$-expansion correction to the channel master
equation), so more channels means smaller fluctuations and $\Nnoise\!\to\!\infty$ recovers the deterministic gate.
We integrate \cref{eq:foxlu} by Euler--Maruyama and substitute the noisy gates into the membrane equation
\eqref{eqn:hh}, making $\Nnoise$ a single knob from near-deterministic to strongly stochastic. Fox--Lu is the
standard cheap channel-noise model; see \citet{goldwyn2011} for how it compares to exact Markov-chain channel
simulation.
We sweep a
  \emph{noise ladder} $\Nnoise\in\{50,100,300,1000,3000\}$ from strongly stochastic to near-deterministic; unless
  noted we use $\Nnoise{=}100$ (fairly noisy), and $\Nnoise\!\to\!\infty$ recovers the deterministic benchmark.

\paragraph{Levels of abstraction.}
It is worth noting that HH is only one point on a spectrum of models at
different levels of abstraction. There are more detailed stochastic
models that capture individual cellular responses at a more granular level.
There are also simplified models, such as the
two-variable FitzHugh--Nagumo model,
and the leaky integrate-and-fire model.
Finally, if we set the
membrane time constant to zero and binarise the output,
we get the  McCulloch--Pitts unit \citep{mcculloch1943},
$y=\phi(\sum_i w_i x_i-b)$, which is the basis of artificial neural networks.
So there is no single ``true model''.
Instead, scientists seek  the \emph{coarsest
valid causal abstraction}  that is sufficient
for the things they want to understand or predict
\citep{beckers2019,rubenstein2017}.

\begin{table}[t]
\centering\footnotesize
\setlength{\tabcolsep}{4pt}
\begin{tabular}{@{}l l l p{2.3cm} l@{}}
\toprule
Channel & carries & current & role in the response & blocker \\
\midrule
Na$^+$ & sodium &
$I_{\mathrm{Na}}=g_{\mathrm{Na}}m_{\mathrm{Na}}^3 h_{\mathrm{Na}}\,(V{-}E_{\mathrm{Na}})$ &
regenerative spike \emph{upstroke} & tetrodotoxin (TTX) \\
K$^+$ (delayed rectifier) & potassium &
$I_K=g_K n_K^4\,(V{-}E_K)$ &
\emph{repolarises} the spike & TEA \\
Ca$^{2+}$ (high-threshold) &calcium &
$I_{\mathrm{Ca}}=g_{\mathrm{Ca}}m_{\mathrm{Ca}}^2 h_{\mathrm{Ca}}\,(V{-}E_{\mathrm{Ca}})$
& \emph{alternative}, slower spike upstroke & cadmium (Cd) \\
M-type K$^+$ & potassium &
$I_{\mathrm{M}}=g_{\mathrm{M}}\,m_{\mathrm{M}}\,(V{-}E_K)$
& slow; \emph{spike-frequency adaptation} & XE991 \\
A-type K$^+$ (transient) & potassium &
$I_{\mathrm{A}}=g_{\mathrm{A}}m_{\mathrm{A}}^p h_{\mathrm{A}}\,(V{-}E_K)$
& transient outward; \emph{delays} firing onset & 4-AP \\
leak & mixed &
$I_L=g_L\,(V{-}E_L)$
& sets the \emph{resting potential}; passive & --- \\
\bottomrule
\end{tabular}
\caption{\textbf{The voltage-gated ion channels --- the building blocks.} A \emph{blocker} is a drug that
removes one channel by setting its conductance $g_c{=}0$; these are the mechanism-level interventions
$\dopo(a)$ available on this rung (e.g.\ TTX abolishes Na$^+$-based spikes but not Ca$^{2+}$-based ones). The
parenthetical drug names identify available simulator interventions;
the current headline acquisition menu does not include blockers.}
\label{tab:hhchan}
\end{table}

\subsection{Our benchmark}
\label{app:HHbench}

We design a benchmark, \HHbench, by creating 6
``mystery neurons'', each composed of
a plain Na$+$K$+$leak spiker plus one extra
membrane mechanism, chosen from the list in \cref{tab:hhnovel}: five
are \emph{novel} mechanisms and the sixth is a recallable textbook
M-current control.
The five nonstandard mechanisms are intended to be difficult to distinguish
using the two weak initial probes. This is not a claim of equivalence under
every textbook stimulus or blocker. The evaluated agent selects among
specified stimulation protocols, rather than inventing arbitrary protocols.

\paragraph{Specification of the novel channels.}
Each novel channel
has roughly the same gated form as the textbook ones:
\begin{equation}
  I_Z = g_Z\, m_Z^{p}\, h_Z^{q}\,(V-E_Z), \quad
  T_{m,Z}^{\infty}(V)=\sigma\!\Big(\tfrac{V-V^m_{1/2}}{k_m}\Big), \quad
  T_{h,Z}^{\infty}(V)=\sigma\!\Big(\!-\tfrac{V-V^h_{1/2}}{k_h}\Big),
  \label{eq:hhnovel}
\end{equation}
with $\sigma(u)=1/(1+e^{-u})$ the logistic (Boltzmann) sigmoid,
half-voltages $V^m_{1/2},V^h_{1/2}$, nonzero signed slopes $k_m,k_h$, and fixed
time constants $\tau_m,\tau_h$: activation rises with $V$ and
inactivation falls, while a \emph{negative} activation slope $k_m{<}0$
instead makes the channel hyperpolarisation-activated (as for $I_h$),
and an inactivation half-voltage $V^h_{1/2}$ below rest makes it
\emph{de-inactivated by hyperpolarisation} (available only after a
hyperpolarising pre-pulse). This Boltzmann form is generic across the
\emph{novel} channels but is \emph{not} the textbook parameterisation
shown in  \cref{eq:HHtarget},
which are  monotonic curves of the same
qualitative shape but not identical logistic
sigmoids.

\begin{table}[t]
\centering\footnotesize
\setlength{\tabcolsep}{4pt}
\begin{tabular}{@{}l l l l p{4.3cm}@{}}
\toprule
Mechanism  & $(g_Z,E_Z)$ & activation$^{\dagger}$ & inact.$^{\dagger}$ & behavioural signature (revealing protocol) \\
\midrule
\zrebound ($I_Z$)              & $(4,\,{+}120)$   & $(-57,5,4,2)$      & $(-88,4,130)$ & spike-count collapse after a hyperpolarising conditioning pulse \\
\hsag ($I_h$)                  & $(5,\,{-}30)$    & $(-95,{-}5,140,1)$ & ---           & voltage sag $+$ post-inhibitory rebound on a hyperpolarising step \\
\nafatigue                     & ---              & \multicolumn{2}{l}{slow inactivation added to $h_{\mathrm{Na}}$} & use-dependent spike-count run-down over paired long pulses \\
\carebound ($I_{\mathrm{CaT}}$) & $(3.2,\,{+}120)$ & $(-54,6,2,2)$     & $(-87,4,22)$  & low-threshold rebound \emph{burst} on release from hyperpolarisation \\
\dtype ($I_D$)                 & $(9,\,{-}77)$    & $(-30,10,3,1)$    & $(-80,5,200)$ & delayed / suppressed firing after a hyperpolarising pre-pulse \\
\midrule
\textbookM ($I_M$)             & $(2.5,\,{-}77)$  & $(-35,10,60,1)$   & ---           & spike-frequency adaptation on a long step (recallable by name) \\
\bottomrule
\end{tabular}
\caption{\textbf{The six worlds of \HHbench}.
  Each current is added to a Na$+$K$+$leak spiker via \cref{eq:hhnovel}.
   $^{\dagger}$the activation$/$inactivation columns are the tuples
$(V_{1/2},k,\tau,\text{power})$ and $(V_{1/2},k,\tau)$ of \cref{eq:hhnovel}. Conductances $g_Z$ in
mS$/$cm$^2$; reversals $E_Z$, half-voltages and slopes in mV; time constants in ms; $p,q$ are gate powers
($q{=}1$ when an inactivation gate is present, else $0$). The final column
lists intended revealing protocols, not a proof that all other stimuli are
non-discriminating. \nafatigue adds no channel: it slows the inactivation of the
existing Na$^+$ gate $h_{\mathrm{Na}}$. The $I_M$ control is a standard non-inactivating K$^+$ current the
LLM \emph{can} name and probe. }
\label{tab:hhnovel}
\end{table}

\paragraph{The task}
The agent is told that it will be presented with some voltage trace data
from a neuron of unknown type,
and is asked to propose various candidate mechanisms
(the exact prompts are shown in
\cref{app:ephysprompt}). It is also given the menu of stimulation
protocols (\cref{tab:hhproto}) and channel blockers, and a fixed
experiment \emph{budget}. From a handful of designed experiments it
must \emph{(i)} \emph{propose its own} candidate mechanisms $m$ and
return a posterior $p(m\mid\data)$ over them, and \emph{(ii)} forecast
the cell's response to held-out interventions it never ran. The truth
is never revealed to the agent; it is used only for scoring.

\paragraph{Design space.}
\label{app:HHdesign}
At each step the agent controls the external current $I_{\mathrm{ext}}(t)=\inputs_t$.
Rather than specifying an (open loop) policy mapping  time step to current value,
we instead let the agent choose one of 90 possible policies.
The action library contains the nine protocols in \cref{tab:hhproto},
plus three 27-action grids: single pulses cross nine amplitudes
$\{4,6,8,11,12,14,16,18,20\}\,\mu\mathrm A$ with durations
$\{60,160,300\}$ ms; conditioning--test protocols cross conditioning amplitude
$\{-35,-20,+15\}\,\mu\mathrm A$, conditioning duration $\{60,150,270\}$ ms,
and test amplitude $\{6,12,18\}\,\mu\mathrm A$; paired pulses independently
choose the first and second amplitudes from $\{6,12,18\}\,\mu\mathrm A$ and the
zero-current gap from $\{30,90,180\}$ ms.  These families probe excitability and
adaptation, rebound and de-inactivation, and fatigue and recovery, respectively.

The benchmark also exposes a single channel blocker --- tetrodotoxin (TTX) zeroing
$g_{\mathrm{Na}}$, TEA zeroing $g_{\mathrm{K}}$, cadmium (Cd) zeroing $g_{\mathrm{Ca}}$, or none --- which
would give a larger action set. Blockers remain available in the released
benchmark and interactive app, but the experiment below searches only the
90 current-clamp protocols.

\paragraph{Experimental setup.}
Each run is seeded with a \emph{fixed} passive set $\D_0$ of two conventional experiments:
a brief $12\,\mu$A$/40$\,ms step and a weak $5\,\mu$A$/120$\,ms step. These probes avoid the
conditioning and paired-pulse families designed to expose the novel currents. The stochastic MDA arm then
selects eight further experiments using its model-disagreement/EIG surrogate.
For acquisition, we average deterministic HH feature predictions over the
weighted parameter particles and score between-model disagreement. Thus this
surrogate has no stochastic-rollout Monte Carlo noise, but it approximates,
rather than integrates, the SDE predictive mean. Likelihood evaluation and
held-out SDE predictions still use stochastic dynamics.
For the headline comparison, all four model-based arms and ICL (+MDA design)
replay the archived stochastic MDA histories. Only ICL (+LLM design)
selects its own eight actions. The earlier controlled inference ladder
in \cref{fig:hhbslladder} also uses shared histories.
Each observed trajectory contains exactly 20 sampled
voltage values, stratified over time with endpoints retained; the dense
simulator grid is not available as training data. Process noise uses
$N_{\rm noise}=100$ channels and observation noise has SD $2$ mV.
Oracle and filtering integration steps are $0.025$ and $0.05$ ms, respectively.

\eat{
We run only a \emph{small} budget
($N_a\le 5$): the original discrete menu is small ($9$ protocols, so $N_a{\le}7$ before it is exhausted), and each
\emph{stochastic} evaluation is an expensive SMC$^3$ rollout (the inner particle filter tracks
$\nparticlesLatent{=}250$ latent gating paths; see \cref{tab:smc} for the full SMC settings).  The current stochastic
comparisons still begin with two shared warm-ups, but use the expanded menu and include the external ICL
forecaster described in \cref{app:hhcurrentresults}.
}

\begin{table}[t]
\centering\footnotesize
\begin{tabular}{@{}rlll@{}}
\toprule
\# & protocol & segments $(\Delta t\,\text{ms},\,I\,\mu\text{A})$ & probes \\
\midrule
1 & brief step          & $(40,12)$                            & fast onset \\
2 & long step           & $(300,10)$                           & spike-frequency adaptation \\
3 & strong step         & $(120,18)$                           & high-rate firing \\
4 & weak step           & $(120,5)$                            & near-threshold f--I \\
\midrule
5 & hyperpol.\ conditioning $+$ test     & $(250,{-}30),(150,12)$      & de-inactivation / depol.\ block \\
6 & hyperpol.\ pre-pulse $+$ weak test   & $(250,{-}30),(120,0),(60,6)$ & rebound at low drive \\
7 & paired long pulses                   & $(300,12),(60,0),(300,12)$   & use-dependence / slow inactivation \\
8 & depol.\ conditioning $+$ test        & $(250,15),(150,12)$          & depolarising history \\
9 & brief hyperpol.\ conditioning $+$ test & $(40,{-}30),(150,12)$      & fast de-inactivation \\
\bottomrule
\end{tabular}
\caption{\textbf{Nine seed protocol types for \HHbench.}
The headline acquisition enumerates the expanded 90-action menu described above.
Each protocol is a
sequence of (duration, amplitude) current segments; a leading hyperpolarising segment is a conditioning
pre-pulse. Rows 1--4 are standard current-clamp steps; rows 5--9 are the non-textbook protocols that expose
the hidden mechanisms of \cref{tab:hhnovel}.
}
\label{tab:hhproto}
\end{table}

\paragraph{Evaluation.}
Because a spike is a ${\sim}1$\,ms all-or-none event, a
sub-millisecond timing mismatch between model and data produces a
${\sim}100$\,mV pointwise error even for an essentially correct
model.
Thus,
rather than asking agents to predict the exact voltage trajectory $y_{1:T}$
in response to a novel perturbation (which is very hard,
as shown in  \cref{fig:hhpredoverlaymain}),
we  just ask it
a target functional $\target(y_{1:T})$  of summary statistics,
which include
spike counts in the
test and conditioning windows, their use-dependent run-down,
within-pulse adaptation, and two sub-threshold voltage summaries,
as illustrated in  \cref{fig:hhstochfeatures}.
These are computed as follows:\footnote{
Run-down shares information
with the pre/post spike counts, so these six components are not independent.
}
\begin{equation}
\target(y)=\Big(\underbrace{n_{\mathrm{test}}}_{\text{test spikes}},\
\underbrace{n_{\mathrm{pre}}}_{\text{pre-pulse spikes}},\
\underbrace{n_{\mathrm{pre}}-n_{\mathrm{test}}}_{\text{run-down}},\
\underbrace{n^{\mathrm{early}}_{\mathrm{test}}-n^{\mathrm{late}}_{\mathrm{test}}}_{\text{adaptation}},\
\underbrace{\min_t V(t)}_{V_{\min}},\
\underbrace{\bar V_{\mathrm{end}}}_{\text{steady state}}\Big),
\label{eq:hhstochfeatures}
\end{equation}
where $n_{\mathrm{test}},n_{\mathrm{pre}}$
count upward zero-crossings of $V$ in the test window (after any
conditioning pre-pulse) and before it,
$n^{\mathrm{early}}_{\mathrm{test}}{-}n^{\mathrm{late}}_{\mathrm{test}}$
splits the test window in half, and $\bar V_{\mathrm{end}}$ is the mean voltage over the steady-state tail
of the trace (the final few percent, after the stimulus ends).
This is chosen so that both
the rate-signature worlds (\nafatigue, \textbookM) and the sub-threshold/burst worlds (\hsag,
\carebound) leave a signal.

We compute this target using a fixed \emph{held-out query set} $\queryDist$ of 24 designs,
with eight each from the single-pulse, conditioning--test, and paired-pulse families. The queries have
no exact overlap with the 90-action design menu, are shared across worlds, and are hidden from the agents
(so the agent is always forecasting interventions it never ran). These test protocols are chosen to be
discriminative --- each is a stimulation sequence under which candidate mechanisms may disagree.
For this benchmark, \cref{eq:nmse} uses $t_c=0$, six features, 24 queries,
one entry per feature per query, and fixed feature scales:
\begin{equation}
 L=\frac{1}{24\cdot6}\sum_{q=1}^{24}\sum_{i=1}^{6}
 \left(\frac{\widehat{\E}[F_i(Y)\mid\design_q,\D]
                  -\widehat{\E}_{\star}[F_i(Y)\mid\design_q]}{s_i}\right)^2,
 \qquad s=(3,3,3,3,12,8).
 \label{eq:hhFixedScaleMSE}
\end{equation}
The first four scales apply to spike-count differences and the last two to mV.
The scaling makes errors in counts and voltages commensurate: an error of
three spikes, 12\,mV in the minimum voltage, or 8\,mV in the tail mean each
contributes one before averaging over features and queries. The unequal
voltage scales assign $ (12/8)^2=2.25 $ times more weight per squared mV
to the sustained tail level than to the single minimum. These are fixed
benchmark weighting conventions, not physiological constants or estimated
noise standard deviations; the precise values are not uniquely determined.
The count-derived features overlap, so the metric deliberately gives more
weight to spike-count behavior; it is not a covariance-whitened distance.
The oracle expectation is estimated using independent stochastic rollouts;
this is error in expected features, not squared error against one noisy path.

\paragraph{Distributional feature scores.}
\label{app:CRPS}
A complementary diagnostic is CRPS for each of the six marginal feature
distributions. If $f_i^{(r)}=F_i(Y^{(r)})$ are $R$ independent
posterior-predictive draws and $f_i^{\rm obs}$ is the feature of an
independent held-out rollout, an unbiased Monte Carlo estimate is
\begin{equation}
 \widehat{\mathrm{CRPS}}_i=
 \frac1R\sum_{r=1}^R|f_i^{(r)}-f_i^{\rm obs}|
 -\frac{1}{2R(R-1)}\sum_{r\ne r'}|f_i^{(r)}-f_i^{(r')}|.
 \label{eq:hh_feature_crps}
\end{equation}
We can average this over held-out rollouts and queries and report each
feature separately; a dimensionless aggregate uses
$\frac16\sum_i\mathrm{CRPS}_i/s_i$ (not $s_i^2$).
This tests uncertainty about counts and voltage summaries without requiring
prediction of precise spike times. Unlike MSE of expected features, it
must be scored against realized held-out features, not their oracle mean.
\eat{
It complements the joint energy score, since marginal scores do not test
cross-feature dependence. Correlation of energy score with MSE does not
establish calibration.
}

\eat{
As a distributional diagnostic, we additionally score the raw held-out voltage on a fixed 1-ms grid using
the time-averaged marginal continuous ranked probability score,
as defined in \cref{eq:CRPS}.
This is a proper score for each time-indexed voltage marginal and tests both sharpness and calibration.  It
does not test the joint temporal copula; the multivariate energy score of the standardized feature vector in
\cref{eq:hhstochfeatures} supplies a complementary path-level summary.
Because millisecond spike-time shifts
can dominate \cref{eq:CRPS}, feature nMSE remains the primary task metric and trajectory CRPS is reported
as a diagnostic rather than a replacement.
}

\begin{figure}[t]
\centering
\input{sectionsv5/fig_hh_features}
\caption{\textbf{From a raw voltage trace to the per-trace feature vector $\target(y)$ of
\cref{eq:hhstochfeatures}}, computed on
real \HHbench\ traces. \emph{(a)} On a paired-pulse protocol the spike-count features are the test- and
pre-pulse counts ($n_{\mathrm{test}}$, $n_{\mathrm{pre}}$; upward $0$\,mV crossings, triangles), their
use-dependent \emph{run-down} $n_{\mathrm{pre}}{-}n_{\mathrm{test}}$ (here the \nafatigue\ (slow-Na) cell fires
less on the second pulse), and the within-pulse \emph{adaptation} (early-half minus late-half of the test
window, dashed divider). \emph{(b)} On a hyperpolarising step the sub-threshold features are the voltage
minimum $V_{\min}$ (the $I_h$ sag / hyperpolarisation depth) and the steady-state tail $\bar V_{\mathrm{end}}$.
}
\label{fig:hhstochfeatures}
\end{figure}

\paragraph{Baseline.}
As a baseline, we use a transductive ICL
which computes
$\hat{F}(\design)$ by passing in the observed data
$\data = \{(\design_i,\{(t_{ij},y_{ij})\}_{j=1}^{20})\}$ into the prompt,
along with the novel test input $\design$,
and asking the LLM to predict the resulting features.
ICL (+MDA design) replays the earlier archived MDA histories, not the new
online-expansion histories (i.e., it does not choose its own experiments),
but ICL (+LLM design) uses an LLM to design the experiments.

\begin{figure}[h]
\centering
\includegraphics[width=0.8\textwidth]{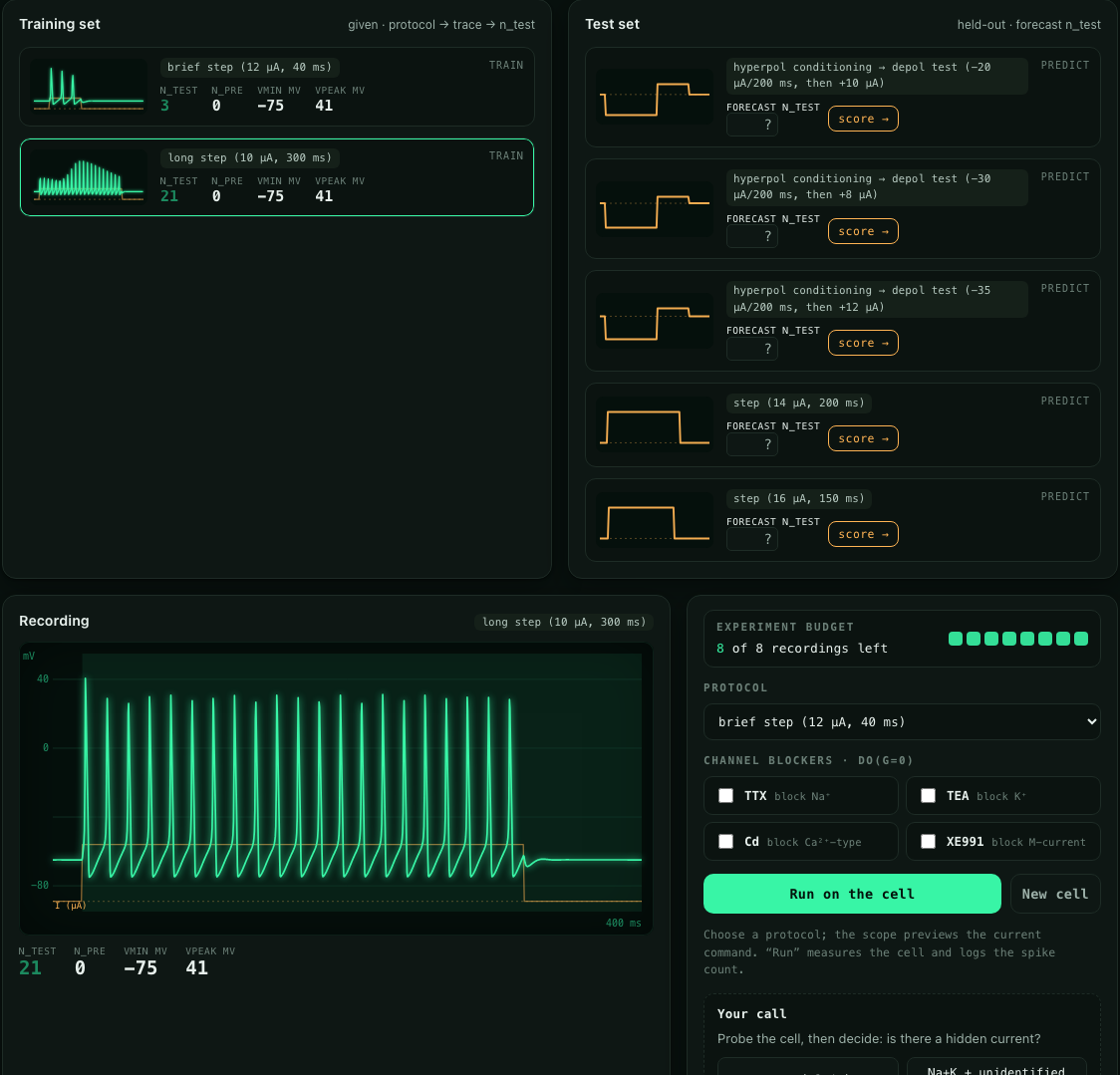}
\caption{\textbf{\HHbench}.
  Screenshot of our app, which lets users interact with the same
  environment we give our agents (except the agents see numerical data, not images.)
  The top left is the training set, $\Dtrain$,
  the top right is the test set, $\Dtest$,
  and the bottom row is the interactive environment.
  The agent can choose a sequence of input currents
  $\inputs_{1:T}$ by specifying the magnitude and duration of a step pulse
  (shown in orange).
  The agent can also choose from a finite set of interventions,
  corresponding to blocking different ion channels
  (shown as white boxes).
  The resulting output voltage $y_{1:T}$ is shown in the green trace.
}
\label{fig:patch}
\end{figure}

\paragraph{Interactive app.}
\Cref{fig:patch} shows a screenshot for a web app we built that lets
users try this benchmark for themselves.
The app is available at
\redacted{\url{https://github.com/murphyk/neuronbench}}.

\subsection{MDA}
\label{app:hhstochmethods}

In this section, we describe how MDA is applied to this domain.

\paragraph{Models.}
The agent is told that the cell is a standard Na$+$K$+$leak spiker
(the familiar null model) that may additionally carry ``an
additional membrane current of unknown identity, voltage-dependence,
and kinetics'' --- an \emph{opaque} prior that does not name or
parameterise the novel current.
The LLM must propose suitable mechanisms,
which are then simulated by the SDE sampler.

\paragraph{Priors.}
\label{app:hhstoch_priors}
\label{app:hh_priors}

In the headline open-library runs, the LLM declares uniform bounds for
conductance, half-activation voltage, and activation time constant (or just
the slow-Na time constant). The backend adds shared leak
$g_{\mathrm L}\sim\mathcal U(0.2,0.45)$ and, for transient extra currents,
$\tau_h\sim\mathcal U(20,300)$ ms.
Reversal potential and activation direction are proposed structural choices;
activation slopes are fixed to $8$ mV (depolarizing) or $-6$ mV
(hyperpolarizing). Inactivation half-voltage is the midpoint of the proposed
activation bounds minus $22$ mV, with slope $5$ mV.
These data-dependent priors differ from the named-archetype reference ranges
in \cref{tab:hhpriors}. Channel-noise scale and observation noise are fixed.
See \cref{app:promptsHH} for the current proposer interface.

\begin{table}[t]
\centering
\footnotesize\setlength{\tabcolsep}{4pt}
\begin{tabular}{@{}lll@{}}
\toprule
Quantity & Symbol & Prior / value \\
\midrule
Na$^+$ and K$^+$ conductances & $g_{\mathrm{Na}},g_{\mathrm K}$ & $120,36$ (fixed) \\
capacitance & $C$ & $1.0\,\mu$F/cm$^2$ (fixed) \\
leak conductance & $g_{\mathrm L}$ & $\mathcal U(0.2,0.45)$ \\
$I_h$ & $g,V_{1/2},\tau$ & $\mathcal U(2,10),\mathcal U(-105,-85),\mathcal U(60,240)$ \\
T-type Ca$^{2+}$ & $g,V_{1/2},\tau_h$ & $\mathcal U(1.5,8),\mathcal U(-65,-45),\mathcal U(12,180)$ \\
D-type K$^+$ & $g,V_{1/2},\tau_h$ & $\mathcal U(4,14),\mathcal U(-40,-20),\mathcal U(100,300)$ \\
M-type K$^+$ & $g,V_{1/2},\tau$ & $\mathcal U(1,6),\mathcal U(-45,-25),\mathcal U(30,90)$ \\
slow Na$^+$ inactivation & $\tau_s$ & $\mathcal U(200,600)$ \\
\bottomrule
\end{tabular}
\caption{\textbf{Named-archetype reference parameter ranges.} These describe
the earlier fixed-archetype setup, not the LLM-declared headline priors. Conductances are
in mS/cm$^2$, voltages in mV, and time constants in ms.  Kinetic functional
forms, reversal potentials, and the canonical excitable backbone are fixed by
the archetype; the extra mechanism's parameters and shared leak are inferred.}
\label{tab:hhpriors}
\end{table}

\paragraph{Likelihoods.}
The (observed data) likelihood for model $m$ is given by
\begin{align}
  p(y_{1:T}\mid m,\theta)=\int p(y_{1:T}\mid z_{0:T},m,\theta)\;p(z_{0:T}\mid m,\theta)\;dz_{0:T},
  \label{eq:HHstochLikelihood}
\end{align}
where $y_{1:T}$ is the observed voltage trace and $z_{0:T}$ the latent gating path.
This requires marginalising
over the stochastic latent path $z_{0:T}$ --- a high-dimensional path integral with no closed form,
because the Fox--Lu transition density $p(z_t\mid z_{t-1})$ is itself intractable.
Below we discuss how to approximate this integral
using a bootstrap particle filter (\cref{alg:pf}).

\paragraph{Particle filtering.}
The agent fits a stochastic state-space model \eqref{eq:ssm}
where the latent state
$z_t=(V_t,\{x_c(t)\})$ (voltage and gates) evolves by the discretised Fox--Lu transition
$p(z_t\mid z_{t-1},\design)$ of \cref{eq:foxlu,eqn:hh}, and the voltage is observed with Gaussian noise,
$y_t\sim\gauss(V_t,\sigma^2)$. Candidate models $m$ differ in \emph{structure} (which channels are present) and
in the conductances $\theta$; the channel count $\Nnoise$ (the noise scale) is a known part of the model here. The
one-step transition density $p(z_t\mid z_{t-1},\design)$ has no closed form --- it is a nonlinear diffusion
over the interval --- but the bootstrap particle filter never needs it. It only \emph{samples}
the transition (successive $0.05$-ms Euler--Maruyama substeps with Gaussian gate noise, \cref{eq:foxlu}) as its proposal,
and only \emph{evaluates} the tractable observation density $\gauss(y_t\mid V_t,\sigma^2)$ to reweight the
particles,
from which $p(y_{1:T}\mid m,\theta)$ is estimated; parameter integration
is a separate outer calculation.
So the intractable-likelihood regime needs only a \emph{simulator}
of the latents plus an evaluable observation
model.

\begin{figure}[t]
\centering
\includegraphics[width=0.86\textwidth]{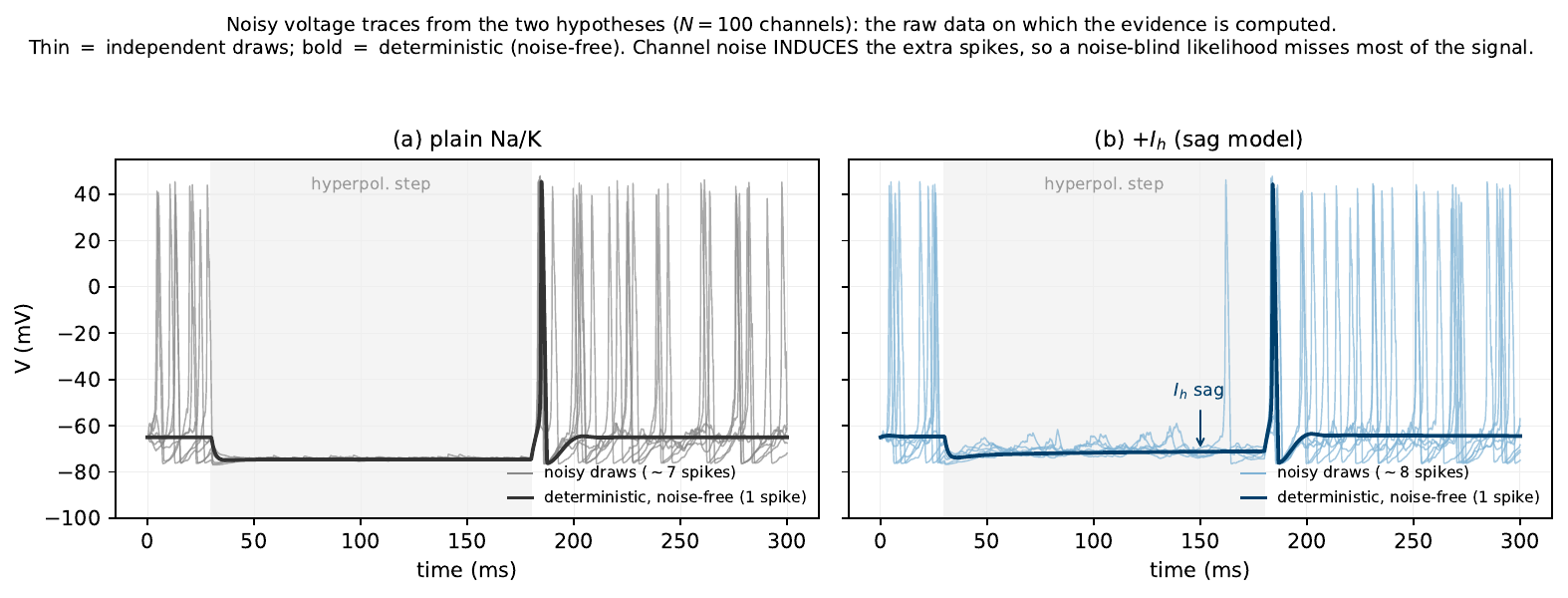}
\caption{\textbf{Stochastic-latent \HHbench: the raw data.} Noisy voltage traces from the two competing
  hypotheses under a moderate hyperpolarising-step protocol, at $\Nnoise{=}100$ channels (thin: independent draws;
  bold: the \emph{deterministic}, noise-free trace). \emph{(a)} A plain Na/K cell. \emph{(b)} The same cell plus a
  hyperpolarisation-activated $I_h$ current (the \hsag model of \cref{tab:hhnovel}), whose only signature is a
  small depolarising \emph{sag} during the step (arrow). Because that sag is comparable in size to the channel
  noise, the two hypotheses overlap and cannot be told apart by eye. Note that channel noise \emph{induces}
  spiking: the deterministic trace fires once where the noisy cell fires ${\sim}7$ times, so a noise-blind
  (deterministic) likelihood misses most of the signal --- the failure mode quantified in \cref{fig:hhstochalg}.}
\label{fig:hhstochdata}
\end{figure}

\begin{figure}[t]
\centering
\input{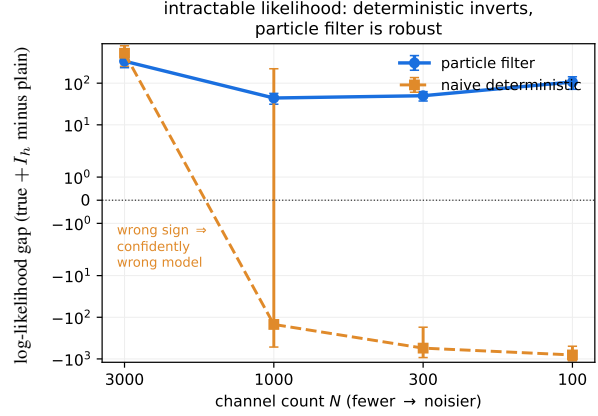}
\caption{\textbf{Stochastic-latent \HHbench: estimating the intractable likelihood by simulation.} We score
  the $I_h$-vs-plain decision on the data of \cref{fig:hhstochdata}, sweeping the channel count $\Nnoise$ (fewer
  $=$ noisier): the fixed-parameter log-likelihood gap vs.\ $\Nnoise$. A likelihood that ignores the
  process noise (a single deterministic rollout $+$ Gaussian observation, orange) degrades and, below
  $\Nnoise{\approx}1000$, \emph{inverts} --- a negative gap means it confidently selects the \emph{wrong} mechanism.
  A bootstrap particle filter (blue), which estimates $p(y\mid m,\theta)$ by propagating the latent gating SDE,
  stays robustly positive. Bars/points are $\pm1$\,SE over independent noise realisations.}
\label{fig:hhstochalg}
\end{figure}

\paragraph{Why the deterministic likelihood breaks.}
To illustrate why we cannot just use a deterministic ODE model
(and hence a deterministic likelihood, as we did in \cref{eq:likelihood}),
we consider a simple example where we need to
distinguish just two hypotheses: a plain Na/K cell vs
the novel \hsag model $I_h$ defined in
\cref{tab:hhnovel}.
The noisy voltage traces from the two hypotheses are shown in \cref{fig:hhstochdata} --- the $I_h$ sag is
subtle relative to the channel noise, so they overlap. In \cref{fig:hhstochalg} we plot
the fixed-parameter log-likelihood gap,
$\log p(y\mid m_{I_h},\theta_{I_h})-\log p(y\mid m_{\rm plain},\theta_{\rm plain})$,
vs noise level $\Nnoise$. Parameters are fixed in this illustrative diagnostic;
it does not integrate parameter uncertainty.
We see that a likelihood that treats the intrinsic
channel noise as zero degrades as the noise grows and, at $\Nnoise{=}1000$
channels, \emph{inverts}: it confidently selects the wrong
mechanism. By contrast, the particle filter  stays robustly correct at every noise level.

\subsection{Results}
\label{app:hhmopenladder}
\label{app:neuroResults}

\paragraph{Sample efficiency.}
The main sample efficiency plot is shown in
\cref{fig:hhmopenmain}.
MDA is given two initial trajectories in $\D_0$,
and then creates an initial hypothesis library using the LLM proposal.
\eat{
The plotted SDE cells reuse the EIG arm of the frozen-archive acquisition
study (manifest \texttt{hh-regime-v2-task-design-v1}, launch commit
\texttt{8e2ee5e}, manifest hash prefix \texttt{649023480430}).
Its archived configuration explicitly disables later model births.
Thus this comparison tests inference and acquisition with a truth-blind,
LLM-initialized library, not successful online expansion.
}
We used particle filtering to approximate
the likelihood (marginalizing over stochastic trajectories),
an initial tempered-SMC parameter fit followed by recursive likelihood-weight
updates on retained parameter particles, and evidence weights over the archive.
The new controlled cohort uses six worlds and three seeds, with $N_p=128$,
$N_z=128$, and 16 initialization tempering steps. Subsequent updates retain
parameter particles without rejuvenation or full refitting.
All four model-based arms share the initial archive and ten observed trajectories;
later model births are disabled. Prediction uses two independent 128-draw
samples from the full joint model--parameter mixture in \cref{eq:BMA},
without top-component truncation. Both ICL baselines are restricted to the
same worlds and seeds.

The factorial crosses ODE/SDE dynamics with recursive SMC/direct optimization+BIC.
ODE likelihoods are Gaussian; SDE likelihoods use a PF. BIC optimization uses
three rounds of 24 candidates and independently validates four finalists with
three likelihood estimates each. All arms share the structural prior
$\log p(m)=-2\dim(\theta_m)+\mathrm{const}$.
SDE/ODE paired area-under-learning-curve ratios are $0.288$ under SMC and
$0.195$ under BIC, favoring SDE in all six worlds. The SDE BIC/SMC ratio is
$1.106$ (95\% hierarchical-bootstrap interval $[0.972,1.354]$), so this study
does not establish a difference between those inference stacks.
At ten trajectories, the MAP-model parameter population has only one distinct
particle in 13 of 18 ODE-SMC runs, versus none of the SDE-SMC runs.
Thus particle impoverishment may contribute to ODE-SMC deterioration;
directly optimized ODE+BIC also performs poorly.
Median worker runtimes are 9.1, 9.2, 11.2, and 18.6 minutes for ODE-SMC,
SDE-SMC, ODE-BIC, and SDE-BIC, respectively, including predictive evaluation.

\begin{figure}[!ht]
 \centering
 \includegraphics[width=\linewidth]{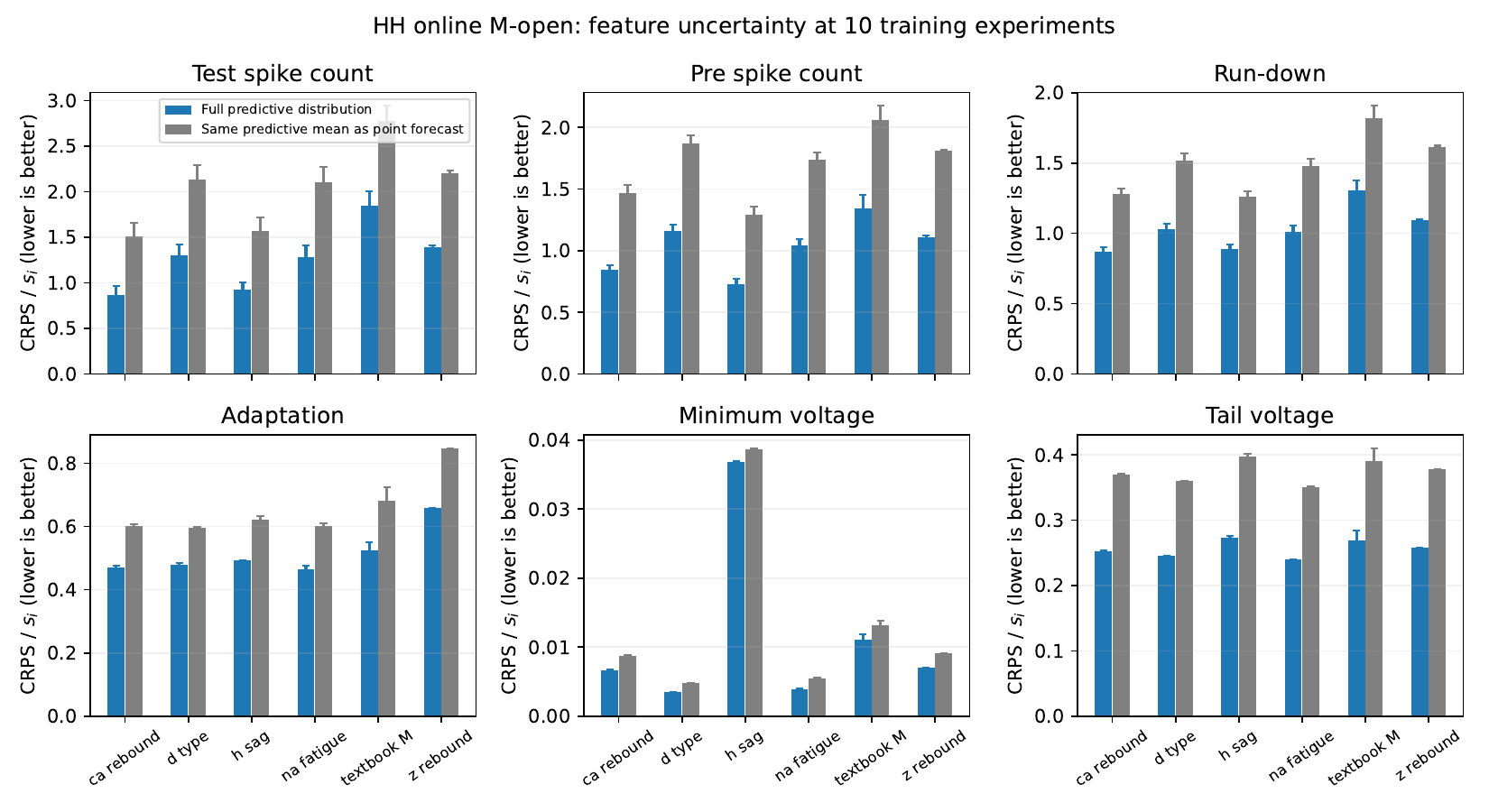}
 \caption{\textbf{HH marginal feature CRPS at ten experiments.}
 Six worlds, six seeds each, and 24 held-out queries per cell. Blue:
 samples from the full model--parameter posterior followed by stochastic
 simulation. Gray: a point distribution at the same estimated predictive
 mean, whose CRPS equals absolute error. Each cell averages two 128-draw
 predictive repeats against independent oracle rollout features; repeats
 share oracle draws. Scores are divided by the fixed $s_i$ of
 \cref{eq:hhFixedScaleMSE}. Bars average seeds; error bars are $\pm1$
 seed-level s.e. within each world. Features are extracted from simulated
 signal voltage before adding measurement noise, matching the existing
 feature target. These are final-budget distributional diagnostics, not
 learning curves or an ablation of parameter integration.
 }
 \label{fig:hh_feature_crps_endpoint}
\end{figure}

\paragraph{CRPS.}
\Cref{fig:hh_feature_crps_endpoint} shows CRPS scores
for all 36 final online
$\mathcal M$-open posteriors, at ten total experiments,
using a stochastic forecast vs just predicting the mean.
Mean standardized
CRPS is $0.687$, versus $1.049$ for a point distribution at the same
predictive mean. Retaining predictive spread improves the score for all
six features in aggregate. This diagnostic does not isolate model or
parameter uncertainty, nor compare independently fitted inference arms.

\eat{
The replay uses the full joint posterior mixture, unlike the historical
top-24 predictive readout, and leaves fitted posteriors unchanged.
Only final full posteriors were retained, so these scores do not establish
improvement over experiment budget.
}

\begin{figure}[t]
\centering
\includegraphics[width=0.78\textwidth]{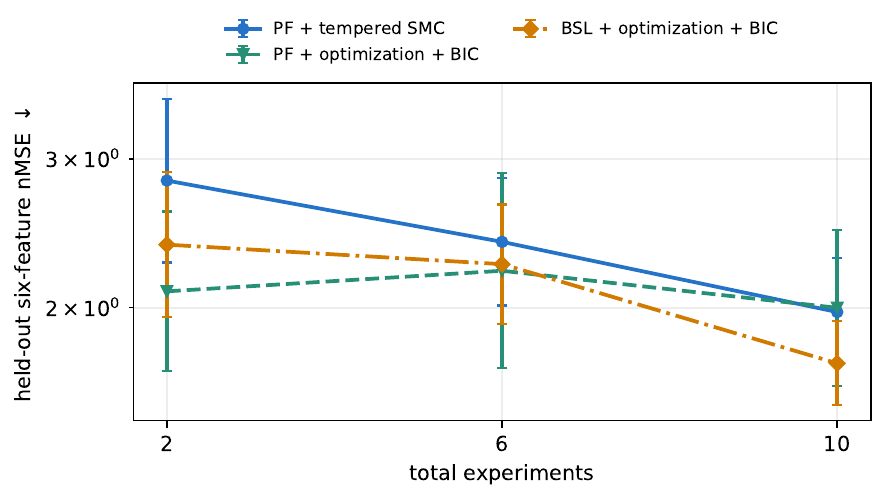}
\caption{\textbf{Standalone stochastic likelihood optimization on \HHbench.}
  All methods receive the same two initial trajectories, truth-blind frozen
  archives, eight EIG-selected actions and realized observations.  PF+tempered
  SMC supplies the reference curve (initial fit followed by recursive updates).
  The other arms independently optimize either the particle-filter likelihood
  or a Gaussian synthetic likelihood, then use BIC model weights; stochastic
  posterior-predictive simulation is retained in both.  Points show total
  budgets $2,6,10$ and are geometric means over 36 paired world--seed units,
  with seeds averaged on the log scale within each of six worlds; bars are
  $\pm1$ world-level s.e.  Relative to PF+tempered SMC, the geometric
  learning-curve ratios are $0.894$ for PF+optimization+BIC and $0.897$ for
  BSL+optimization+BIC; neither paired six-world contrast is statistically
  resolved.
}

\label{fig:hhbslladder}
\end{figure}

\paragraph{Controlled inference ablations.}
Our main method uses the SDE model, but also uses PF for estimating
the likelihood and tempered SMC for estimating the evidence.
We wanted to determine how important each of these components were,
so we did a controlled ablation,
where we retain SDE+PF but replace parameter integration with optimization
and use BIC model weights, keeping archives, designs, and observations fixed.
We also optimize Bayesian Synthetic Likelihood (\cref{app:BSL}) and use BIC.
The results are shown in \cref{fig:hhbslladder}.
The optimization+BIC configurations have slightly lower point estimates,
but neither paired contrast is statistically resolved.

To quantify this, we consider the area under the learning curve (AULC),
which we define as
\begin{equation}
  \operatorname{AULC}_{a,w,s}
  = \int_{b_{\min}}^{b_{\max}} L_{a,w,s}(b)\,\mathrm db.
  \label{eq:hhAULC}
\end{equation}
where $L_{a,w,s}(b)>0$ denotes the held-out loss of method $a$
at the displayed total-experiment budget $b$, in world $w$ and replicate $s$.
When comparing methods $a$ and $c$,  we report the geometric learning-curve ratio
\begin{equation}
  R_{a/c}
  = \exp\!\left\{
      \mathbb E_{w,s}\!\left[
      \frac{1}{b_{\max}-b_{\min}}
      \int_{b_{\min}}^{b_{\max}}
      \log\frac{L_{a,w,s}(b)}{L_{c,w,s}(b)}\,\mathrm db
      \right]\right\},
  \label{eq:hhGeometricCurveRatio}
\end{equation}
using trapezoidal integration over the displayed budgets, averaging
replicates on the log scale within each world and then averaging worlds.
This geometric learning-curve loss ratio is not generally the ratio of the
ordinary AULCs in \cref{eq:hhAULC}; it averages log loss ratios before exponentiation.
Thus $R_{a/c}<1$ favors $a$; for example, $R_{a/c}=0.80$ means that $a$ has
20\% lower error than $c$ on the multiplicative learning-curve summary.
The results are shown in \cref{tab:hhInferenceLadder},
and show once again that all SDE methods are much better than ODE,
while the stochastic inference configurations have similar predictive accuracy.
This is not a direct measurement of evidence-estimation error.

\paragraph{Optimizing the stochastic likelihood.}
For the PF+optimization+BIC arm, we use
\begin{equation}
 \mathrm{BIC}^{\mathrm{PF}}_m
 =-2\log\widehat p_{\mathrm{PF}}(\D\mid m,\hat\theta_m)
   +\nparams_m\log N,
 \qquad p(m\mid\D)\approx
 \frac{\exp(-\mathrm{BIC}^{\mathrm{PF}}_m/2)}
 {\sum_{m'}\exp(-\mathrm{BIC}^{\mathrm{PF}}_{m'}/2)}.
 \label{eq:hh-pf-opt-bic}
\end{equation}
Here $\nparams_m$ counts fitted parameters, the model prior is uniform,
and $N$ is 20 times the number of observed trajectories.
The PF integrates stochastic latent trajectories with fixed 2-mV observation
noise; no additional residual-scale multiplier is fitted.
To obtain $\hat\theta_m$, we use bounded derivative-free cross-entropy search:
three batches of 24 parameter candidates, retaining the top quarter to adapt
the sampling distribution. Search evaluations use 64 PF particles and common
random numbers within each batch to reduce noise in candidate comparisons.
The four best candidates are then rescored using three independent
128-particle PF evaluations each. We average these likelihood estimates
(using log-mean-exp numerically), select the best finalist, and use its
validated score in \cref{eq:hh-pf-opt-bic}. This stabilizes the noisy search
without claiming to find the exact likelihood maximum.
BSL gave a smoother optimization objective in our pilot.
In the matched optimization sweep, BSL used 1.71 aggregate A10G-hours versus
2.97 for PF, i.e., about 42\% less GPU time.
(These are aggregate compute times, not elapsed latency.)
The speedup is limited because BSL still needs to sample trajectories
to compute $\mu_{m,\theta}$ and $\Sigma_{m,\theta}$.
True speedups would require avoiding these rollouts
and the use of amortized inference methods,
such as neural likelihood estimation, neural posterior estimation, or neural ratio estimation;
we leave this to future work.

\begin{table}[t]
\centering
\scriptsize
\setlength{\tabcolsep}{3.2pt}
\begin{tabular}{@{}llllr@{}}
\toprule
Latent dynamics & Likelihood & Parameter treatment & Model evidence & $R$ \\
\midrule
SDE & bootstrap PF & initial SMC + recursive updates & SMC normalizer & 1.000 \\
SDE & bootstrap PF & bounded optimization & BIC & 0.894 \\
SDE & Gaussian BSL & bounded optimization & BIC & 0.897 \\
ODE & Gaussian residual & bounded optimization & BIC & 2.974 \\
\bottomrule
\end{tabular}
\caption{\textbf{Controlled inference ladder.}
  Every row replays the same frozen archives, actions, and realized data;
  hence this table isolates inference and prediction rather than comparing
  independently selected acquisition policies.  $R$ is the geometric
  learning-curve nMSE ratio in \cref{eq:hhGeometricCurveRatio}, relative to
  SDE+PF+tempered SMC (lower is better).  The PF and BSL plug-in rows are
  standalone estimators: each obtains $\hat\theta_m$ by bounded optimization,
  independently validates the finalists, and then applies BIC.  The ODE row
  is the matched deterministic replay from the controlled ladder.}
\label{tab:hhInferenceLadder}
\end{table}

\clearpage
\section{\boxing}
\label{app:boxing}

\subsection{Details on the benchmark}
\label{app:boxingBench}

In this section, we briefly describe the  \textsc{BoxingGym} benchmark from
\citep{gandhi2025boxinggym}.
This implements ten scientific domains as generative
probabilistic models; an agent interactively chooses designs (for up to 10 steps),
observes outcomes, and is scored by
(i) held-out predictive loss;
(ii) the \emph{EI-regret} of its chosen designs (compared to the
EIG estimated using the ground truth model and the best of $100$ random
designs);  and (iii) an \emph{explain-to-a-novice} model-discovery metric.
We focus on predictive loss to match the rest of the paper. We found the
``explain-to-a-novice'' metric
gave very high variance results.
Thus we just report predictive loss (under novel perturbations),
to be consistent with the rest of this paper.

Note: Unlike the main paper, all results in this section use Opus~4.7
(with reasoning turned off, which is its default setting).
Comparisons within this auxiliary study all share that LLM,
so we can compare the relative performance of methods.
However we make no claims about absolute top performance,
or the relative performance of LLMs across domains.

\paragraph{Domains.}
Two of the ten domains (\textsc{emotion}, \textsc{moral-machines})
use a large language model \emph{as the
participant-simulator}; they test elicitation of an LLM's implicit preferences rather than discovery of a
mechanistic generative process, so we exclude them from our experiments.
We also exclude the Lotka-Volterra (predator-prey) example from this
auxiliary study. This leaves us with the seven domains shown in
\cref{tab:boxingdomains}, which we
partition into two clusters.

The first cluster consists of covariate-regression models,
where the agent needs to learn the form
of the function that defines the mean of the (scalar) output,
analogous to learning the symbolic rate law in the  chemistry domain.
Since there are no latent variables, the GLM likelihood
is fast to compute.

The second cluster consists of latent variable models, where the number of parameters
can grow with the size of the data.
To represent these  models compactly, we use a probabilistic programming language (PPL),
as we explain in \cref{app:codemda}.

\paragraph{Initial data $\D_0$ and query set $\queryDist$.}
Every \textsc{BoxingGym} run is seeded with a passive set $\D_0$ of $n_0$ designs drawn uniformly at random
from the candidate pool ($n_0{=}2$ for the GLM/latent domains; the \textsc{location} panels of
\cref{fig:codelocation} use a shared $n_0{=}3$ seed set so every agent starts from identical data), after
which the agent designs its experiments. The query set $\queryDist$ on which we score the held-out predictive
loss is the set of \emph{unobserved} designs: the held-out $(s,q)$ response cells for \textsc{irt}, the
held-out probe locations $x\in\real^2$ for \textsc{location}, and fresh covariate points for the GLM domains
--- always disjoint from the experiments the agent ran.

\begin{table*}[!hb]
\centering\scriptsize
\caption{The seven  \textsc{BoxingGym} domains we run MDA
  on. $\Phi$ is the standard-normal CDF; $\sigma(\cdot)$ the
  logistic. The first five domains  are
  covariate-regression domains with global parameters.
  Below the line we show
  \textsc{irt} and \textsc{location}, which also have
 latent variables whose number grows with the data size
  (\textsc{irt}: one ability per student and one difficulty per
  question) or model size (\textsc{location}: one position $\theta_k$ per source).
  We use code synthesis (in \textsc{NumPyro}) to represent these
  latent-variable models.
  }
\label{tab:boxingdomains}
\begin{tabular}{@{}lll@{}}
\toprule
domain & ground-truth generative model & design $\design$\\
\midrule
\textsc{dugongs} & $p(y|t;\theta)= \gauss(\mu=\alpha-\beta\lambda^{t}, \sigma^2)$ & age $t\in[0,5]$ \\
\textsc{death} & $p(y|t;\theta)=\text{Binom}(N, \eta=1-e^{-\theta t})$ & time $t\in(0,2)$ \\
\textsc{peregrines} & $p(y|t;\theta)=\text{Poiss}(\lambda=e^{\alpha+\beta_1 t + \beta_2 t^2 + \beta_3 t^3})$
  & time $t\in[0,5]$ \\
\textsc{hyperbolic} & $p(y|R_i,R_d,D;\theta)=\text{Ber}(
\varepsilon+(1-2\varepsilon)\Phi\!\big(\tfrac{R_d/(1+k D)-R_i}{\alpha}\big))$
& $(R_i,R_d,D) \in \real^3$ \\
\textsc{survival} & $p(y_p|t_p,x_p;\theta)=\text{Ber}(\sigma\!\big(t_p\,\lambda_0 e^{\beta\,x_p}\big))$
& patient $p$ \\
\midrule
\textsc{irt} & $p(y_{sq}|\theta) =
\text{Ber}(\sigma\!\big(\theta^{\mathrm{disc}}_q(\theta^{\mathrm{abil}}_s-\theta^{\mathrm{diff}}_q)\big))$
& student $s$, question $q$ \\
\textsc{location} & $p(y|x,\theta)=\gauss(\mu=b+\sum_{k=1}^K \tfrac{\alpha}{c+\lVert x-\theta_k\rVert^2}, \sigma^2)$
& $x \in \real^2$ \\
\bottomrule
\end{tabular}
\end{table*}

\paragraph{Baselines.}
\label{app:apprentice}
As the main baseline, we reimplement the \textsc{Box's Apprentice} agent from
\citep{li2024automated}, which was  used to create the results in
\citep{gandhi2025boxinggym}.
In more detail, following their code, we design the Apprentice
as follows: at each step it prompts the LLM to synthesize a
single model $\hat m$ from the data so far, fits $\hat m$, injects
$\hat m$ (with a residual critique) into the LLM's context, and asks
the LLM to choose the next experiment directly (``where should we observe
next to best improve the model?'').
At the end, it uses $\hat{m}$ to make predictions, similar to MDA.
As another baseline, we use ICL (reusing MDA's data/ design decisions).

\subsection{Priors and likelihoods}
\label{app:boxing_priors}

For the GLM domains,
each candidate is a mean function $\mu_\theta(x)$.
The likelihood is tractable in closed form --- Gaussian
with a known scale $\sigma$ for the real-valued domains, or
the domain's specified count likelihood or Bernoulli likelihood for binary outcomes --- so no particle
filter is needed.
The parameters are given a uniform prior over their range
(proposed by the LLM), and are integrated out using
particle tempered SMC run per model,
with $\nparticlesParams{=}1500$ particles (which are cheap to compute
due to the closed form likelihood).
For the latent variable models, we group the latents together with the fixed parameters,
and use tempered SMC to compute
$p(\D|m) = \int p(\D|m,z,\theta) p(z,\theta|m) dz d\theta$,
rather than nesting PF inside of SMC.

\subsection{Results on GLMs}
\label{app:boxingresults}

\begin{figure}[t]
\centering
\includegraphics[width=0.32\textwidth]{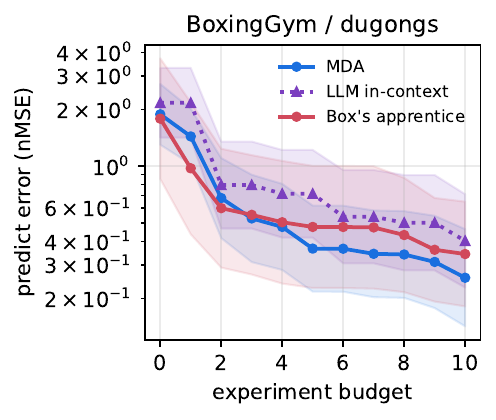}
\includegraphics[width=0.32\textwidth]{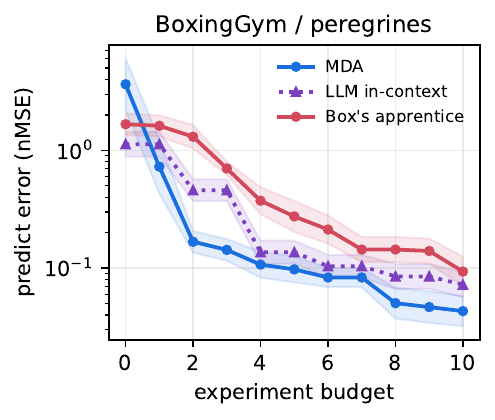}
\includegraphics[width=0.32\textwidth]{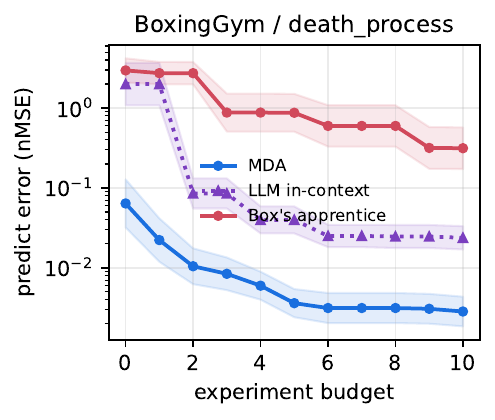} \\
\includegraphics[width=0.32\textwidth]{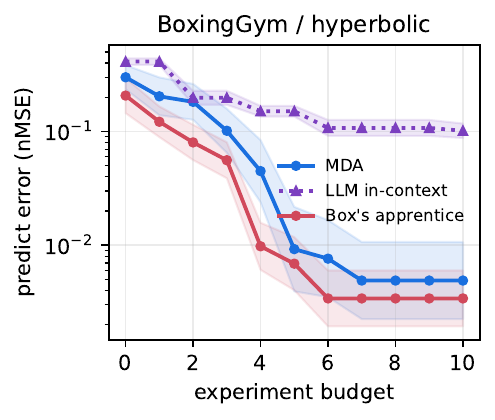}
\includegraphics[width=0.32\textwidth]{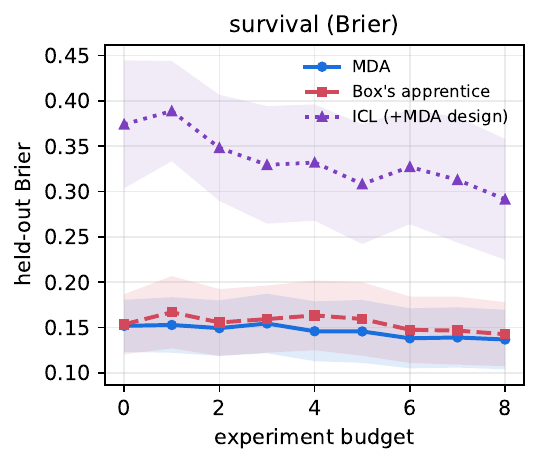}
\caption{\textbf{\textsc{BoxingGym} learning curves for the five covariate-regression domains.}
  MDA, Box's Apprentice, and the model-free ICL (+MDA design) baseline (mean\,$\pm$\,SE; held-out nMSE against the
  noise-free mean $\E[Y\mid x]$, except \textsc{survival}, whose Bernoulli outcome is scored by the proper
  Brier score).
}
\label{fig:boxing_grid}
\end{figure}

The results on the 5 GLM domains are shown in \cref{fig:boxing_grid}.
MDA is better than or on par with Apprentice on all five domains --- by orders of
magnitude on \textsc{death-process}, more modestly on the noisier \textsc{dugongs}, and
only marginally (within overlapping bands) on \textsc{survival} --- and it ties on \textsc{hyperbolic}, a binary-choice domain
that both agents solve almost perfectly.
We also see that MDA consistently beats the model-free ICL baseline
across all five domains; the apprentice is itself out-forecast by the
model-free ICL baseline on \textsc{death-process} and \textsc{peregrines}.
In addition, MDA is much more stable in its predictions than Apprentice,
consistent with averaging over models rather than using a single MAP estimate;
acquisition and model evolution also differ, so this is not an isolated averaging ablation.
(Both average over their parameters.)

\subsection{Inference over latent variable models using a PPL}
\label{app:codemda}

In this section, we describe how to extend MDA to work with models represented as
code, using a probabilistic programming language.
The original paper used \textsc{PyMC}; our implementation uses
\textsc{NumPyro}. This is an implementation choice, not a controlled
comparison of the two probabilistic-programming systems.
The LLM proposes each candidate as a \textsc{NumPyro} program --- a
\verb|def model(ctx)| that declares the priors over latents
$(z,\theta)$ with \verb|numpyro.sample| (a \verb|numpyro.plate| for
the per-unit latents) and registers the expected observable at every
candidate design as a \verb|numpyro.deterministic('mu',...)| node. We
compile it into exactly the callables MDA's loop consumes (prior
sampler, prior log-density, expected observable) by reading
\textsc{NumPyro}'s own \verb|biject_to| transform for each site, so
the tempered SMC does its random-walk moves in the
\emph{unconstrained} space --- where a positive scale, a $[0,1]$
probability, or a plate of positions all mix --- while the model reads
back constrained values. The observation family (Bernoulli or
Gaussian) is specified in the context (part of the benchmark design).
The resulting PPL model is then passed to our standard
adaptive-tempering SMC, to compute the marginal likelihood
\[
p(\D|m) = \int p(\D|m,z,\theta) p(z,\theta) dz d\theta
\]
These evidences determine normalized weights over the finite candidate pool.

\subsection{Results on Item Response Theory}
\label{sec:irt}

\paragraph{Models.}
The \textsc{irt} domain requires the agent to predict
the probability that student $s$ answers item $q$ correctly.
A variety of models have been proposed for this task in the literature,
involving latent per-student \emph{ability} and per-item \emph{difficulty}/\emph{discrimination}
variables.
There is a ladder of models of increasing complexity, including
\begin{equation}
\begin{aligned}
  \textrm{Student}&:\ p_{sq}=\sigma\big(\theta^{\mathrm{abil}}_s\big) \\
  \textrm{1PL (Rasch)}&:\ p_{sq}=\sigma\big(\theta^{\mathrm{abil}}_s-\theta^{\mathrm{diff}}_q\big) &&\text{\citep{rasch1960}}\\
  \textrm{2PL}&:\ p_{sq}=\sigma\big(\theta^{\mathrm{disc}}_q\,(\theta^{\mathrm{abil}}_s-\theta^{\mathrm{diff}}_q)\big) &&\text{\citep{birnbaum1968}}\\
  \textrm{3PL}&:\ p_{sq}=\theta^{\mathrm{guess}}_q+(1-\theta^{\mathrm{guess}}_q)\,\sigma\big(\theta^{\mathrm{disc}}_q(\theta^{\mathrm{abil}}_s-\theta^{\mathrm{diff}}_q)\big) &&\text{\citep{birnbaum1968}}\\
  \textrm{MIRT}&:\ p_{sq}=\sigma\big(\theta^{\mathrm{disc}\top}_q \theta^{\mathrm{abil}}_s-\theta^{\mathrm{diff}}_q\big),\ \ \theta^{\mathrm{abil}}_s\in\real^D &&\text{\citep{reckase2009}}
\end{aligned}
\label{eq:mirt}
\end{equation}
The simplest model
just has one  ability parameter per student;
1PL (one-parameter logistic, the \emph{Rasch} model) has  student ability
and question difficulty; 2PL adds a per-item \emph{discrimination}
$\theta^{\mathrm{disc}}_q$ (how sharply the item separates abilities);
3PL adds a per-item \emph{guessing} floor $\theta^{\mathrm{guess}}_q$;
multidimensional IRT (MIRT) makes ability a $D$-vector. The simulator's
2PL latents have $\gauss(0,1)$ priors; candidate programs specify their own
priors (\cref{fig:irtcode}). These latents are marginalized out, so there are
no fixed parameters in the model. We use conventional IRT $\theta$ notation
here; the $z$ labels in \cref{fig:irttruth} denote the same per-unit latent variables.
The rungs differ only in \emph{which latents exist}, and hence in their complexity
(expressive power).

\textsc{BoxingGym}'s ground truth is 2PL
(see \cref{fig:irtground} for their code).
The resulting predictive distribution has the form
\begin{equation}
  p(Y_{sq}{=}1|m)=\int \mathrm{Ber}\!\Big(1\,\Big|\,\sigma\big(\theta^{\mathrm{disc}}_q(\theta^{\mathrm{abil}}_s-\theta^{\mathrm{diff}}_q)\big)\Big)\,
  p(\theta^{\mathrm{abil}}_s)\,p(\theta^{\mathrm{diff}}_q)\,p(\theta^{\mathrm{disc}}_q)\;\mathrm{d}\theta^{\mathrm{abil}}_s\,\mathrm{d}\theta^{\mathrm{diff}}_q\,\mathrm{d}\theta^{\mathrm{disc}}_q .
  \label{eq:irt}
\end{equation}

\begin{figure}[t]
\begin{lstlisting}[style=pytiny]
# BoxingGym ground truth (PyMC): a 2PL IRT model over an S x Q grid.
with pm.Model():
    alpha = pm.Normal('alpha', 0, 1, shape=S)       # per-student ability      (latent)
    beta  = pm.Normal('beta',  0, 1, shape=Q)       # per-item  difficulty     (latent)
    gamma = pm.Normal('gamma', 0, 1, shape=Q)       # per-item  discrimination (latent, 2PL)
    p = pm.math.invlogit(gamma[None, :] * (alpha[:, None] - beta[None, :]))
    responses = pm.Bernoulli('responses', p=p, shape=(S, Q))
\end{lstlisting}
\caption{\textbf{The \textsc{irt} ground-truth generative model} (\textsc{BoxingGym}, in \textsc{PyMC};
$S{=}Q{=}6$, \texttt{mode=2pl}). Note all three of ability, difficulty, \emph{and} discrimination are
$\gauss(0,1)$ latents --- \cref{eq:mirt} --- so there are no fixed parameters, and because the latent scale is
  $1$ the induced $P(\text{correct})$ sits near $\tfrac12$.
  }
\label{fig:irtground}
\end{figure}

\paragraph{Designs.}
The agent designs which of the $S{\times}Q$ \emph{cells} $(s,q)$ to query; it
sees only the queried cells --- a \emph{sparse, partially observed} response grid --- and is scored on the
held-out cells.

\paragraph{Results.}

\begin{figure}[t]
\centering
\begin{subfigure}[t]{0.46\textwidth}\centering
  \includegraphics[width=\textwidth]{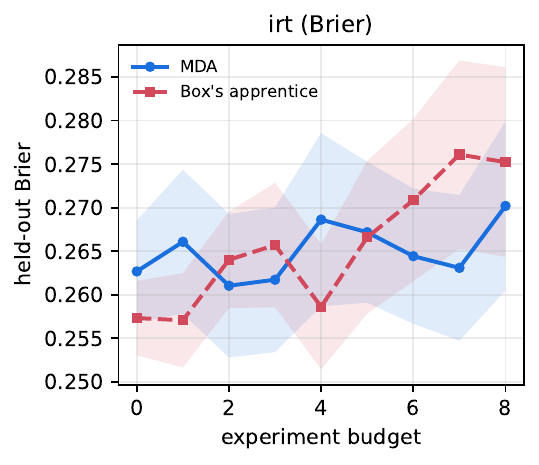}
  \caption{default $6{\times}6$ \textsc{irt} (near-chance)}\label{fig:codelat-6x6}\end{subfigure}\hfill
\begin{subfigure}[t]{0.46\textwidth}\centering
  \includegraphics[width=\textwidth]{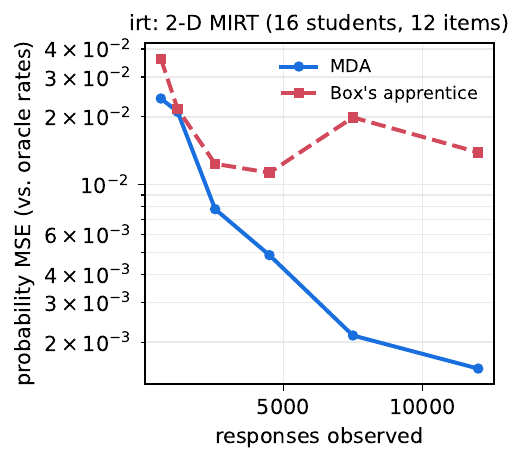}
  \caption{higher-signal $2$-D MIRT \textsc{irt}}\label{fig:codelat-irt}\end{subfigure}
\caption{\textbf{Probabilistic-program comparisons on \textsc{irt}, at two signal levels.}
(\subref{fig:codelat-6x6}): the default $6{\times}6$ \textsc{BoxingGym} instance uses LLM-authored
\textsc{NumPyro} programs. Its data are
weakly informative (\cref{fig:irtground}), and MDA and the apprentice tie within
noise (held-out Brier ${\sim}0.27$, $20$ seeds). (\subref{fig:codelat-irt}): on the higher-signal $2$-D MIRT
instance (\cref{fig:irttruth}: $16$ students $\times$ $12$ items with a hidden second ability axis), the
probability MSE against the true response rates falls as
the response matrix fills. This controlled illustration compares a fixed
five-model ladder against a single evolving LLM program; both use numerical
Brier-risk design. Scores cover all cells, not just unobserved ones.
}
\label{fig:codelatent}
\end{figure}

\begin{figure}[t]
\centering
  \includegraphics[width=\textwidth]{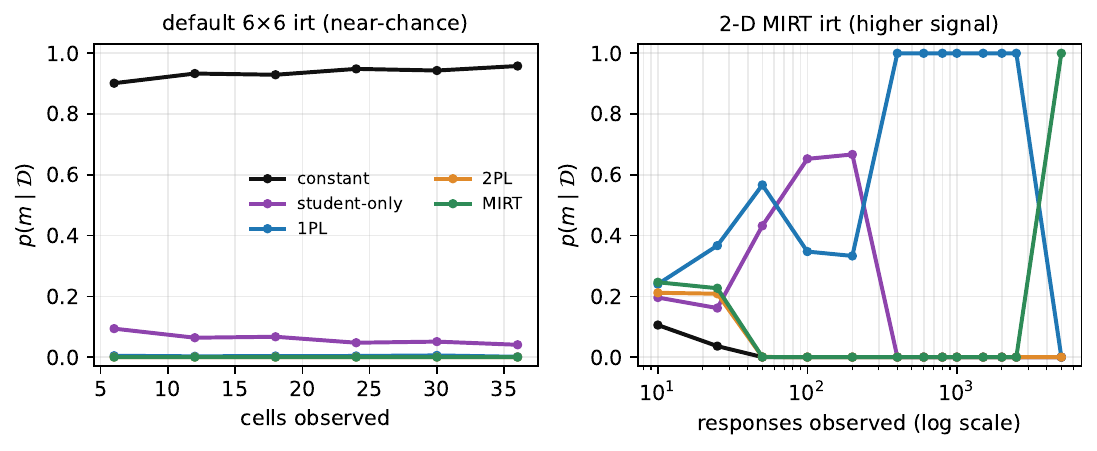}
  \caption{\textbf{Posterior over fixed model ladders for the two \textsc{irt} domains.}
    These controlled inference illustrations do not generate new structures.
    \emph{Left:} low-signal \boxing regime --- with few experiments / little data,
    MDA keeps most of its mass on a simple constant model.
    \emph{Right:} high-signal regime --- as the data grows, MDA climbs the complexity
    ladder, eventually converging on the true $2$-D MIRT model.
    }
\label{fig:IRTpost}
\end{figure}

\Cref{fig:codelat-6x6}
shows the Brier score --- defined as $(y-p)^2$,
where $y \in \{0,1\}$ is the true outcome and $p$ is the predicted
probability --- for  MDA and Apprentice as a function
of number of experiments.
Both are similar in performance, but curiously, both get worse with more data.
The domain is very small:
there are only 6 students and 6 questions, and only 8 of the cells are observed,
and each one only contains a single binary response.
MDA's LLM proposes multiple possible models, given the context and observed data
(see \cref{fig:irtcode}),
but there is not enough evidence to move away from a constant predictor\footnote{
At a global base rate $\mu\approx0.667$, the constant-predictor risk is
$\mu(1-\mu)\approx0.22$. The Bayes floor is instead
$\E_i[p_i(1-p_i)]$, which is smaller when conditional rates vary.
The observed risk near $0.27$ is above the constant-predictor reference;
small samples and nonrepresentative acquisition are possible contributors,
not a demonstrated causal explanation.},
as shown in \cref{fig:IRTpost}~(left).
This is an example of the Bayesian Occam's razor penalizing overly
complex models.
Below we design a larger scale experiment to see if either agent can discover
structure if there is enough signal in the data to warrant it.

\begin{figure}[t]
\begin{lstlisting}[style=pytiny]
# IRT: per-student/per-item latents via numpyro.plate. Verbatim LLM-generated NumPyro programs (mda2.propose_numpyro).
# ctx["features"] is the (C,d) matrix of all candidate designs; the observation
# family (Bernoulli/Gaussian) is a task constant applied externally by the SMC.
import numpyro, numpyro.distributions as dist, jax, jax.numpy as jnp

def constant(ctx):   # Null model: single global success probability
    F = ctx['features']
    C = F.shape[0]
    p = numpyro.sample('p', dist.Beta(1.0, 1.0))
    mu = jnp.ones(C) * p
    numpyro.deterministic('mu', mu)

def rasch(ctx):   # Rasch IRT: student ability minus question difficulty via logistic
    F = ctx['features']
    s_idx = F[:, 0].astype(jnp.int32)
    q_idx = F[:, 1].astype(jnp.int32)
    with numpyro.plate('students', 6):
        ability = numpyro.sample('ability', dist.Normal(0.0, 1.5))
    with numpyro.plate('questions', 6):
        difficulty = numpyro.sample('difficulty', dist.Normal(0.0, 1.5))
    logits = ability[s_idx] - difficulty[q_idx]
    mu = jax.nn.sigmoid(logits)
    numpyro.deterministic('mu', mu)

def twopl(ctx):   # 2PL IRT: per-question discrimination scales ability minus difficulty
    F = ctx['features']
    s_idx = F[:, 0].astype(jnp.int32)
    q_idx = F[:, 1].astype(jnp.int32)
    with numpyro.plate('students', 6):
        ability = numpyro.sample('ability', dist.Normal(0.0, 1.5))
    with numpyro.plate('questions', 6):
        difficulty = numpyro.sample('difficulty', dist.Normal(0.0, 1.5))
        discrimination = numpyro.sample('discrimination', dist.LogNormal(0.0, 0.5))
    logits = discrimination[q_idx] * (ability[s_idx] - difficulty[q_idx])
    mu = jax.nn.sigmoid(logits)
    numpyro.deterministic('mu', mu)
\end{lstlisting}
\caption{\textbf{The \textsc{irt} candidate programs, as \emph{verbatim} LLM-generated \textsc{NumPyro} code.}
Each is one rung of the ladder \cref{eq:mirt}: the model index $m$ selects \emph{which} per-unit latents
are present (a \texttt{numpyro.plate} over students / questions), so the programs have different latent counts
($1,\ S,\ S{+}Q,\dots$). The \texttt{deterministic('mu',...)} node returns $P(\text{correct})$ at every $(s,q)$
cell; the evidence-SMC marginalises the plated latents to score each program. The LLM also emits 3PL and MIRT
(\cref{eq:mirt}) across proposals.}
\label{fig:irtcode}
\end{figure}

\paragraph{Results on a larger problem.}

\begin{figure}[t]
\centering
\includegraphics[width=\textwidth]{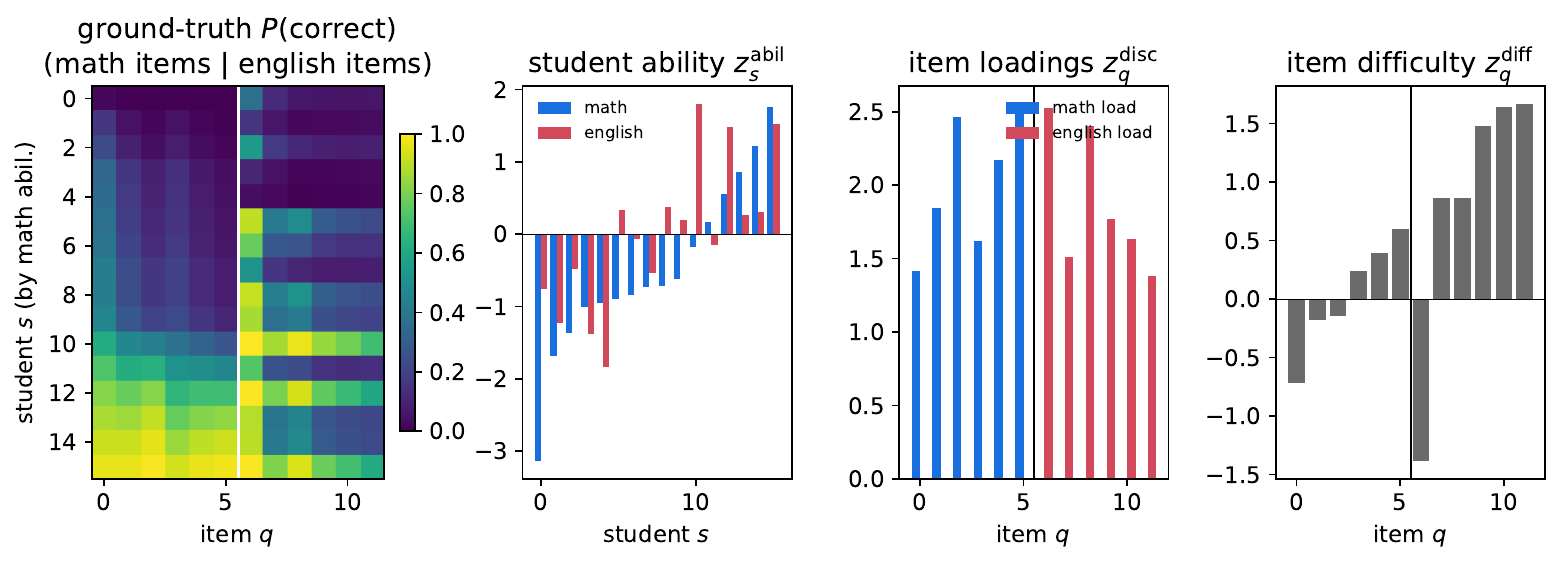}
\caption{\textbf{Ground truth for the stronger-signal MIRT \textsc{irt} instance.} Left: the ground-truth
$P(\text{correct})$ response grid ($16$ students $\times$ $12$ items, sorted by math ability; the vertical rule
splits the math and english item blocks). Right: the generative latents --- a \emph{two}-dimensional student
ability (math, english), per-item loadings on each axis, and per-item difficulty. The hidden second (english)
ability axis is what a 1PL/2PL model cannot capture and MIRT can.}
\label{fig:irttruth}
\end{figure}

To test the agents in a problem setting where there is more data,
we create an example
(not part of \boxing) as follows.
The true data generating process is a MIRT model with two latent dimensions,
representing the domains of math and English.
There are 16 students and 12 items;
each question loads onto one of these two domains,
as specified by $\theta_q^{\text{disc}}$,
and each student has different abilities
on these two domains, as specified by $\theta_s^{\text{abil}}$.
In addition each question has an intrinsic difficulty,
as specified by $\theta_q^{\text{diff}}$.
The resulting parameters are shown in
\cref{fig:irttruth}.
We also show the mean of the predictive distribution $p(y_{sq}=1)$,
which has a clear block structure (math vs English),
as well as a low-rank structure.

At each round $\expt$, the agent gets to pick
$N_\expt$ cells to examine; for each cell it observes a binary label
$y_{sq} \sim \text{Ber}(\mu_{sq})$.
The total number of counts for a cell is then
$N_\expt(s,q)$, of which $S_\expt(s,q)$ are successful,
which the agent models using a Binomial likelihood.
Evaluation measures squared error against the oracle probabilities over all
192 cells, including queried cells; it is not unseen-cell generalization.

In \Cref{fig:codelat-irt}, we let the agent pick $N_\expt=150$ cells per round,
chosen freely from all 192 cells.
This controlled PPL illustration uses a fixed five-model ladder for MDA
(constant, student-only, 1PL, 2PL, MIRT), and a single evolving LLM-written
program for Apprentice. Both use numerical expected Brier-risk reduction
to allocate responses; this is distinct from the headline EIG surrogate.
MDA has lower probability error in this illustration; no significance
claim is made from the single synthetic instance.
In \cref{fig:IRTpost}~(right), we see that as the sample size increases,
the MDA agent shifts its probability mass to more complex models,
eventually converging on  the true MIRT model.

\subsection{Results on Location}
\label{app:locationResults}

The true predictive model  for the
\textsc{location} (source finding) problem,
at location $x\in\real^2$, is
\begin{equation}
  p(Y_x\mid m)
  =\int \gauss\!\Big(Y_x\,\Big|\, b+\sum_{k=1}^{K}\frac{\alpha}{c+\lVert x-\theta_k\rVert^2},\ \sigma^2\Big)\,
  \left[\prod_{k=1}^K p(\theta_k) \right]
  p(\theta_b,\theta_c,\theta_{\alpha},\theta_{\sigma}) d\theta
  \label{eq:location}
\end{equation}
with latent source positions $\theta_k\sim\gauss(0,I_2)$ and
fixed globals $(b,c,\alpha,\sigma)$ all marginalized out.
The model structure $m$ corresponds to the number of sources $K$.
The agent-generated code for each of these models is shown in
\cref{fig:locationcode}.

\begin{figure}[t]
\begin{lstlisting}[style=pytiny]
# Location: source-count (K) discovery. Verbatim LLM-generated NumPyro programs (mda2.propose_numpyro).
# ctx["features"] is the (C,d) matrix of all candidate designs; the observation
# family (Bernoulli/Gaussian) is a task constant applied externally by the SMC.
import numpyro, numpyro.distributions as dist, jax, jax.numpy as jnp

def one_source(ctx):   # Single point source with unknown location and amplitude over constant background.
    F = ctx['features']
    b = numpyro.sample('b', dist.HalfNormal(1.0))
    alpha = numpyro.sample('alpha', dist.HalfNormal(5.0))
    m = numpyro.sample('m', dist.HalfNormal(1.0)) + 0.05
    theta = numpyro.sample('theta', dist.Uniform(-2.5*jnp.ones(2), 2.5*jnp.ones(2)))
    d2 = jnp.sum((F - theta)**2, axis=1)
    mu = b + alpha / (m + d2)
    numpyro.deterministic('mu', mu)

def two_sources(ctx):   # Two point sources with unknown locations sharing amplitude/scale.
    F = ctx['features']
    b = numpyro.sample('b', dist.HalfNormal(1.0))
    alpha = numpyro.sample('alpha', dist.HalfNormal(5.0))
    m = numpyro.sample('m', dist.HalfNormal(1.0)) + 0.05
    with numpyro.plate('sources', 2):
        tx = numpyro.sample('tx', dist.Uniform(-2.5, 2.5))
        ty = numpyro.sample('ty', dist.Uniform(-2.5, 2.5))
    d2 = (F[:,0:1] - tx[None,:])**2 + (F[:,1:2] - ty[None,:])**2
    mu = b + jnp.sum(alpha / (m + d2), axis=1)
    numpyro.deterministic('mu', mu)

def three_sources(ctx):   # Three point sources with shared kernel parameters.
    F = ctx['features']
    b = numpyro.sample('b', dist.HalfNormal(1.0))
    alpha = numpyro.sample('alpha', dist.HalfNormal(5.0))
    m = numpyro.sample('m', dist.HalfNormal(1.0)) + 0.05
    with numpyro.plate('sources', 3):
        tx = numpyro.sample('tx', dist.Uniform(-2.5, 2.5))
        ty = numpyro.sample('ty', dist.Uniform(-2.5, 2.5))
    d2 = (F[:,0:1] - tx[None,:])**2 + (F[:,1:2] - ty[None,:])**2
    mu = b + jnp.sum(alpha / (m + d2), axis=1)
    numpyro.deterministic('mu', mu)
\end{lstlisting}
\caption{\textbf{The \textsc{location} programs, as \emph{verbatim} LLM-generated \textsc{NumPyro} code.} Model
discovery is over the \emph{number} of sources $K$: the LLM writes a ladder \texttt{one\_source},
\texttt{two\_sources}, \texttt{three\_sources},\dots, each a \texttt{numpyro.plate} of $K$ latent source
positions $\theta_k$ over the exact
inverse-square signal field $b+\sum_k\alpha/(c+\lVert x-\theta_k\rVert^2)$ (\cref{eq:location}). The
\texttt{deterministic('mu',...)} node is the mean observable at every candidate probe.
}
\label{fig:locationcode}
\end{figure}

\begin{figure}[t]
\centering
\begin{subfigure}[t]{0.40\textwidth}\centering
  \includegraphics[width=\textwidth]{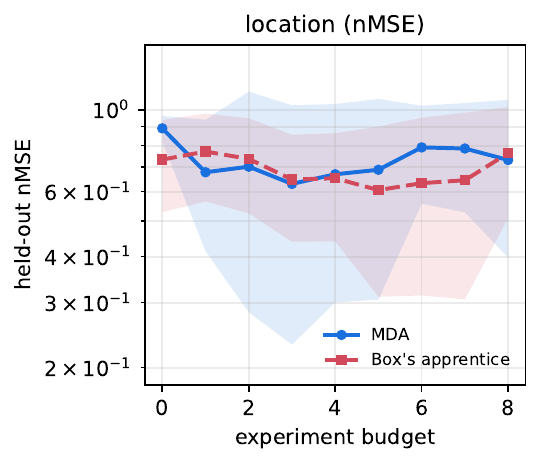}
  \caption{data efficiency}\label{fig:codeloc-curve}\end{subfigure}\hfill
\begin{subfigure}[t]{0.40\textwidth}\centering
  \includegraphics[width=\textwidth]{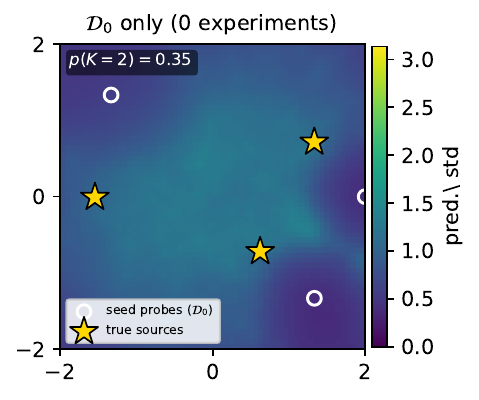}
  \caption{shared $\mathcal{D}_0$ ($0$ experiments)}\end{subfigure}

\medskip
\begin{subfigure}[t]{0.40\textwidth}\centering
  \includegraphics[width=\textwidth]{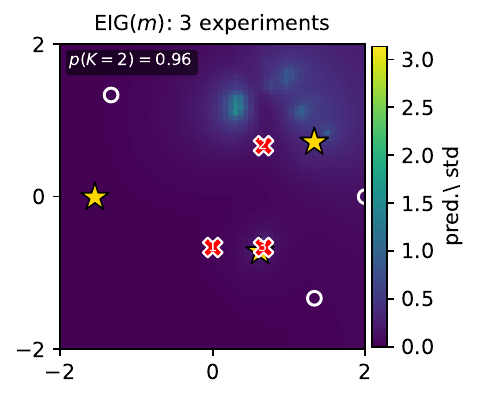}
  \caption{$3$ EIG experiments}\end{subfigure}\hfill
\begin{subfigure}[t]{0.40\textwidth}\centering
  \includegraphics[width=\textwidth]{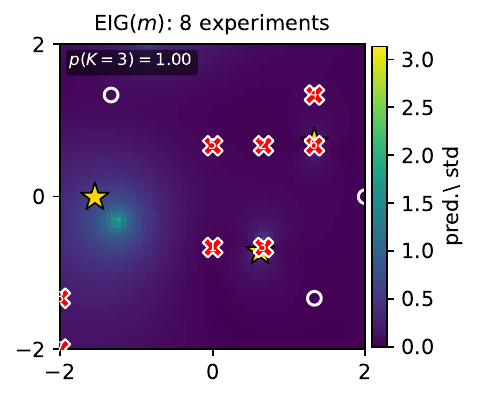}
  \caption{$8$ EIG experiments}\end{subfigure}
\caption{\textbf{Code-MDA on \textsc{location} (source finding over LLM-authored \textsc{NumPyro} programs).}
  (\subref{fig:codeloc-curve}): held-out nMSE vs.\ budget ($20$ seeds, median $\pm$ IQR/2).
  The remaining panels show the posterior-predictive std of the
log-signal field (viridis) as EIG experiments accumulate data,
starting from the \emph{same} shared passive dataset
$\mathcal{D}_0$ (white circles $=$ the $3$ seed probes given to every agent), with the red $\times$ the
probes EIG chooses, and gold stars the true $K{=}3$ sources. EIG
places probes where the $K$-source programs disagree, collapsing the field uncertainty and driving the
posterior over the source count to the truth ($p(K{=}3\mid\data){\to}1$) --- MDA's evidence-SMC \emph{discovers} the
right structure.
}
\label{fig:codelocation}
\end{figure}

The performance of MDA and Apprentice on this domain are shown in \cref{fig:codeloc-curve}.
Both perform very similarly.
The other panels show how the EIG objective causes the agent to place probes
in locations that reduce its uncertainty about how many sources there are
(i.e., the  model identity).

We also tried maximizing the EIG for the model and its parameters
(\cref{eq:voiFull}) --- which would additionally pin down the source
locations $\theta_k$ --- but we found that the
greedy variance
objective is too myopic.
In particular, because the signal field is near-singular next to
a source, it piles probes onto a single hypothesized source (where a
few particles' predictions blow up) rather than spreading them, giving
unstable results worse than random design.

\eat{
In principle one might think that  task-driven VoI,
using the task-driven predictive objective,
would work better, but it suffers from the same problem:
occasionally task-VoI steers a probe into that near-singular region,
and then the prediction error blows up, and it takes many seeds
to average this effect away, thus making the VoI estimate unreliable.
}

In summary, 
on this domain active design gives no advantage: the EIG policy and
random sampling perform equally well.
(\cref{fig:codelocation}(c,d) illustrate where the EIG policy places its probes;
the advantage it would buy on a
more identifiable field simply vanishes here.)

\clearpage
\section{Further related work}
\label{app:related}

Here we expand on connections to prior work that we did not have space
for in \cref{sec:related}.

\subsection{Other interactive learning benchmarks}

Various benchmarks have been developed to evaluate agents that learn
about a world by \emph{interactive experimentation}.  In this paper,
we build on \ActiveChem
\citep{kabra2026autoscilab}, and \textsc{BoxingGym}
\citep{gandhi2025boxinggym}.
We discuss some other benchmarks below.

\paragraph{NewtonBench.}
\textsc{NewtonBench} \citep{zheng2026newtonbench} contains 324
interactive law-discovery tasks across 12 areas of physics.  It generates
memorization-resistant problems by systematically modifying canonical laws
(while preserving dimensional consistency), and varies both the complexity of
the hidden equation and the complexity of the larger system in which it is
embedded.  The agent chooses experiments and must recover the hidden equation,
with success assessed primarily by symbolic equivalence. This complements
our emphasis on predictive accuracy under new interventions, including
partially observed latent dynamics.

\paragraph{Science-Gym.}
\textsc{Science-Gym} \citep{cerrato2026sciencegym} is a Gym-compatible
testbed in which an agent collects experiments and then seeks an interpretable
equation.  Its seven environments cover the lever, projectile motion, an
inclined plane, Lagrange points, the brachistochrone, an epidemiological SIRV
model, and droplet friction; the interface can vary the observations and amount
of reward supervision.  Its reported ``Threshold-and-Save'' baseline uses an
RL policy to find high-reward episodes and applies guided symbolic regression
to the retained tabular data. Our experiments instead score forecasts on
held-out interventions, without a discovery-directed reward.

\paragraph{SciGym.}
\textsc{SciGym}  is a  systems-biology dry lab in
which an agent recovers the missing reactions of a curated
\textsc{SBML} network under a fixed experiment budget, scored on both
structural recovery and simulated-trajectory error.
The underlying models are ODEs whose laws must be discovered.

\paragraph{ARC-AGI-3.}
ARC-AGI-3 \citep{ARC3} evaluates interactive exploration and goal-directed
action in unfamiliar visual environments. Unlike the earlier static ARC
input/output tasks, it is not simply transductive prediction at a supplied
test input. MDA instead evaluates scientific-model forecasts on held-out
experimental conditions, with numerical inference in an explicit mechanistic
model class.

\paragraph{ZendoWorld.}
\citep{Koehler2026} introduces ZendoWorld, which requires solving
perception (from simple 3d scenes) in addition to rule induction and
active hypothesis testing.  In this paper, we focus on numeric input,
but ultimately an AI scientist system needs to be able to process raw
visual inputs (and other modalities) as well.

\paragraph{Other benchmarks.}
\textsc{LLM-SRBench} \citep{shojaee2025srbench} targets equation discovery
with problems designed to reduce memorization, while \textsc{Auto-Bench}
\citep{chen2025autobench} uses causal-graph-based interactive environments.

\subsection{Other ``AI-scientist'' systems}
\label{app:aiscientist}

A fast-growing line of work builds LLM \emph{agents} that automate
parts --- or all --- of the scientific workflow. The literature-grounded
systems below contrast with MDA along two useful axes. \emph{(a) What is discovered:} an
open-ended \emph{natural-language hypothesis} or a candidate design,
rather than an explicit, uncertainty-quantified \emph{mechanistic
model} that forecasts unseen interventions. \emph{(b) How candidates
are adjudicated and chosen:} by \emph{LLM judgement} ---
self-critique, simulated debate, or Elo tournaments --- and literature
support, rather than by a Bayesian posterior with model evidence and
an information-theoretic experiment-design objective. We summarise the
closest exemplars.

\paragraph{Co-Scientist \citep{Gottweis2026}.}
A Gemini-based multi-agent ``structured thinking engine'' for
hypothesis generation. A \emph{Supervisor} orchestrates specialised
\emph{Generation}, \emph{Reflection}, \emph{Ranking},
\emph{Evolution}, \emph{Proximity}, and \emph{Meta-review} agents that
continuously generate, critique (via ``simulated scientific debate''),
and refine hypotheses; candidates are ordered by an \emph{Elo
tournament} and improved by scaling test-time compute. It is validated
on drug repurposing (acute myeloid leukaemia, confirmed \emph{in
vitro}), novel-target discovery, and mechanisms of antimicrobial
resistance. In our terms it performs \emph{black-box optimisation in
natural-language hypothesis space} against an internal LLM-judge
fitness (the tournament): there is no explicit posterior over
mechanisms, no likelihood, and no experiment-design step --- the
system proposes and scores hypotheses, while wet-lab validation is
external. MDA instead maintains approximate numerical weights over
\emph{mechanistic} models and chooses interventions by expected
information gain (\cref{app:algo}).

\paragraph{Robin \citep{Ghareeb2026}.} A lab-in-the-loop system that couples literature-search agents
(\emph{Crow} and \emph{Falcon}, built on PaperQA2) with a
data-analysis agent (\emph{Finch}). Robin selects an experimental
assay by literature review, generates therapeutic candidates by
literature synthesis, ranks them with an \emph{LLM-judged tournament}
on the strength of the supporting literature, and --- after a human
runs the wet-lab experiment --- has Finch analyse the raw data (a
consensus over eight stochastic analysis trajectories) and propose
follow-up assays. It drove a real, validated discovery: ripasudil
(ROCK inhibition) to enhance retinal-pigment-epithelium phagocytosis
in dry age-related macular degeneration, with a follow-up RNA-seq
experiment implicating \textsc{abca1}. Like Co-Scientist, its
hypotheses are grounded in the \emph{published literature} and
adjudicated by LLM judgement rather than a posterior; experiments are
proposed from literature, not by an information-theoretic design
objective, and the deliverable is a candidate plus its interpretation,
not a mechanistic forecaster of interventions the agent never ran.

\paragraph{Tool-using and lab-automation agents.} Earlier systems automate narrower slices, with an LLM
orchestrating external tools. Coscientist \citep{boiko2023}, distinct from
Gottweis et al.'s Co-Scientist described above, uses GPT-4
to plan and \emph{physically execute} chemistry experiments on
robotic/cloud platforms; ChemCrow \citep{bran2024chemcrow} augments an
LLM with a suite of chemistry tools for synthesis planning and
execution. In materials and biology, multi-agent design systems such
as SciAgents \citep{ghafarollahi2024sciagents} (intelligent
graph-reasoning agents) and the Virtual Lab
\citep{swanson2024virtuallab} (an agent ``team'' that designed
experimentally validated nanobodies) follow the same
LLM-proposes-and-adjudicates template. These add genuine tool use and
real actuation, but none maintains an explicit posterior over
\emph{competing} mechanisms, nor selects experiments to maximise
information about them.

\paragraph{Numerical equation discovery.}
SINDy \citep{brunton2016sindy} fits sparse combinations of candidate nonlinear
functions to measured dynamics; SINDYc \citep{brunton2016sindyc} includes control inputs.
It is a relevant non-LLM comparator, but applying the basic formulation to our
partially observed systems requires choosing observable or delay-embedded
coordinates and handling noise and missing observations, rather than providing
the hidden simulator states. Our glucose comparison includes an observation-space SINDYc control with
training-only library selection and stability screening (\cref{app:glucosesindy});
we do not yet evaluate a corresponding sparse-dynamics baseline for HH.
The PySR system \citep{cranmer2023pysr} uses
evolutionary structure search and coefficient fitting
for symbolic regression of  static equations;
it is used internally by \SciLab, so we already indirectly compare to
it in \cref{fig:mainchem-curve}.

\paragraph{Active theory learning.}
ATLAS \citep{Elteto2026} alternates fitting an ensemble of interpretable
DisRNNs and optimizing experiments to distinguish their predictions.
Its evaluation recovers Q-learning and actor--critic agents from bandit
behavior; it also examines GRU ensembles. MDA shares the population-based
active-learning principle but searches explicit non-neural algebraic,
compartmental, and stochastic channel models across domains.
The different interfaces prevent a plug-in baseline comparison.

\paragraph{Automated research pipelines.} At the far end, the AI Scientist \citep{lu2024aiscientist} automates
the entire machine-learning research loop --- ideation, coding,
running experiments, and writing the manuscript --- with LLM agents
and an LLM reviewer. This targets the \emph{breadth} of the pipeline
rather than the depth of a single inference step, and again uses an
LLM as the evaluator rather than a calibrated model posterior.

\paragraph{Automated cognitive scientists: \emph{auto-psych} \citep{Prystawski2026} and \emph{AutoCog}
\citep{Jagadish2026}.} Two concurrent systems automate the discovery
of \emph{computational cognitive} theories and are the closest prior
work to MDA in \emph{method}, not merely in ambition: unlike the
literature-grounded, LLM-judged systems above, both represent
hypotheses as \emph{executable probabilistic models} and adjudicate
them by \emph{quantitative predictive fit} rather than an LLM
referee. The \emph{auto-psych} system runs nested agentic loops --- an
inner loop that conjectures, fits, and critiques probabilistic
cognitive models, and an outer loop that designs experiments, launches
them on online participants, and analyses the returned data ---
demonstrated on the classic problem of judging which coin-flip
sequences look ``random''.  Notably, it evaluates the discovered
theories \emph{almost exactly as we do}: models are scored by
\emph{held-out} expected log predictive density (ELPD, via
Pareto-smoothed-importance-sampling leave-one-out cross-validation,
with per-stimulus RMSE and $R^2$), and ground-truth \emph{recovery} is
quantified by the RMSE between the ground-truth model's responses and
the best-fitting model's responses over an enumerated set of stimuli
--- precisely our pairing of a held-out predictive loss
($\ell_{\mathrm{predict}}$) with a recovery error against the true
model over a query set ($\ell_{\mathrm{recover}}$), and pointedly
\emph{not} the high-variance ``explain-to-a-novice'' judge that we and
\textsc{BoxingGym} set aside. It also \emph{designs} experiments by
expected information gain over the model set,
$\mathrm{EIG}(c)=H(M)-\sum_{r}P(r\mid c)\,H(M\mid r,c)$, greedily
selecting the stimuli that best discriminate the competing models, and
reports theories that fit human data better than ones drawn from the
scientific literature. \emph{AutoCog} closes the same loop for
\emph{decision-making}: LLM agents propose competing executable
theories, design maximally \emph{discriminative} experiments, collect
behaviour from online participants, score each theory by its
generative fit, and diagnose failures into successor theories; it
discovered --- then \emph{preregistered and confirmed} on fresh
participants --- a novel multi-cue decision rule with diminishing
sensitivity to feature values.

Both converge with MDA on the two commitments we argue are essential
--- \emph{predictive/posterior} adjudication in place of an LLM judge,
and \emph{information-driven} experiment selection --- which we read
as independent evidence for the recipe. MDA is distinguished by three
choices. \emph{(i)~$\mathcal M$-open search:} auto-psych's EIG is a
mutual information taken under a \emph{uniform prior over a model
registry}, whereas MDA maintains an evidence-weighted posterior and
\emph{expands} the hypothesis set when a predictive check fails
(\cref{alg:mda}), rather than re-ranking a fixed
slate. \emph{(ii)~Numerical design:} our implemented acquisition is also a
model-discrimination EIG surrogate (\cref{eq:voi}); task-oriented VoI is
a future direction, not a distinguishing empirical claim. \emph{(iii)~Scope:} each is
specialised to a single cognitive-science paradigm with a bespoke
likelihood, whereas MDA is one domain-general engine (\cref{app:algo})
spanning chemistry, glucose dynamics, electrophysiology, and
the \textsc{BoxingGym} suite.

\paragraph{Amortized design.}
ASIG \citep{hartmann2026asig} trains LLM information-gathering policies
with an EIG reward. This complements our online numerical acquisition:
it amortizes the design policy rather than fitting executable scientific
models at each experiment.

\paragraph{BED-LLM \citep{bedllm}.}
This recent paper applies sequential Bayesian experimental design to make an
LLM gather information \emph{adaptively} --- multi-turn question
asking (the 20-Questions game) and eliciting a user's latent
preferences (movie recommendation) --- rather than to discover a
mechanistic model. Like MDA, it chooses each query by maximising the
expected information gain (EIG),
but it uses probability values from the LLM itself,
rather than generating an explicit probability model.
 MDA differs in what is discovered and how the posterior is
maintained: BED-LLM targets a \emph{single categorical latent} (the
secret object, the preferred item) and reasons in the LLM's space of
\emph{answers}, whereas MDA maintains an evidence-weighted, $\mathcal
M$-open archive of \emph{executable models} (rate laws, compartment ODEs,
channel mechanisms) with numerically fitted parameters,
and evaluates held-out predictive accuracy.

\paragraph{Where MDA sits.}
The breadth-first systems above (Co-Scientist, Robin, the tool-using agents, the
AI Scientist) automate the \emph{breadth} of discovery --- literature
synthesis, hypothesis ideation, lab actuation, manuscript generation
--- and prioritise by LLM judgement or self-play tournaments (the two
automated cognitive scientists are the exception --- they share MDA's
quantitative, predictive-fit core, but on a single paradigm). MDA
automates the \emph{depth} of one step: fitting executable predictive
models and selecting informative interventions, using numerical likelihood
and evidence approximations rather than an LLM judge. The two are
complementary --- a Co-Scientist- or Robin-style system could invoke
MDA as a quantitative inner loop where an explicit mechanistic model
is needed, and their literature agents could supply MDA's structural
priors. This is precisely the division of labour our $\mathcal M$-open
loop makes concrete (\cref{alg:mda}): \emph{the LLM proposes; numerical
inference scores models and guides design}.

\clearpage
\section{LLM prompts}
\label{app:llm-prompts}
\label{app:prompts}

This appendix documents agent interfaces and prompt templates.
Historical examples are marked explicitly and are not the exact prompts
of every current run; archived per-call transcripts are the source of record.
For the existing domains, our prompts are very
similar to the ones used in the original papers (reproduced below);
we only make changes when our required output is different
(because we use LLMs in a different way to the baselines).
 MDA uses the LLM only to  \emph{propose} structures.
 ICL (+MDA design) forecasts from MDA's history; ICL (+LLM design) also
chooses its own experiments.
 For each existing benchmark, we also run their own
 agent (\SciLab\ and \Apprentice),
 which use the LLM in different ways
 (e.g., experiment design and/or forecasting).
The main-text runs use GPT-5.6-Luna at medium reasoning effort.
Older templates and BoxingGym results used Opus~4.7; their generation settings
must not be read as universal defaults for the current comparisons.

\subsection{\CHEMbench (enzyme rate laws, \S\ref{sec:chem})}
\label{app:chemprompt}

\subsubsection{MDA proposer}

The LLM proposer
sees the experiments
collected so far (design inputs $\to$ observed initial rate $r_0$) and --- when refining an existing pool ---
the forms already tried together with their residuals (so it does not re-propose dead ends, plus which input
each residual correlates most with) and the remaining budget and phase.

\promptfile{MDA proposer --- rate law (\CHEMbench)}{prompts/chem_proposer.txt}

\subsubsection{\SciLab agent}
The baseline's upstream equation-discovery system prompt is shown below
(from \texttt{autoscilab/llm/prompts.py}). In the audited run it is paired with the generic goal
``Discover the governing rate law for this enzyme-catalyzed biochemical reaction,'' the benchmark's universal
grammar, and no domain tags. Ground-truth evaluation is disabled during discovery and becomes available only
after the final submitted law is frozen.
\promptfile{\SciLab\ --- equation discovery}{prompts/scilab_baseline.txt}

\subsection{\boxing (\cref{app:boxing})}
\label{app:boxingprompt}

\subsubsection{MDA proposer}

For the covariate-regression (GLM) domains
(\textsc{dugongs}/\textsc{death}/\textsc{peregrines}/\textsc{hyperbolic}),
the LLM proposes the mean function $f$
  as JSON (name, expression, parameter
  bounds).
\promptfile{MDA proposer --- GLM mean function (\boxing)}{prompts/boxing_glm.txt}

For the PPL implementations (\textsc{irt}/\textsc{location}/\textsc{survival}),
the LLM proposes a  \textsc{NumPyro} program,
which defines the priors over latents via \texttt{numpyro.sample}, a
\texttt{numpyro.plate} for per-unit latents,
and a \texttt{deterministic('mu',...)} node for the mean.
\promptfile{MDA proposer --- NumPyro program (\boxing)}{prompts/boxing_numpyro.txt}

\subsubsection{\Apprentice agent}
The original Box's Apprentice \citep{li2024automated} (a.k.a.\ Box's~LOOP) is an all-LLM loop: an LLM
\emph{synthesises} a \textsc{PyMC} probabilistic program for the data; the program is fit; an LLM \emph{critic}
inspects the posterior-predictive fit and returns revision hypotheses; and the loop repeats, with a separate
LLM call choosing the next experiment. Its two core prompts --- the \textsc{PyMC} program-synthesis prompt and
the critic prompt --- are reproduced here for reference (verbatim from
\texttt{boxing\_gym/agents/box\_loop\_prompts.py}, with runtime placeholders shown as \texttt{<...>}):
\promptfile{Box's~LOOP (original baseline) --- PyMC program synthesis \textnormal{[reference]}}{prompts/boxloop_pymc_proposal.txt}
\promptfile{Box's~LOOP (original baseline) --- model critic \textnormal{[reference]}}{prompts/boxloop_pymc_critic.txt}

Our reimplementation keeps this propose~$\to$~fit~$\to$~critique~$\to$~refine loop but makes two changes so the
comparison shares a model generator and numerical fitter. (i)~It uses \textsc{NumPyro} instead of \textsc{PyMC}, fit by the \emph{same}
evidence-SMC as MDA. (ii)~In place of the heavily-scaffolded prompt above, the program is generated by
\emph{MDA's own} structured proposer --- the GLM proposer on the covariate-regression domains and the
\textsc{NumPyro} proposer on the latent-variable domains (both shown in \S\ref{app:boxingprompt}, above) --- so
the \Apprentice\ and MDA share an \emph{identical} model generator and differ only in the inference/design
strategy (a single evolving program vs.\ MDA's posterior over programs with VoI design).
This is an adapted single-program control, not an evaluation of the unmodified
published Box's Apprentice system. The only
\Apprentice-specific prompts are therefore the experiment-design call and the residual critique fed back to that
shared proposer. First it picks its next experiment by an LLM call conditioned on its current program:
\promptfile{\Apprentice\ (ours) --- experiment design}{prompts/boxing_design.txt}
Then each round it re-fits the single evolving program and feeds its residual back to the shared proposer:
\promptfile{\Apprentice\ (ours) --- residual critique fed to the shared proposer}{prompts/boxing_apprentice_critique.txt}

\subsubsection{ICL forecaster}
The ICL baseline  forecasts the held-out outcomes from the data table alone,
with no model:
\promptfile{ICL forecaster (\boxing)}{prompts/boxing_icl.txt}

\subsection{Electrophysiology (ion channels, \S\ref{sec:bio})}
\label{app:ephysprompt}
\label{app:promptsHH}

This is our new benchmark, so there are no existing prompts to compare to.
Both methods receive the same observation type and protocol information.
Their processing differs: MDA computes summaries and fit diagnostics from
these observations, whereas ICL forecasts directly from the observation table.
These diagnostics contain no additional oracle measurements.

\paragraph{Current sparse-observation interface.}
The headline HH agents receive the protocol and exactly 20 observed
time--voltage pairs per experiment, not a dense trace or the oracle's true
six-feature vector. The numerical PF conditions on these sparse voltages.
The ICL forecaster sees the accumulated observations and the 24 query schedules,
then predicts six features per query. ICL (+LLM design) sees the unused action
menu but not the held-out query schedules when choosing actions; it receives
no mechanism archive or model-fitting tools. A malformed action receives one
format-only retry and then the smallest unused index; a missing forecast is
scored as a zero feature vector. These fallbacks are retained in the audit.
The fixed text below reproduces the current implementation's design and
forecast templates; braces around data/menu blocks indicate substituted
run-specific observations rather than extra instructions.
\promptfile{Current HH ICL (+LLM design): action selection}{prompts/hh_icl_llm_design_current.txt}
\promptfile{Current HH ICL: sparse-voltage forecast}{prompts/hh_icl_forecast_current.txt}

\paragraph{Historical feature-summary template (not the headline interface).}
The following older example compresses each trajectory to six features:
\promptfile{What the LLM sees as ``data'' (\HHbench)}{prompts/hh_data_example.txt}
$n_{\mathrm{test}}$/$n_{\mathrm{pre}}$ are the test-window / pre-pulse spike counts;
$\mathrm{run\text{-}down}=n_{\mathrm{pre}}-n_{\mathrm{test}}$; $\mathrm{adaptation}$ is early-minus-late
test spikes; $V_{\min}$/$V_{\mathrm{end}}$ are the sub-threshold sag depth and steady-state tail (mV).
This older representation must not be confused with the raw 20-point
observations used in the current comparison.

\subsubsection{MDA proposer}
\label{app:hhproposer}

The current generic proposer specifies additional currents or slow-Na
inactivation through executable JSON, not a named-archetype menu. It sees
intervention schedules, 20 observed voltages, and summaries computed from
those sparse observations, not dense oracle features. Later calls include
archived mechanisms and predictive-check diagnostics.
The legacy system sentence still says ``spike-count data''; the payload
below includes the sparse voltages used in the headline run.
\promptfile{Current MDA proposer --- generic HH mechanisms}{prompts/hh_proposer_current.txt}

\eat{
A more restricted variant instead asks
for candidate channel \emph{compositions} from a fixed channel menu, given a one-line phenotype description:
\promptfile{MDA proposer --- channel compositions (\HHbench)}{prompts/hh_channels.txt}
}

\subsubsection{ICL forecaster}
\label{app:hhbaseline}

The following is the historical six-feature forecaster template, not the
current sparse-observation prompt described above:
\promptfile{ICL forecaster (\HHbench)}{prompts/hh_icl_stoch.txt}
Every method (MDA and the ICL forecaster) is conditioned on the \emph{same} passive seed $\data_0$ ($2$
observations); the $N_a{=}0$ point is this forecaster given $\data_0$ alone --- there is no data-free arm, and
no method is uniquely handicapped. These templates mirror the code (\texttt{icl\_hh.py} / the released
\texttt{neuronbench}).

\subsection{\glucose: compartment models and direct forecasting}
\label{app:glucose_prompts}
All glucose LLM calls use GPT-5.6 Luna with medium reasoning effort.
The following system text reproduces the implemented prompts; user payloads
are shown schematically, with run-specific arrays substituted at execution.
Patient identity, age, hidden states, and simulator parameters are withheld.
MDA sees the fitted candidate pool and its training predictions, whereas ICL
sees observations and input schedules, not MDA's fitted models.

\subsubsection{MDA model proposer}
The census prompt includes explicit parent IDs and edit descriptions for
genealogy logging. This metadata is stored separately from executable ODE
expressions; it does not require a single parent when a proposal combines
mechanisms. The earlier illustrative pilot used the same base prompt without
the genealogy instructions. The multi-cut follow-up reuses these frozen
models and makes no new model-proposal calls.
\promptfile{Glucose MDA: ODE proposal and genealogy}{prompts/glucose_proposer_current.txt}

\subsubsection{ICL forecasting and design}
The completed census forecaster conditions on a 45-minute prefix.
ICL (+MDA design) replays MDA's history; ICL (+LLM design) uses the following
additional acquisition prompt. Both share the same forecast prompt.
\promptfile{Glucose ICL: completed census forecast}{prompts/glucose_icl_forecast_current.txt}
\promptfile{Glucose ICL (+LLM design): experiment selection}{prompts/glucose_icl_design_current.txt}
The separate multi-cut evaluation uses an explicit empty-prefix contract
at cut zero and variable-length forecast targets. This prompt is not the
one used for the completed census curves.
\promptfile{Glucose ICL: multi-cut forecast}{prompts/glucose_icl_multicut.txt}

\clearpage
\end{document}